\documentclass[11pt, logo, onecolumn, copyright, colorlinks=true, allcolors=blue]{nvidiatechreport}
\usepackage{url}
\usepackage{graphicx} 
\usepackage{subcaption,ragged2e}
\usepackage[round]{natbib}

\usepackage[normalem]{ulem}
\usepackage[utf8]{inputenc} 
\usepackage[T1]{fontenc}    
\usepackage{tikz}
\usepackage{pgf-pie}
\usepackage{enumitem}
\usetikzlibrary{positioning}
\usetikzlibrary{calc}
\usepackage{array}
\usepackage{booktabs}
\usepackage{wrapfig}
\usepackage{caption} 
\usepackage{listings}
\usepackage{adjustbox}
\usepackage{footnote}
\usepackage{multirow}
\usepackage{amsmath}
\usepackage{amsfonts}
\usepackage{breakcites}
\usepackage{blindtext}
\usepackage{svg}
\usepackage[most]{tcolorbox}
\usepackage{tabularx}
\usepackage{graphicx} 
\usepackage{xcolor}
\usepackage{float}
\usepackage{needspace}
\usepackage{makecell}
\usepackage{colortbl}
\usepackage{stfloats}
\setlist[itemize]{topsep=1pt,itemsep=1pt}
\setlist[enumerate]{topsep=1pt,itemsep=1pt}
\usepackage{xparse}
\RenewDocumentCommand{\paragraph}{s o m}{%
  \par\medskip\noindent\textbf{#3}\quad\ignorespaces
}

\newcommand{\ourmodelfull}{NVIDIA Nemotron 3 Ultra\xspace}
\newcommand{\ourmodel}{Nemotron 3 Ultra\xspace}
\newcommand{\ourbasemodel}{Nemotron 3 Ultra 550B-A55B Base\xspace}

\newcommand{\supermodel}{Nemotron 3 Super\xspace}
\newcommand{\nanomodel}{Nemotron 3 Nano\xspace}

\title{\ourmodel: Open, Efficient Mixture-of-Experts Hybrid Mamba-Transformer Model for Agentic Reasoning}
\author{\large NVIDIA}
\date{}

\begin{document}

\begin{abstract}
\large \textbf{Abstract.}
\normalsize




We introduce \ourmodel, a 550 billion total and 55 billion active parameter Mixture-of-Experts Hybrid Mamba-Attention language model. We pre-trained \ourmodel on 20 trillion text tokens, then extended the context length to 1M tokens, and post-trained using Supervised Fine Tuning (SFT), Reinforcement Learning (RL), and Multi-teacher On-Policy Distillation (MOPD). Nemotron 3 Ultra is our most capable model yet, employing multiple key technologies - LatentMoE, Multi Token Prediction (MTP), NVFP4 pre-training, multi-environment RLVR, MOPD, and reasoning budget control. Nemotron 3 Ultra achieves up to $\sim6\times$ higher inference throughput as compared to state-of-the-art publicly available LLMs while attaining on-par accuracy. The state-of-the-art accuracy, high inference throughput, and 1M token context length make Nemotron 3 Ultra ideal for long-running autonomous agentic tasks. We open-source the base, post-trained, and quantized checkpoints, along with the training data and recipe on HuggingFace.

\end{abstract}

\maketitle

\section{Introduction}
\label{sec:intro}

We present Nemotron 3 Ultra, the largest and most capable model in the Nemotron 3 family~\citep{blakeman2025nvidia}. As LLM applications evolve from simple chatbots to long-running agents capable of autonomously writing code, conducting research, and completing complex tasks, the ability to deliver fast and efficient inference becomes increasingly important. Nemotron 3 Ultra addresses this by employing a Mixture-of-Experts hybrid Mamba-Attention architecture, leading to improvements along the inference-throughput-to-accuracy frontier. While Mixture-of-Experts help Nemotron 3 Ultra achieve better accuracy per active parameter, the hybrid Mamba-Attention architecture significantly improves inference throughput by reducing attention cost and KV cache footprint. Nemotron 3 Ultra achieves 5.9$\times$, 4.8$\times$, and 1.6$\times$ higher inference throughput compared to GLM-5.1-754B-A40B, Kimi-K2.6-1T-A32B, and Qwen-3.5-397B-17B respectively on 8K input / 64K output token setting while also attaining on-par accuracy across a wide range of agentic and reasoning benchmarks.

\begin{figure}[ht]
    \centering
    \includegraphics[width=\linewidth]{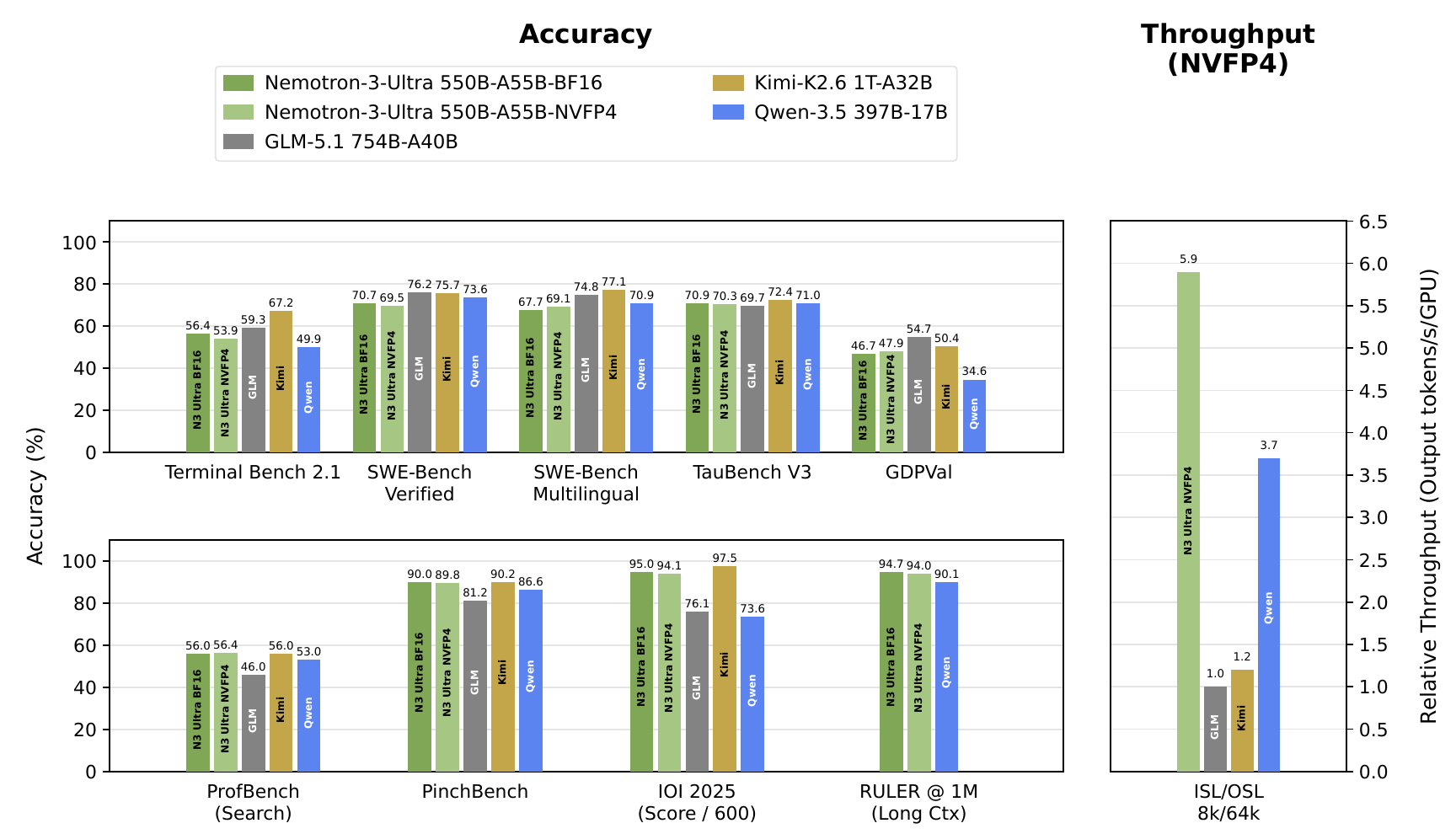}
    \caption{Accuracy and throughput comparisons for \ourmodel. Our model achieves on-par accuracy with other open LLMs while achieving significantly higher inference throughput on the 8K input / 64K output token setting. All throughput numbers are reported at max-throughput using NVFP4 precision on GB200. For Nemotron 3 Ultra, throughput numbers are obtained from TRT-LLM, while all other model numbers use vLLM. We run with and without speculative decoding, where available, and choose the best numbers for each model.}
    \label{fig:intro}
\end{figure}

We pretrained \ourbasemodel with all the key Nemotron 3 features and technologies~\citep{blakeman2025nvidia}, including NVFP4 pre-training, LatentMoE~\citep{latentmoe_tr}, and Multi Token Prediction (MTP)~\citep{gloeckle2024better}. We pretrained our base model in NVFP4 with 20 trillion text tokens using a Warmup-Stable-Decay learning rate schedule. Pretraining was divided into two phases with 15 trillion tokens of data in the first phase focusing on diversity and broad domain coverage followed by 5 trillion tokens of data in the second phase focusing on high quality data to refine model accuracy. LatentMoE helped us achieve better accuracy per parameter than standard Granular MoEs~\citep{dai2024deepseekmoe} while Multi Token Prediction (MTP) leads to faster inference with speculative decoding. Our pretrained base model achieves significantly higher accuracy than other publicly available base models, such as DeepSeek v3.2~\citep{deepseekai2024deepseekv32}, Mistral Large 3, Kimi-K2~\citep{kimiteam2025kimik2openagentic}, and GLM-4.5~\citep{5team2025glm45agenticreasoningcoding}.

We trained \ourmodel with an agent-focused post-training pipeline to improve its long-horizon reasoning, tool use, and autonomous task completion capabilities. The initial SFT stage used a carefully curated data mixture to build the model's foundational capabilities. This was followed by unified RLVR over a wide mix of reasoning, agentic, code, safety, usability, and chat environments. In parallel, more than ten domain-specialized teacher models were trained using targeted recipes, including agentic teachers built on a dedicated agentic SFT path. Finally, Multi-teacher On-Policy Distillation (MOPD) consolidated these teachers into Ultra through dense token-level guidance on student-generated rollouts. Ultra is also equipped with reasoning effort control, which supports inference-time adjustment of the accuracy–compute trade-off.

We are releasing the Base, Post-Trained, and NVFP4 quantized checkpoints on HuggingFace. Alongside the checkpoints, we are also open-sourcing the training recipes\footnote{\href{https://github.com/NVIDIA-NeMo/Nemotron}{https://github.com/NVIDIA-NeMo/Nemotron}}, data, and RL environments. 

\paragraph{Checkpoints}
\begin{itemize}
    \item \href{https://huggingface.co/nvidia/NVIDIA-Nemotron-3-Ultra-550B-A55B-NVFP4}{\texttt{Nemotron 3 Ultra 550B-A55B NVFP4} \includegraphics[height=0.9em]{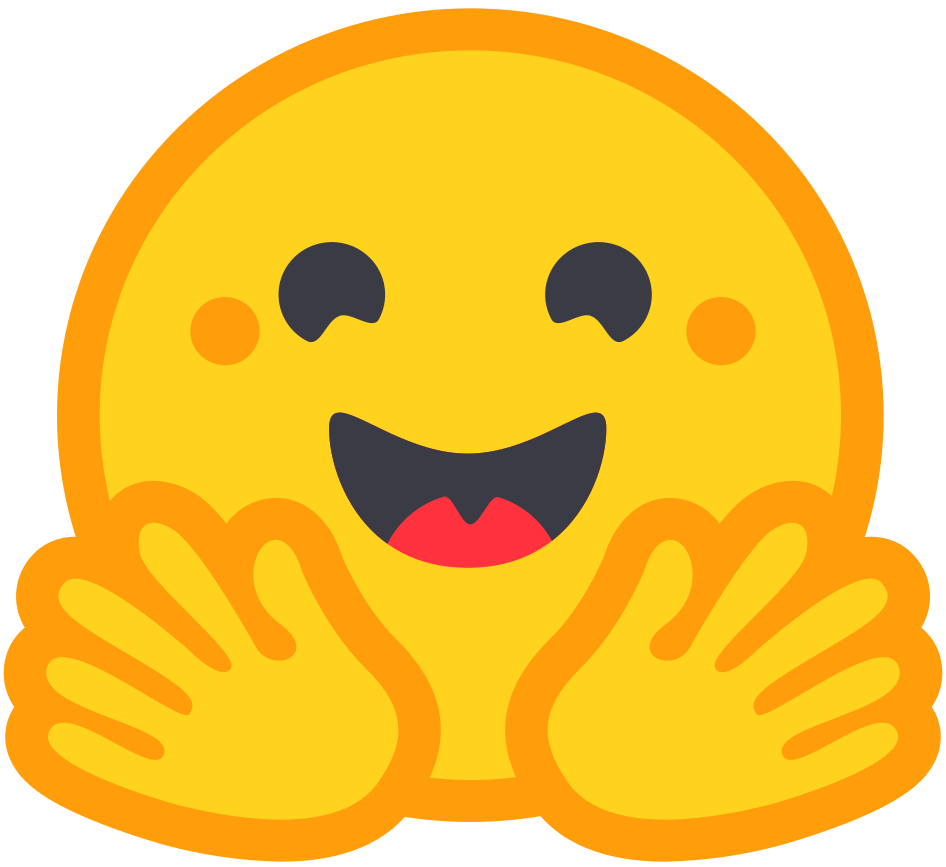}} : post-trained and NVFP4 quantized model
    \item \href{https://huggingface.co/nvidia/NVIDIA-Nemotron-3-Ultra-550B-A55B-BF16}{\texttt{Nemotron 3 Ultra 550B-A55B BF16} \includegraphics[height=0.9em]{assets/huggingface-color.png}} : post-trained model
    \item \href{https://huggingface.co/nvidia/NVIDIA-Nemotron-3-Ultra-550B-A55B-Base-BF16}{\texttt{Nemotron 3 Ultra 550B-A55B Base BF16} \includegraphics[height=0.9em]{assets/huggingface-color.png}} : base model
    \item \href{https://huggingface.co/nvidia/NVIDIA-Nemotron-3-Ultra-550B-A55B-GenRM}{\texttt{Nemotron 3 Ultra 550B-A55B GenRM} \includegraphics[height=0.9em]{assets/huggingface-color.png}} : GenRM used for RLHF

\end{itemize}

\paragraph{Data}
\begin{itemize}
    \item \href{https://huggingface.co/datasets/nvidia/Nemotron-Pretraining-Code-v3}{\texttt{Nemotron-Pretraining-Code-v3} \includegraphics[height=0.9em]{assets/huggingface-color.png}} : 173B tokens of fresh code data from GitHub through September 30, 2025.
    \item \href{https://huggingface.co/datasets/nvidia/Nemotron-Pretraining-Legal-v1}{\texttt{Nemotron-Pretraining-Legal-v1} \includegraphics[height=0.9em]{assets/huggingface-color.png}} : A collection of synthetic datasets intended to improve the legal capabilities of LLMs. In one ablation, adding these datasets to Nemotron 3 Nano pretraining boosted a proxy LegalBench average accuracy from 64.6 to 74.7.
    \item \href{https://huggingface.co/datasets/nvidia/Nemotron-Pretraining-Specialized-v1.2}{\texttt{Nemotron-Pretraining-Specialized-v1.2} \includegraphics[height=0.9em]{assets/huggingface-color.png}} : A collection of synthetic datasets aimed to improve LLM capabilities on factual recall, moral scenarios, and diverse generative and multiple choice questions.
    \item \href{https://huggingface.co/collections/nvidia/nemotron-post-training-v3}{\texttt{Nemotron-Posttraining-v3} \includegraphics[height=0.9em]{assets/huggingface-color.png}} : A collection of post-training datasets for improving agentic, reasoning, and general model capabilities during SFT and RL.
\end{itemize}


We organize the remainder of the report into 4 sections: Pretraining~(\S\ref{sec:pretraining}), Post-training~(\S\ref{sec:posttraining}), Quantization~(\S\ref{sec:quantization}), and Inference~(\S\ref{sec:inference}).
\section{Pretraining}
\label{sec:pretraining}

In this section, we provide details on the pretraining of \ourmodel. Specifically, we detail the architecture, NVFP4 pretraining recipe, data, and hyperparameters used for \ourbasemodel. We also share details on the long context extension phase, model training instabilities, and benchmark accuracies.

\subsection{Model Architecture}
\label{sec:architecture}

\begin{figure}[!ht]
\centering
\includegraphics[width=0.9\linewidth]{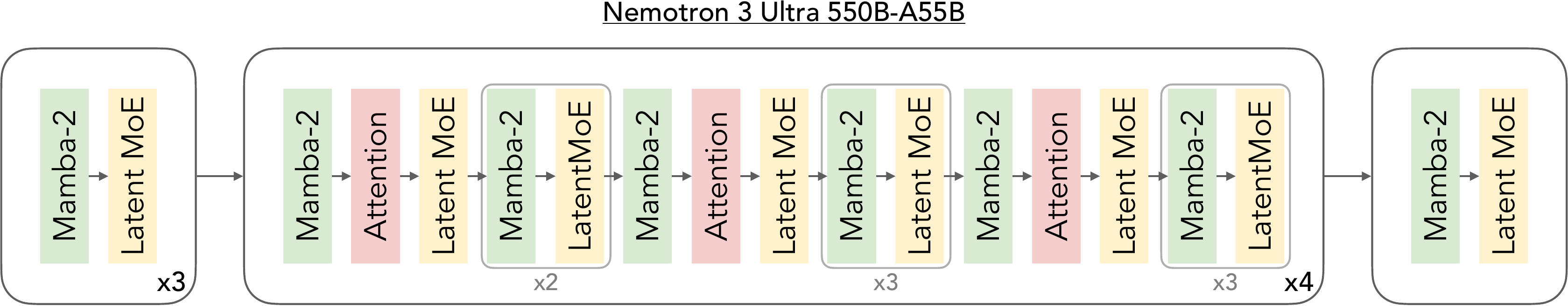}
\caption{\ourmodel layer pattern. Similar to Nemotron 3 Super, we use a hybrid Mamba-Attention architecture scaled sparsely using LatentMoE layers.}
\label{fig:layer-pattern}
\end{figure}

\ourmodel uses the same hybrid Mamba-Attention Mixture-of-Experts (MoE) architecture as Nemotron 3 Super~\citep{nvidia2026nemotron3superopen}, extended to 550B total parameters with 55B active parameters per token. As with Nemotron 3 Super, we leverage LatentMoE~\citep{latentmoe_tr} for MoE layers and native Multi-Token Prediction for inference acceleration with two heads during pre-training. Both MTP heads share the same parameters to enable robust autoregressive drafting as described in~\cite{nvidia2026nemotron3superopen} and consist of a single attention layer followed by a single MoE layer. The layer pattern and configuration for \ourmodel are shown in Figure~\ref{fig:layer-pattern} and Table~\ref{table:model_arch}, respectively.

\begin{table*}[!ht]\small\centering
\renewcommand{\arraystretch}{1.2}
\setlength{\tabcolsep}{10pt}
\begin{tabular}{l|c}
\toprule
\textbf{Configuration} & \textbf{\ourmodel} \\
\midrule
Total Layers & 108 \\
Model Dimension  & 8192 \\
\midrule
Q-Heads ($n_{q}$) & 64 \\
KV-Heads ($n_{kv}$) & 2 \\
Head Dimension & 128 \\
\midrule
Mamba State Dimension & 128 \\
Mamba Groups & 8  \\ 
Mamba Heads & 256 \\
Mamba Head Dimension & 64 \\ 
\midrule
Expert Hidden Dimension & 5120 \\
Shared Expert Intermediate Size & 10240 \\
Total Experts per Layer & 512 \\
Top-$k$ (Activated Experts) & 22 \\
\midrule
MoE Latent Size & 2048 \\
\midrule
MTP layers (shared weight) & 2 \\ 
\bottomrule
\end{tabular}
\caption{\ourmodel Architecture Dimensions.}
\label{table:model_arch}
\end{table*}

\subsection{NVFP4 Pretraining}

\begin{figure*}[!h]
    \centering
    \includegraphics[width=0.8\linewidth]{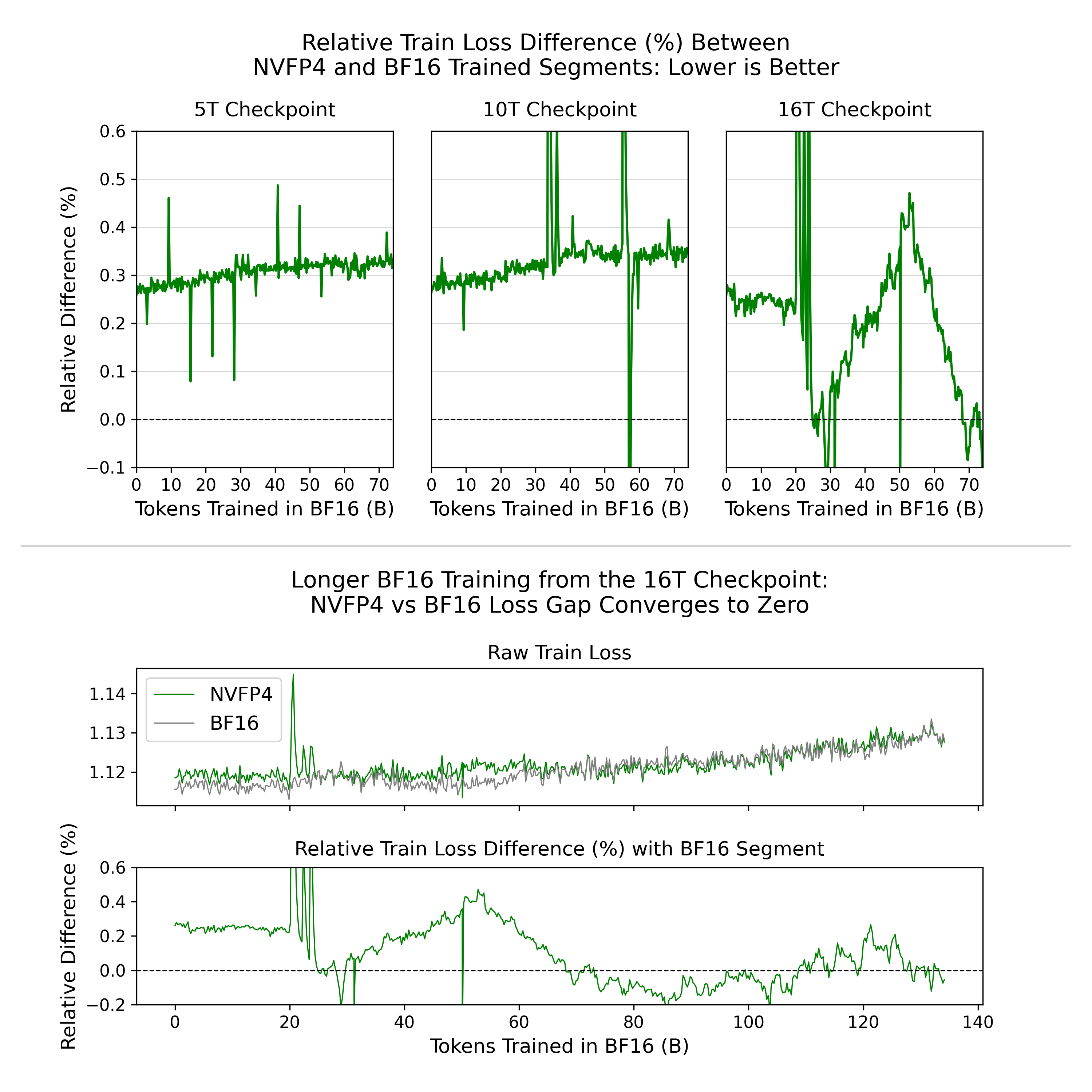}
    \caption{Top: Ablation studies switching all tensors to BF16 from 5T, 10T, 16T token checkpoints, respectively, shown as relative percent difference in train loss between NVFP4 and BF16. The starting loss gap for each ablation study was within 0.28\%. After 74B tokens of BF16 training, the loss gap increases to 0.33\% and 0.34\% from the 5T and 10T checkpoints, and decreases to 0.03\% from the 16T checkpoint. Bottom: Longer training in BF16 from the 16T token checkpoint. Raw train loss shows similar pattern of divergence for both BF16 and NVFP4 models. Relative difference in train loss (\%) converges toward zero during the training divergence. Switching all tensors to BF16 did not resolve the training divergence.}
    \label{fig:nvfp4_pretrain}
\end{figure*}

We trained \ourmodel using the same NVFP4 pretraining recipe as Nemotron 3 Super \cite{nvidia2026nemotron3superopen}, leveraging Transformer Engine's open-source cuBLAS NVFP4 GEMM kernels for fprop, dgrad, and wgrad. NVFP4 layers use the E2M1 datatype with two-dimensional block quantization on weights, Random Hadamard Transforms on inputs to wgrad, and stochastic rounding on gradients \cite{nvidia2025pretraininglargelanguagemodels}. We kept the final 15\% of the network (16 layers), Mamba output projections, latent projections, QKV and attention projections, MTP layers, and embedding layers in higher precision following \cite{nvidia2025nvidianemotron3efficient, nvidia2026nemotron3superopen}. To our knowledge, this is the largest-scale demonstration of stable and accurate NVFP4 training to date.

To monitor training health, we branched ablations from checkpoints at 5T, 10T, and 16T tokens, switched all tensors to BF16, and continued pretraining for 74B tokens. We tracked the relative train loss difference between BF16 segments and \ourmodel (NVFP4). As seen in prior work, switching all tensors to BF16 substantially recovers the high-precision loss, providing a proxy for high-precision training \cite{nvidia2025pretraininglargelanguagemodels}.  These three ablation studies on \ourmodel showed a relative train loss gap against the BF16 segments below 0.4\% on average (Figure \ref{fig:nvfp4_pretrain}, top), which is lower than NVFP4 vs. BF16 train loss gaps observed on smaller model variants \cite{nvidia2025nvidianemotron3efficient}. The relative train loss gap averaged over the first 5B tokens of BF16 training was 0.27\%, 0.28\%, and 0.25\% starting from the 5T, 10T, and 16T checkpoints, respectively. After 74B tokens of BF16 training, the relative train loss gap increases to 0.33\% and 0.34\% from the 5T and 10T checkpoints, and decreases to 0.03\% from the 16T checkpoint, averaged over the final 5B tokens. Switching all tensors to BF16 did not resolve the training divergence discussed in section \ref{sec:stability} (Figure \ref{fig:nvfp4_pretrain}, bottom).

\subsection{Pretraining Data}


In this section, we describe the new data that we added to pretraining since Nemotron 3 Super~\citep{nvidia2026nemotron3superopen}. We are releasing these new datasets on HuggingFace.\footnote{\href{https://huggingface.co/datasets/nvidia/Nemotron-Pretraining-Code-v3}{\texttt{Nemotron-Pretraining-Code-v3}}, \href{https://huggingface.co/datasets/nvidia/Nemotron-Pretraining-Legal-v1}{\texttt{Nemotron-Pretraining-Legal-v1}}, \href{https://huggingface.co/datasets/nvidia/Nemotron-Pretraining-Specialized-v1.2}{\texttt{Nemotron-Pretraining-Specialized-v1.2}}}


\subsubsection{Code refresh}

We refreshed our raw source code data from GitHub, adding 173B new tokens with a cut-off date of September 30, 2025.

\subsubsection{Nemotron-Pretraining-Multiple-Choice and Nemotron-Pretraining-Generative}

We generated large-scale, task-seeded synthetic Q\&A data from the training splits of many public datasets spanning a wide range of domains, including STEM, factual knowledge, commonsense reasoning, logical reasoning, math, code, reading comprehension, and multilingual QA.
Held-out test splits were not used for data generation. The source benchmark training examples were used as seeds to capture task structure, domain, difficulty, and answer format, while the generated examples were newly synthesized to preserve the underlying capability being tested rather than reproduce evaluation instances.

We organize the resulting data into two dataset families: Nemotron-Pretraining-Multiple-Choice, which contains synthetic questions with answer options and normalized correct answers, and Nemotron-Pretraining-Generative, which contains open-ended Q\&A examples with free-form answers. For both formats, we generate answer-enriched samples that include task-relevant knowledge, reasoning, or explanatory context when appropriate. We apply formatting checks, schema validation, deduplication, and task-specific filtering to improve data quality. These datasets are designed to promote cross-task capability transfer by exposing the model to diverse task formats, reasoning patterns, and knowledge domains during pretraining.

To validate the quality of this data, we conducted a 100B-token phase-3 continued-pretraining ablation on a Nemotron-family base checkpoint. Adding the benchmark-oriented synthetic data improved MMLU-Pro from 64.8 to 66.6, average code from 73.2 to 75.1, commonsense understanding from 72.9 to 74.5, and GPQA from 30.8 to 41.9, while average math remained stable (87.6 to 87.9).

\subsubsection{Nemotron-Pretraining-Fact-Seeking}

This dataset contains fact-seeking questions generated from Finewiki~\citep{penedo2025finewiki}. We generate the questions in two stages:
extracting informative, factual statements from Finewiki articles, and prompting Qwen3-30B-A3B-Instruct-2507 with each statement and its original context to generate either a short-answer or multiple-choice question.

To verify the usefulness of the data, we conducted an ablation study using an intermediate checkpoint from Nemotron 3 Nano pretraining. We injected the fact-seeking data during the final 100B tokens of training, improving accuracy on SimpleQA from 40.24 to 50.16. Since we converted SimpleQA questions into multiple-choice format for easier evaluation, these scores are not directly comparable to the original SimpleQA scores.

\subsubsection{Nemotron-Pretraining-Moral-Scenarios}

In the SFT data we previously released\footnote{\url{https://huggingface.co/datasets/nvidia/Nemotron-Pretraining-SFT-v1}}, we included multiple-choice questions about moral scenarios. These questions were constructed using situations and norms from Moral Stories~\citep{emelin-etal-2021-moral} and actions from Social Chemistry~\citep{forbes-etal-2020-social}. In this work, we sampled a subset of these examples and created a chain-of-thought version using Qwen3-235B-A22B-Thinking-2507.

\subsubsection{Nemotron-Pretraining-Legal}

We curated and generated a number of datasets targeting the legal domain as follows.

\begin{itemize}
    \item Datasets extracted from HTML files
    \begin{itemize}
        \item \textbf{Nemotron-Pretraining-Legal-California-Code-Of-Regulations}: California Code of Regulations\footnote{\url{https://govt.westlaw.com/calregs/Index}}, excluding Title 6 and Title 24.
        \item \textbf{Nemotron-Pretraining-Legal-NYCourts-Judicial-Ethics-Opinions}: New York Court Judicial Ethical Opinions\footnote{\url{https://www.nycourts.gov/ipjudicialethicsopinions}}.
        \item \textbf{Nemotron-Pretraining-Legal-eCFR}: Code of Federal Regulations\footnote{\url{https://www.ecfr.gov/}}.
    \end{itemize}

    \item LLM-cleaned datasets
    \begin{itemize}
        \item \textbf{Nemotron-Pretraining-Legal-Case-Law-Summary}: 5.4M summaries generated from a filtered version of Caselaw\footnote{\url{https://huggingface.co/datasets/common-pile/caselaw_access_project_filtered}} using Qwen3-235B-A22B-Instruct-2507.
    \end{itemize}

    \item Reformatted datasets
    \begin{itemize}
        \item \textbf{Nemotron-Pretraining-Legal-CaseHOLD}: We transformed the CaseHOLD dataset\footnote{\url{https://huggingface.co/datasets/casehold/casehold}} into a multiple-choice format.
        \item \textbf{Nemotron-Pretraining-Legal-Contract-NLI}: For each non-disclosure agreement in the ContractNLI dataset\footnote{\url{https://stanfordnlp.github.io/contract-nli/}}, we extracted the annotated hypotheses, answers, and evidence statements and appended them to the source document.
    \end{itemize}

    \item Synthetic datasets
    \begin{itemize}
        \item \textbf{Nemotron-Pretraining-Legal-Canadian-Case-Law-Outcome}: We identified passages from the CHRT, RPD, RAD, and RLLR subsets of the Canadian Case Law dataset\footnote{\url{https://huggingface.co/datasets/a2aj/canadian-case-law}} that clearly state the outcome of an appeal (allowed, dismissed, or other), as well as random passages that do not include the outcome using Qwen3-235B-A22B-Instruct-2507.
        \item \textbf{Nemotron-Pretraining-Legal-Definition-Classification}: From Caselaw, we extracted passages containing defining language as positive examples using Qwen3-235B-A22B-Instruct-2507 and randomly selected passages not containing defining language as negative examples. We use these passages to construct questions that classify whether a text from a judicial opinion defines a term.
        \item \textbf{Nemotron-Pretraining-Legal-Diversity-Jurisdiction}: This dataset contains questions that ask whether complete diversity exists between plaintiffs and defendants. The questions are generated from templates using random person names sampled from Nemotron Persona, along with states and causes of action sampled from two predefined lists. We also rephrase questions using Qwen3-235B-A22B-Instruct-2507 to increase diversity.
        \item \textbf{Nemotron-Pretraining-Legal-Function-Of-Decision}: We randomly sampled paragraphs from Caselaw documents and prompted Qwen3-235B-A22B-Instruct-2507 to classify its function into 7 pre-defined categories (facts, procedural history, issue, rule, analysis, conclusion, decree). We further balanced the number of examples for each category.
        \item \textbf{Nemotron-Pretraining-Legal-GlobalCit}: This dataset contains questions related to global nationality laws, converted from the GLOBALCIT dataset\footnote{\url{https://globalcit.eu/databases/globalcit-citizenship-law-dataset}} based on its codebook. We rephrased each question into three different version using Qwen3-235B-A22B-Instruct-2507.
        \item \textbf{Nemotron-Pretraining-Legal-LegalBench-CUAD-v2}: This dataset contains questions that ask whether a clause is a specific type of clause in a contract defined in the Contract Understanding Atticus Dataset (CUAD)\footnote{\url{https://www.atticusprojectai.org/cuad/}}. Using Qwen3-235B-A22B-Instruct-2507, we cleaned raw CUAD contracts shorter than 8k tokens, extracted the first qualifying clause of each type from each contract, and generated a negative example from each extracted clause. For some categories (\texttt{affiliate\_license\_licensor}, \texttt{affiliate\_license\_licensee}, \texttt{post\_termination\_services}, \texttt{exclusivity}, \texttt{effective \_date}, \texttt{non\_disparagement}, \texttt{unlimited\_all\_you\_can\_eat\_license}) where the identification accuracy is low, we composed longer prompts with detailed instructions based on the labeling handbook to extract qualifying clauses.
        \item \textbf{Nemotron-Pretraining-Legal-ToS-Clause-Understanding}: This dataset contains terms of service clause understanding questions. We generated a relevant legal question about each clause from the TOS Dataset\footnote{\url{https://huggingface.co/datasets/CodeHima/TOS_Dataset}} using Qwen3-235B-A22B-Instruct-2507.
        \item \textbf{Nemotron-Pretraining-Legal-ToSDR-QA}: This dataset contains Yes/No questions that address different sections or issues covered by each contract in the ToSDR Terms of Service Corpus\footnote{\url{https://www.kaggle.com/datasets/sonu1607/tosdr-terms-of-service-corpus}} using Qwen3-235B-A22B-Instruct-2507.
        \item \textbf{Nemotron-Pretraining-Legal-eCFR-QA}: This dataset contains DiverseQA-like data generated from the Code of Federal Regulations. We generated a variety of question from CFR excerpts and evaluate the correctness of each answer using Qwen3-235B-A22B-Instruct-2507.
    \end{itemize}

\end{itemize}

We conducted an ablation study using an intermediate checkpoint from Nemotron 3 Nano pretraining, an MoE with 30B total and 3B active parameters. Starting from a 14.9T token checkpoint, we trained for an additional 100B tokens using a phase 2 blend and evaluated models on over 100 subtasks in LegalBench. The experiment results show that these legal-specific datasets substantially improve our model’s accuracy on LegalBench tasks across multiple categories, boosting the average accuracy from 64.6 to 74.7.

\subsubsection{Data Mixture and Ordering}

\begin{figure}[!t]
    \centering
    \begin{subfigure}[b]{0.49\textwidth}
        \centering
        \includegraphics[width=\linewidth]{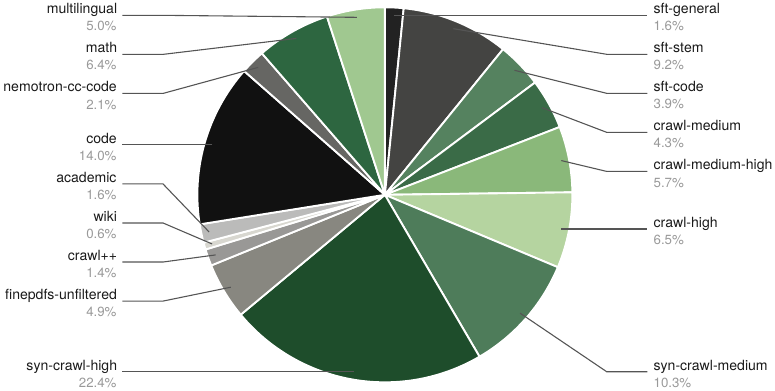}
        \caption{Data mixture of phase 1}
        \label{fig:data-mixture-phase1}
    \end{subfigure}
    \hfill
    \begin{subfigure}[b]{0.49\textwidth}
        \centering
        \includegraphics[width=\linewidth]{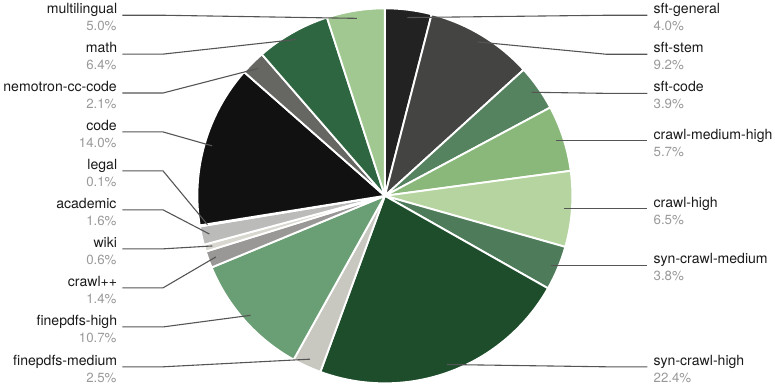}
        \caption{Data mixture of phase 2}
        \label{fig:data-mixture-phase2}
    \end{subfigure}
    \caption{The data mixtures for both pretraining phases. We design the phase 1 data mixture to have a bias for diversity and the phase 2 data mixture to have a bias for quality.}
    \label{fig:data-mixtures}
\end{figure}

The data mixtures used to train \ourmodel are an adaptation of the data mixtures used to train Nemotron 3 Super and Nano \citep{nvidia2025nemotron3nanoopen, nvidia2026nemotron3superopen}, and incorporate new and refreshed datasets. Following \citet{feng2024maximizedataspotentialenhancing}, we design our data mixtures to balance diversity and quality. We adopt the proposed two-phase curriculum and transition from a data mixture which biases dataset diversity (phase 1) to a data mixture which biases dataset quality (phase 2). This transition occurs after ${\sim}15$ trillion tokens corresponding to ${\sim}75\%$ of pretraining. We present the high-level breakdown for the phase 1 and phase 2 data mixtures in Figure~\ref{fig:data-mixtures}. More detail on quality estimation and dataset composition are available in \citet{feng2024maximizedataspotentialenhancing} and the Nemotron 3 Super and Nano reports \citep{nvidia2025nemotron3nanoopen,nvidia2026nemotron3superopen}.

The pretraining corpus spans 19 high level categories across both data mixtures. The largest component, accounting for ${\sim}49\%$ of phase 1 tokens and ${\sim}38\%$ of phase 2 tokens, consists of quality-filtered and synthetic web crawl data: crawl-medium, crawl-medium-high, crawl-high, syn-crawl-medium, and syn-crawl-high. Other categories include finepdfs \citep{kydlicek2025finepdfs} which we quality-filter and upweight for inclusion in phase 2, math data \citep{mahabadi2025nemotronccmath133billiontokenscalehigh,akter2025mindmathinformedsynthetic}, code data, Nemotron-CC-Code, Wikipedia, academic texts, legal data, multilingual data spanning 11 languages (Arabic, Chinese, French, German, Hebrew, Hindi, Italian, Japanese, Korean, Portuguese, Spanish), Crawl++, and synthetic SFT-style data. Crawl++ consists of OpenWebText, BigScience \citep{laurencon2023bigsciencerootscorpus16tb}, and Reddit datasets. The SFT-style data, subcategorized as sft-code, sft-stem, and sft-general, we include per \citet{akter2026frontloading} who demonstrate its effectiveness.

\subsection{Hyperparameters}
\label{sec:hyperparameters}
We follow the same training recipe and hyperparameters as Nemotron 3 Super~\citep{nvidia2026nemotron3superopen}, with a few adjustments: For \ourmodel we use a Warmup-Stable-Decay (WSD) learning rate schedule over a total horizon of 20 trillion tokens. We warmup the learning rate for 200 billion tokens to a peak value of $2.5 \times 10^{-4}$. For the final 5 trillion tokens we then decay the learning rate according to a minus-sqrt decay schedule to a minimum of $2.5 \times 10^{-6}$. 
As in Nemotron 3 Super, we used offline checkpoint merging for evaluation analysis~\citep{tian2025wsm} throughout pretraining with a sliding merge window size of 500B tokens at a checkpointing interval of 25B tokens weighted to emulate our learning rate decay schedule.
At the end of pretraining, final checkpoint selection was performed over a large set of checkpoint merges created using different merge settings: varying tokens seen, merge windows from 125B to 1T tokens, and using sequential, random, and reversed orderings. A 500B token merge window checkpoint that exhibited a balanced trade-off between knowledge, math, and code was selected for the long-context phase.
We use an MTP loss scaling factor of 0.1. All other hyperparameters remain the same as for Nemotron 3 Super. 

\subsection{Long-Context Extension}

Similar to Nemotron 3 Super \& Nano, we added a long-context phase (LC-Phase) at the end of pretraining. In the LC-Phase, we performed continuous pretraining (CPT) to equip the base model with long-context ability. We used a constant learning rate of $2.5*10^{-6}$. We used 32-way context parallelism, 8-way tensor parallelism, 128-way expert parallelism, and 2-way pipeline parallelism to train on GB200 GPUs.

Besides the long-context document QA data we used in Nemotron 3 Super \& Nano, we further added long-context SFT-style data into the blend. We did not use any RULER-style data in the blend. Overall, the long-context data was 46\% and Phase 2 data was 54\% in the blend. We performed CPT on 1,048,576 (1M) context length for 92\% of the iterations, while we trained on 4,096 (4K) for the remaining 8\% of the time in order to maintain the accuracy of the short benchmarks. Note that each iteration was trained with either 1M or 4K length and we did not mix sequence lengths within an iteration. Each iteration we constantly trained for 25,165,824 tokens. We only put math and code SFT-style data into the 4K iterations, since we found it worked best to maintain the short benchmark metrics while achieving strong long-context RULER scores. Eventually, the LC-Phase was trained for 33B tokens.

\subsection{Base Model Evaluations}
\label{subsec:base_model_evals}

\begin{table}[ht!]
\centering
\small
\setlength{\tabcolsep}{5pt}
\renewcommand{\arraystretch}{1.3}
\resizebox{\textwidth}{!}{%
\begin{tabular}{lr|ccccc}
\toprule
\textbf{Task} & \textbf{Metric} &
\multicolumn{1}{c}{\textbf{Nemotron-3-Ultra}} &
\multicolumn{1}{c}{\textbf{DeepSeek-V3.2}} &
\multicolumn{1}{c}{\textbf{Mistral-Large-3}} &
\multicolumn{1}{c}{\textbf{Kimi-K2}} &
\multicolumn{1}{c}{\textbf{GLM-4.5}} \\
& &
\multicolumn{1}{c}{\textbf{550B-A55B-Base}} &
\multicolumn{1}{c}{\textbf{Exp-Base}} &
\multicolumn{1}{c}{\textbf{675B-Base-2512}} &
\multicolumn{1}{c}{\textbf{Base}} &
\multicolumn{1}{c}{\textbf{Base}} \\
\midrule

\rowcolor{black!5}
\multicolumn{7}{l}{\textbf{General Knowledge}} \\
MMLU & \textit{5-shot, acc} & \textbf{89.08} & 87.82 & 87.35 & 87.60 & 86.50 \\
MMLU-Pro & \textit{5-shot, CoT EM} & \textbf{79.07} & 63.26 & 67.42 & 69.15 & 65.78 \\
AGIEval-En & \textit{3/5-shot, CoT EM} & \textbf{78.73} & 70.13 & 69.30 & 72.55 & 70.06 \\
GPQA & \textit{5-shot, CoT EM} & \textbf{50.00} & 31.82 & 34.85 & 43.43 & 34.85 \\
\midrule

\rowcolor{black!5}
\multicolumn{7}{l}{\textbf{Math}} \\
GSM8K & \textit{8-shot, CoT EM} & 88.10 & 84.38 & \textbf{91.21} & 91.05 & 85.37 \\
MATH & \textit{4-shot, EM} & \textbf{82.00} & 60.12 & 62.88 & 68.40 & 57.58 \\
\midrule

\rowcolor{black!5}
\multicolumn{7}{l}{\textbf{Code}} \\
HumanEval & \textit{sampled pass@1 n=32, EvalPlus sanitized} & \textbf{83.84} & 61.85 & 66.71 & 78.20 & 78.16 \\
MBPP-Sanitized & \textit{3-shot pass@1 n=32, EvalPlus sanitized} & \textbf{85.97} & 58.66 & 84.08 & 72.14 & 76.69 \\
\midrule

\rowcolor{black!5}
\multicolumn{7}{l}{\textbf{Commonsense Understanding}} \\
ARC-Challenge & \textit{25-shot, acc\_norm} & \textbf{97.35} & 95.22 & 97.27 & 95.82 & 96.59 \\
HellaSwag & \textit{10-shot, acc\_norm} & 90.51 & 89.44 & 88.88 & \textbf{90.92} & 90.17 \\
OpenBookQA & \textit{0-shot, acc\_norm} & 48.60 & 48.20 & \textbf{51.40} & 50.80 & 49.60 \\
PIQA & \textit{0-shot, acc\_norm} & 83.79 & 85.09 & 84.82 & \textbf{85.47} & 85.09 \\
WinoGrande & \textit{5-shot, acc} & 79.32 & 83.43 & 82.08 & 84.21 & \textbf{85.24} \\
\midrule

\rowcolor{black!5}
\multicolumn{7}{l}{\textbf{Reading Comprehension}} \\
RACE & \textit{0-shot, acc} & 92.15 & 93.21 & \textbf{93.30} & 91.96 & 92.15 \\
\midrule

\rowcolor{black!5}
\multicolumn{7}{l}{\textbf{Multilingual}} \\
MMLU Global Lite & \textit{5-shot, avg} & \textbf{90.13} & 85.59 & 87.34 & 85.63 & 85.81 \\
MGSM & \textit{8-shot, native CoT avg} & \textbf{87.73} & 82.33 & 82.93 & 85.20 & 81.27 \\
\midrule

\rowcolor{black!5}
\multicolumn{7}{l}{\textbf{Long Context}} \\
RULER 64K & \textit{0-shot} & \textbf{95.30} & 93.30 & 90.11 & 93.79 & 16.12 \\
RULER 128K & \textit{0-shot} & \textbf{92.49} & 91.88 & 55.77 & 88.61 & 0.00 \\
RULER 256K & \textit{0-shot} & \textbf{86.22} & -- & 35.50 & -- & -- \\
RULER 512K & \textit{0-shot} & \textbf{84.54} & -- & -- & -- & -- \\
RULER 1M & \textit{0-shot} & \textbf{76.83} & -- & -- & -- & -- \\
\bottomrule
\end{tabular}%
}
\caption[Base model comparison]{
    Comparison of \textbf{Nemotron-3-Ultra-550B-A55B-Base}, \textbf{deepseek-ai/DeepSeek-V3.2-Exp-Base}, \textbf{mistralai/Mistral-Large-3-675B-Base-2512}, \textbf{moonshotai/Kimi-K2-Base}, and \textbf{zai-org/GLM-4.5-Base}. Best available results are marked in bold.
}
\label{tab:base-model-comparison}
\end{table}

All evaluation results reported for \ourbasemodel were collected via Nemo Evaluator SDK\footnote{\url{https://github.com/NVIDIA-NeMo/Evaluator}} and NVIDIA's open source container of LM Evaluation Harness\footnote{\url{https://github.com/EleutherAI/lm-evaluation-harness}}, unless otherwise stated. For reproducibility purposes, more details on the evaluation settings can be found in the Nemo Evaluator SDK examples folder\footnote{\url{https://github.com/NVIDIA-NeMo/Evaluator/blob/main/examples/nemotron/nemotron-3-ultra}}.
The open source LM Evaluation Harness container packaged via NVIDIA's Nemo Evaluator SDK used for evaluations can be found here\footnote{\url{https://catalog.ngc.nvidia.com/orgs/nvidia/teams/eval-factory/containers/lm-evaluation-harness}}. This container is built on top of LM Evaluation Harness, with the following evaluation settings:
\begin{enumerate}

  \setlength{\itemsep}{0.4em}

    \item For general knowledge, we evaluate MMLU~\citep{hendrycks2021measuringmassivemultitasklanguage}, MMLU-Pro, AGIEval-En and GPQA. We report exact match or accuracy metrics according to the benchmark-specific evaluation protocol shown in Table~\ref{tab:base-model-comparison}. Missing entries indicate that the corresponding result was not available in the comparison set.

    \item For mathematical reasoning, we evaluate GSM8K~\citep{cobbe2021trainingverifierssolvemath} with 8-shot chain-of-thought exact match and Minerva Math with 4-shot exact match.

    \item For code tasks, we evaluate HumanEval~\citep{chen2021evaluatinglargelanguagemodels} and MBPP~\citep{austin2021programsynthesislargelanguage} using EvalPlus-sanitized variants~\citep{Liu_Is_Your_Code_2023}. We report sampled pass@1 estimated from 32 generations per prompt for HumanEval and MBPP-Sanitized where available.

    \item For commonsense reasoning, we report ARC-Challenge~\citep{Clark2018ThinkYH}, OpenBookQA~\citep{mihaylov2018suitarmorconductelectricity}, PIQA~\citep{bisk2019piqareasoningphysicalcommonsense}, HellaSwag~\citep{zellers2019hellaswagmachinereallyfinish}, and WinoGrande~\citep{sakaguchi2019winograndeadversarialwinogradschema} using the accuracy or normalized accuracy metrics shown in Table~\ref{tab:base-model-comparison}.

    \item For multilingual capability, we evaluate MGSM~\citep{shi2022languagemodelsmultilingualchainofthought} using 8-shot native chain-of-thought exact match and Global MMLU-Lite~\citep{singh2024globalmmluunderstandingaddressing} using 5-shot accuracy. The reported aggregate scores average the available language-specific results.

    \item For long context capability, we evaluate RULER~\citep{hsieh2024ruler} from 64K to 1M context length. Missing entries indicate that the corresponding result was not available in the comparison set.

\end{enumerate}

Evaluation results reported for \ourbasemodel with comparisons to deepseek-ai/DeepSeek-V3.2-Exp-Base, mistralai/Mistral-Large-3-675B-Base-2512, moonshotai/Kimi-K2-Base, and zai-org/GLM-4.5-Base are shown in Table~\ref{tab:base-model-comparison}.


\subsection{Model Stability}
\label{sec:stability}

\begin{figure*}[h]
    \centering
    \includegraphics[width=\linewidth]{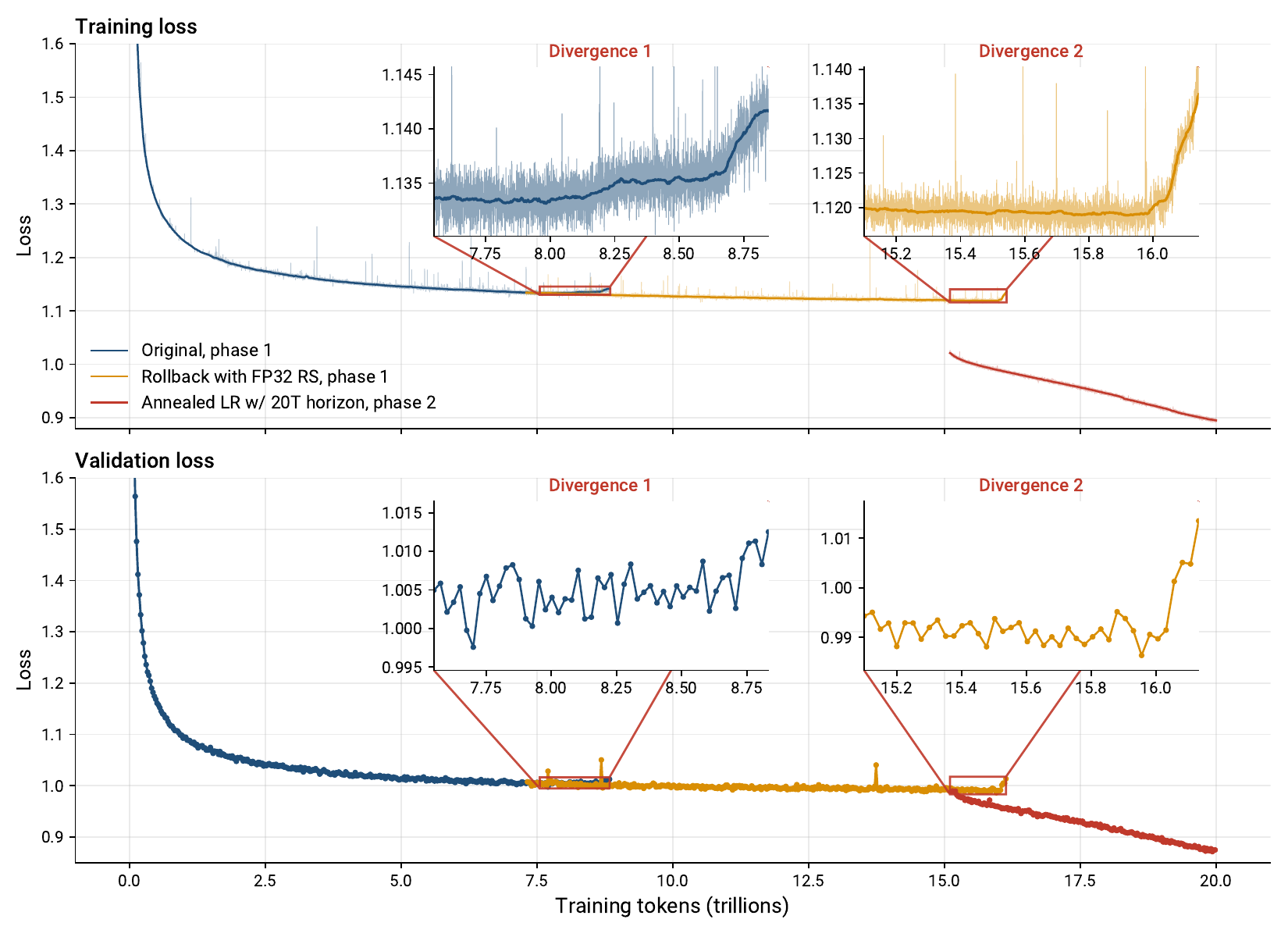}
    \caption{Training and validation loss versus number of tokens. Two separate runs with ``phase 1'' data (``Original'' and ``Rollback with FP32 RS'') are shown in different colors; both resulted in loss divergences. Figure insets show zoomed-in versions of each divergence. ``Rollback with FP32 RS'' was obtained by starting from a checkpoint in the original run before the first loss divergence and using the original FP32 gradient reduction recipe.} \label{fig:loss_vs_tokens_full}
\end{figure*}

During pretraining, we observed two instances of training divergence characterized by simultaneous increases in both the training cross-entropy loss and \texttt{wgrad} L2 norm.  These are shown in Figure~\ref{fig:loss_vs_tokens_full}.

\begin{figure*}[!ht]
    \centering
    \includegraphics[width=\linewidth]{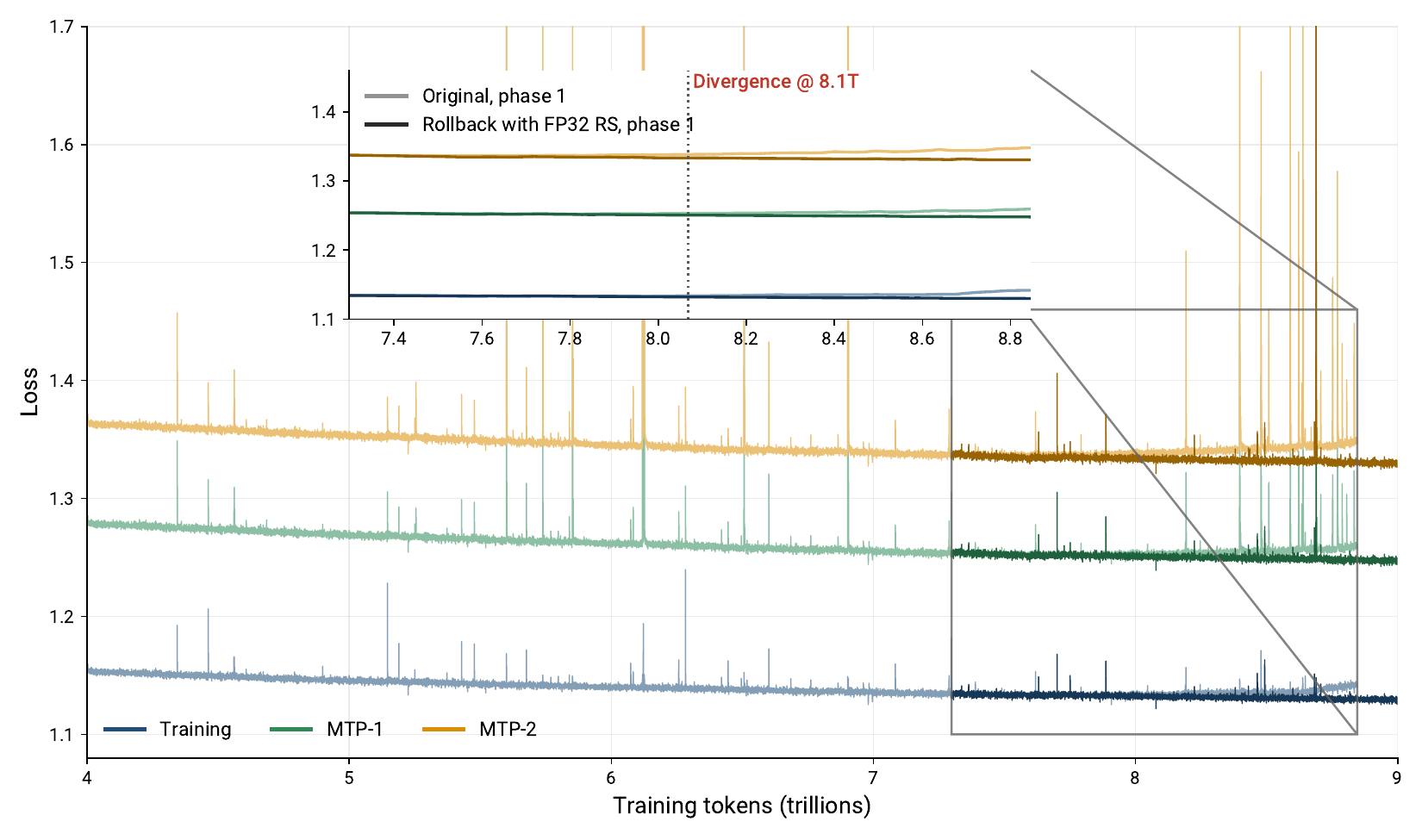}
    \caption{Training (cross-entropy loss for next token), MTP-1 and MTP-2 loss versus number of tokens around the region of the first divergence shown in Figure~\ref{fig:loss_vs_tokens_full}. MTP-2 loss diverges first with frequent large spikes.}
    \label{fig:loss_vs_tokens_divergence1_mtp}
\end{figure*}

\subsubsection*{Divergence 1: Local Gradient Accumulation Precision for Output Layer}

The first divergence, which occurred at around 8T tokens, was attributed to a reduction in local gradient accumulation precision for the output layer from FP32 to BF16 (in a bid to move data-parallel gradient reductions to BF16 over the wire as a throughput optimization). As mentioned in \S\ref{sec:architecture} and \S\ref{sec:hyperparameters}, \ourmodel uses 2 MTP blocks with a MTP loss scaling factor of 0.1 (0.05 for each MTP block); as a result, the MTP blocks' \texttt{wgrad} contribution to the shared output layer is essentially lost when using BF16, which has only 7 mantissa bits. Figure~\ref{fig:loss_vs_tokens_divergence1_mtp} shows MTP-2 loss started spiking / diverging before training (and validation) loss. Rolling back to an earlier checkpoint and moving back to the full FP32 gradient reduction recipe re-stabilized training (as shown in Figure~\ref{fig:loss_vs_tokens_full}).

\subsubsection*{Divergence 2: Undetermined}

\begin{figure*}[!ht]
    \centering
    \includegraphics[width=\linewidth]{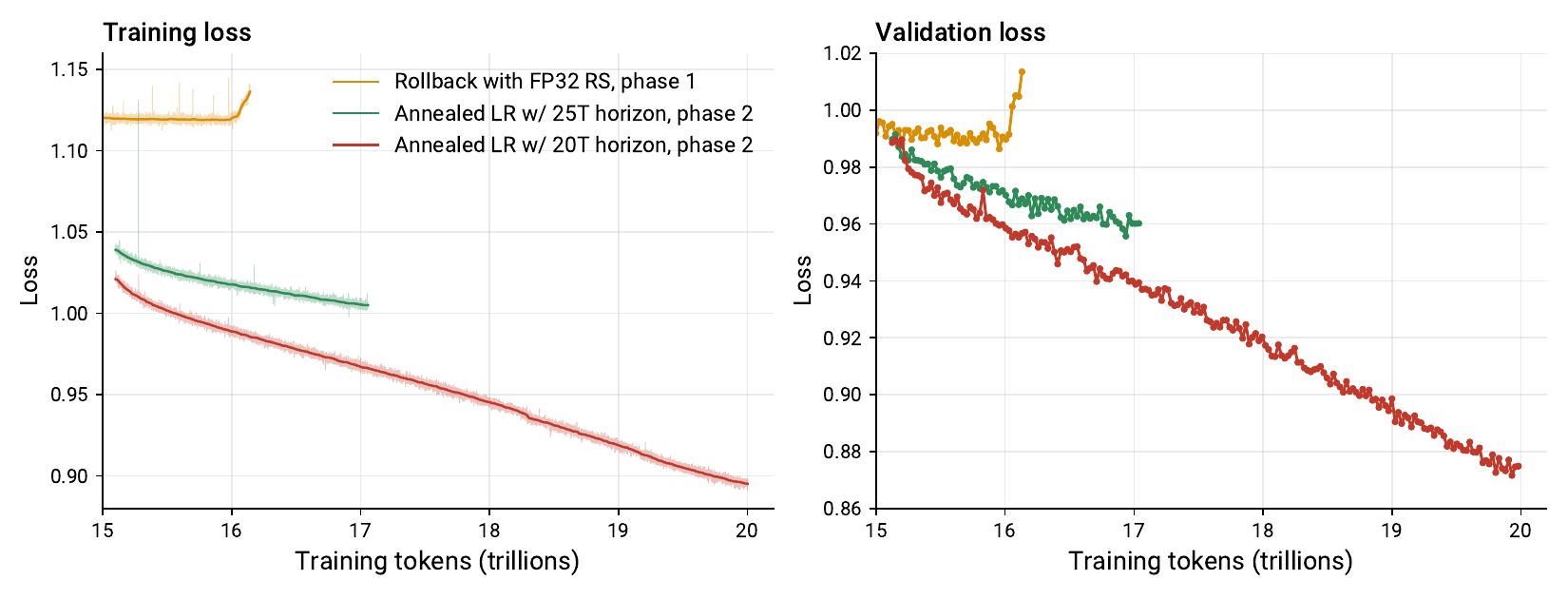}
    \caption{Training and validation loss around the second divergence (shown in Figure~\ref{fig:loss_vs_tokens_full}) for the original run that diverged (``Rollback with FP32 RS''), and two runs with early LR annealing targeting different token horizons after rewinding to a pre-divergence checkpoint (around 15T tokens).}
    \label{fig:loss_vs_tokens_divergence2}
\end{figure*}

\begin{figure}[!htp]
  \centering
  \begin{subfigure}[b]{0.49\columnwidth}
    \centering
    \includegraphics[width=\columnwidth]{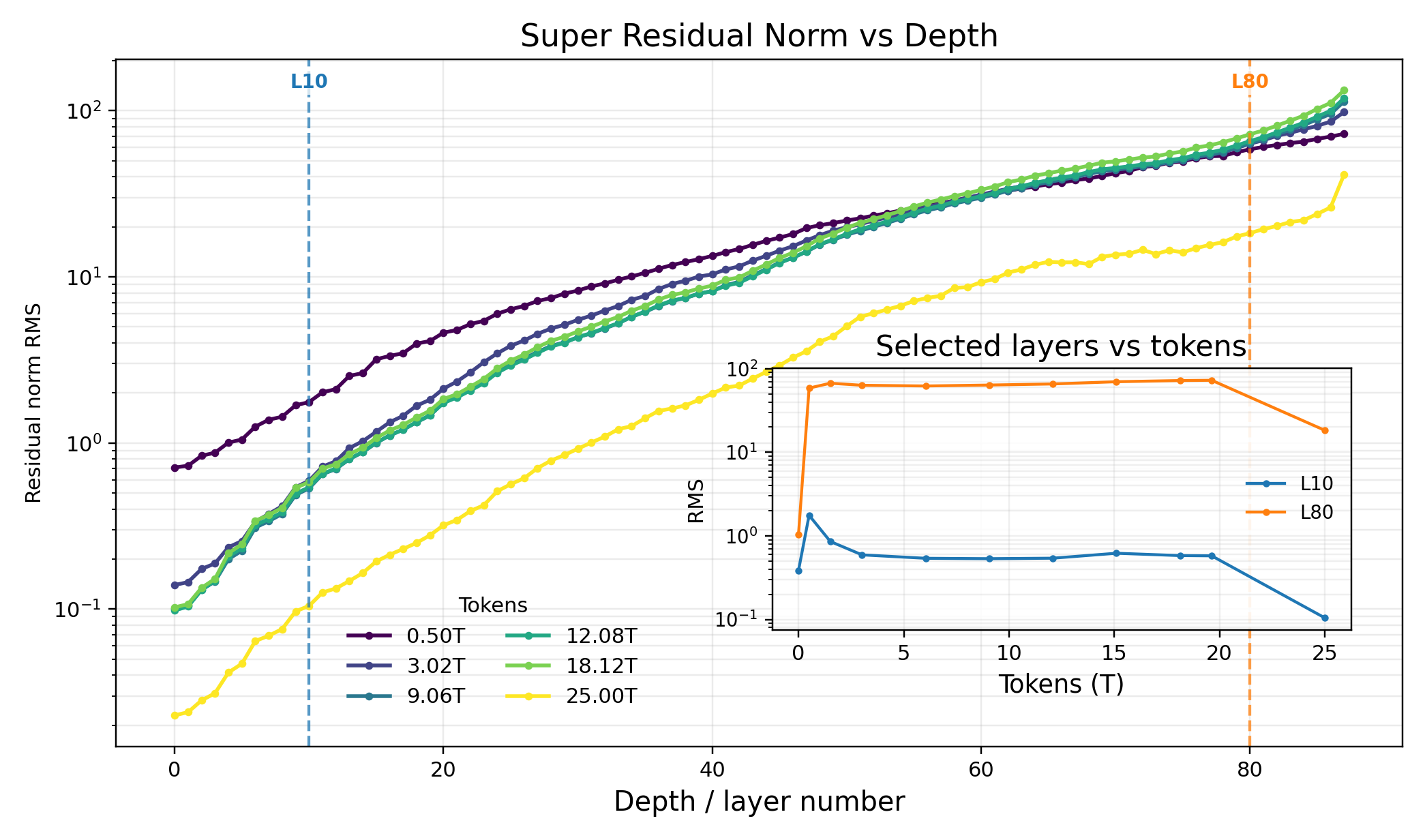}
    \caption{Nemotron-3-Super.}
    \label{fig:super_residual_norm}
  \end{subfigure}
  \hfill 
  \begin{subfigure}[b]{0.49\columnwidth}
    \centering
    \includegraphics[width=\columnwidth]{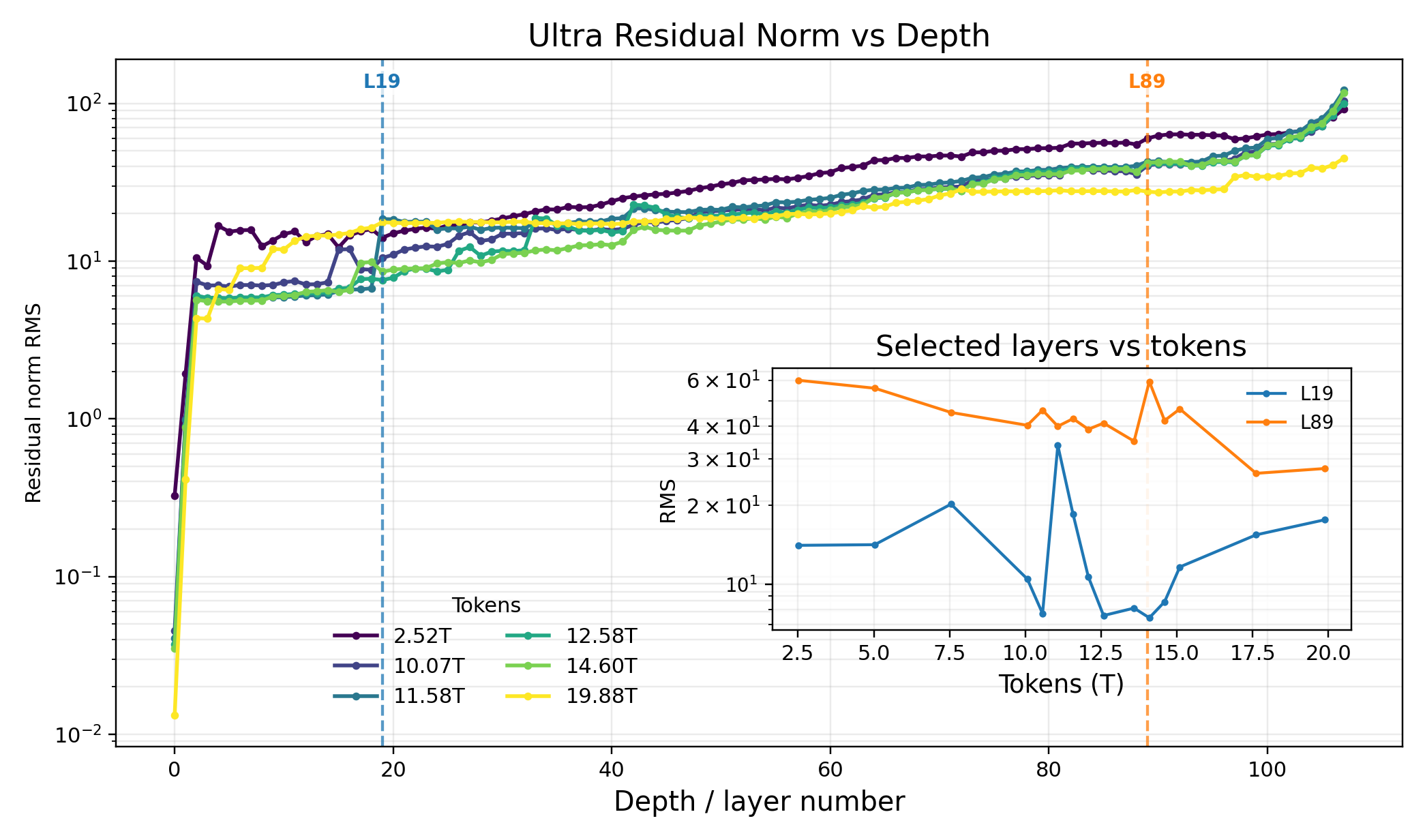}
    \caption{Nemotron-3-Ultra.}
    \label{fig:ultra_residual_norm}
  \end{subfigure}

  \vspace{1em} 

  \caption{Residual activation norm growth during training for Super and Ultra. The early layers for Super had increasing and then decreasing residual norms, with eventual stabilization. Later layers are characterized by norm growth. For Ultra, however, early layer residual norms started spiking about 7.5T tokens into training, with drastic spiking happening around 11T tokens, indicating increasing training instability.}
  \label{fig:super_ultra_residual_norms}
\end{figure}

For the second training divergence which occurred around $16$T tokens, we found through ablations that starting learning rate annealing (both a 5T and 10T decay) immediately after rolling back to the 15T token checkpoint mitigates divergence (Figure~\ref{fig:loss_vs_tokens_divergence2}). We eventually made the practical decision to cut the total pretraining token horizon down to 20T tokens.

To better understand the \emph{cause} of divergence, we studied the behavior of different model tensors across the full pretraining token horizon, and also between Super and Ultra. While we could not find a smoking gun that caused this instability, we found two interesting phenomena:

\paragraph{1. Imbalanced and Dead Experts.} As one possible proxy for pretraining health, the distribution of tokens across the available experts within the Mixture-of-Experts (MoE) layers can be continuously monitored. When a model begins to diverge or experience optimization difficulties, the routing mechanism often degrades, leading to severe token skew. In extreme cases, this results in ``dead experts'' that receive zero or near-zero tokens and effectively drop out of the learning process.
To quantify expert imbalance, we measure the MaxVio metric \citep{deepseekai2025deepseekv3technicalreport} which calculates the peak load on any single expert compared to the perfectly balanced mean. 

$\texttt{MaxVio} = \frac{\max_{1 \le i \le E} (T_i)}{\mu}$, where $E$ is the total number of experts, $T_i$ is the total number of tokens routed to expert $i$, and $\mu$ is the mean number of tokens per expert (calculated as $\frac{\sum T_i}{E}$).

We note that the maximum attainable MaxVio is $\text{MaxVio}_{\text{max}} = \frac{E}{k}$. For \ourmodel and Super, this gives us $\text{MaxVio}_{\text{max}} = 23.27$, while for Nano, it is $21.33$. 

We calculated MaxVio on 20 iterations worth of tokens (from both training and validation datasets) per checkpoint, for a total of 500M tokens. Across Nemotron-3 Nano, Super and Ultra, we observed that MaxVio on the training data was always lower than on the validation data over the course of pretraining. For Nano, the value was generally $\approx 1.3$ for training and $\approx 5$ for validation; for Super, it was $\approx 2$ and $\approx 6$, respectively. For Ultra, routing started balanced, with the median (across layers) MaxVio being $1.2$ and the maximum being $4.8$ (first MoE layer). As training progressed, expert routing became increasingly unbalanced; the median layer's MaxVio stayed around $1.2$ but the maximum kept increasing to $\approx 12$ by 12T tokens (again first layer). Although not causal by itself, MaxVio seems  correlated with training instability.

\paragraph{2. Imbalanced Residual Stream Activation Norms.} We found residual norms differed by 3 orders of magnitude across model depth for Super and 4 orders of magnitude for Ultra. Additionally, the dynamics of residual norms in \ourmodel were qualitatively different from what was observed during Nano and Super pretraining, as shown in Figure~\ref{fig:super_ultra_residual_norms}. For those models, residual norms in the early layers would increase, and then decrease and stabilize. Later layers had their residual norms slowly increase during training. For Ultra, residual norms initially followed this pattern, but norms in the early layers started rising around 7.5T pretraining tokens, with large residual norm spikes happening around 11T tokens, indicating poor signal propagation.

\clearpage

\clearpage
\section{Post-Training}
\label{sec:posttraining}

\begin{figure}[h]
    \centering
    \includegraphics[width=0.9\linewidth]{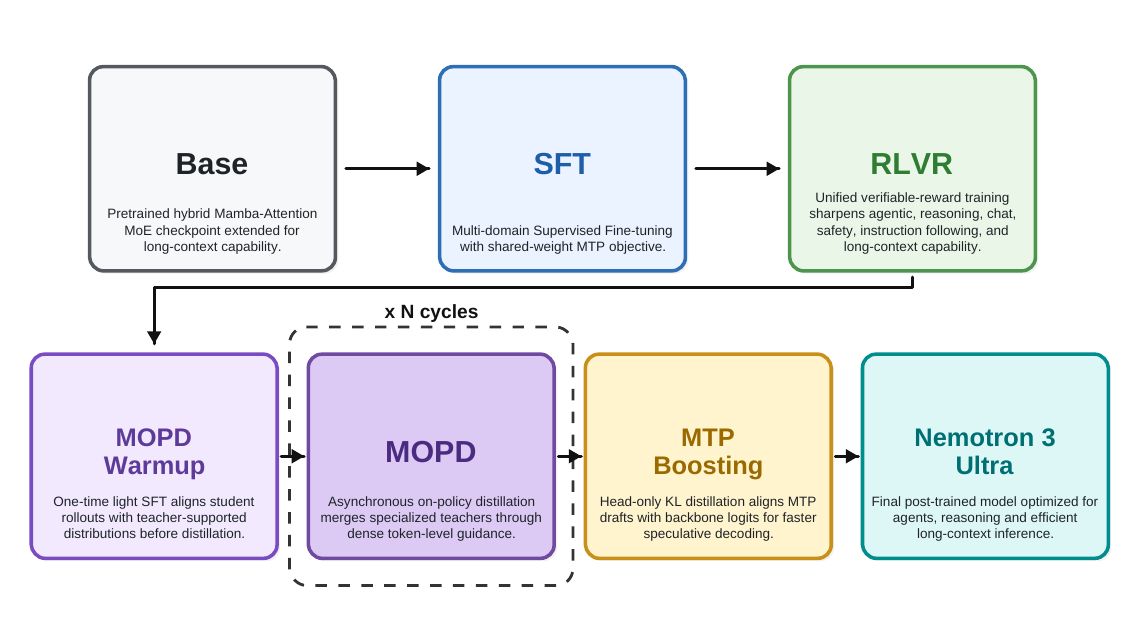}
    \vspace{-0.5em}
    \caption{Overview of the post-training pipeline for \ourmodel.}
    \label{fig:posttraining_stages}
    \vspace{-0.5em}
\end{figure}

Given the pretrained model, we conduct a post-training phase that has been substantially redesigned from the pipeline used for Nemotron 3 Super~\citep{nvidia2026nemotron3superopen}. As illustrated in Figure~\ref{fig:posttraining_stages}, the pipeline starts with a general Supervised Fine-tuning stage. Instead of relying solely on consecutive reinforcement learning stages, we augment the pipeline with Multi-teacher On-Policy Distillation (MOPD)~\citep{yang2026nemotron,lu2025onpolicydistillation,xiao2026mimo}, enabling both broad capability acquisition and targeted specialization. The remainder of this section is organized around the MOPD process: we first describe how the student model is prepared through SFT, RLVR, and MOPD warmup, then present the training of specialized teacher models, and finally detail the iterative MOPD optimization and MTP Boosting procedures.

\subsection{Supervised Fine Tuning}
\label{subsec:SFT}

To train the student model, we start from the pre-trained base model and perform Supervised Fine-Tuning (SFT) in two stages, following \supermodel~\citep{nvidia2026nemotron3superopen}. In Stage~1, we train on packed sequences of length $294{,}912$ tokens with global batch size $64$ for $204{,}800$ samples, using a cosine learning-rate schedule with peak learning rate $1.5 \times 10^{-5}$, minimum learning rate $1 \times 10^{-6}$, and $9{,}600$ warmup samples. In Stage~2, we extend packed sequence to $515{,}000$ tokens, augmenting the mixture with additional long-context data up to $512$K tokens. We train with global batch size $64$ for $19{,}200$ samples, using the same learning-rate schedule with peak $1 \times 10^{-5}$ to minimum $2 \times 10^{-6}$ and $6{,}400$ warmup samples. 
As in pre-training, we retain the shared-weight MTP objective during SFT, using two MTP layers with a per-token auxiliary-loss scaling factor of $0.1$.

\subsubsection{Data}
\paragraph{Long Context.}
We prepare 512K long-context SFT data based on the synthetic data pipeline following~\citep{nvidia2026nemotron3superopen}. 
Our data is aimed at improving long-context abilities including, but not limited to multi-document reasoning, sequential scanning, and querying synthetic tables.
\paragraph{Efficiency and Control.}
The SFT data include two components for reasoning efficiency and control. The first is training samples
generated by GPT-OSS-120B in its medium-effort mode on prompts of math reasoning, STEM question answering and instruction following. These SFT data initiate Ultra's medium-effort mode which is later optimized during the RLVR stage. The second component is training samples where the reasoning traces are truncated to random reasoning budgets while the responses remain the same. This is similar to Nemotron 3 Nano and Super with one design change: the \texttt{</think>} tokens in the truncated samples are masked from SFT training loss.

\paragraph{Safety.}
To instill robust safety behaviors, first, we retain the $45$K safety data blend curated in \supermodel~\citep{nvidia2026nemotron3superopen} with prompts curated from diverse sources and responses synthetically generated conditioned on a response policy mapped to each prompt. Our two-stage response and reasoning generation framework enables deliberate reflection on safety guidelines in the reasoning traces and ensures that the final responses are consistent, policy-compliant, and contextually appropriate. Further, to curate a multilingual set for \ourmodel training, we translated the \supermodel safety data blend into six languages — German, Spanish, French, Japanese, Italian, and Chinese - using NeMo Skills' chunked translation capability, ensuring sentence-by-sentence parity, and paired with NVIDIA Riva Translate 4B v1.1 as the translation backbone. 

To improve translation quality, each translated example was back-translated into English and compared against the original English prompt-response pair. Examples with semantic similarity below 0.8 were filtered out, removing approximately $\approx10 - 15\%$ of examples per translated language. The highest- and lowest-scoring translations were manually spot-checked to verify that this filter was removing examples with clear translation failures or structural issues. After filtering, we used stratified sampling to create a balanced dataset across the six translated languages. The final safety blend contains about $\approx135$K samples: $\approx45$K in English and $\approx15$K from each of the translated languages.

\paragraph{Search Capabilities.}
During SFT, we expose the model to search data spanning different difficulty levels and tool-use harnesses to improve generalization.

First, we retain the search trajectories from the \supermodel~\citep{nvidia2026nemotron3superopen} dataset, whose seed prompts are grounded in the Wikidata knowledge graph. Specifically, the dataset selects well-connected hub entities and performs 4--8 hop random walks over factual relations to construct multi-hop search prompts. These prompts are then solved using Tavily Search Engine (\url{https://www.tavily.com/}), with MiniMax 2.1 serving as the teacher model.

In addition, we include a new search dataset: a commercially cleared subset of OpenResearcher~\citep{li2026openresearcher}, a public SFT dataset designed for long-horizon research agents. OpenResearcher synthesizes over $97$K trajectories using \texttt{gpt-oss-120b}~\citep{agarwal2025gpt} as the teacher model in a fully offline browser environment. The environment is backed by a local search index over $15$M FineWeb documents~\citep{penedo2024fineweb} augmented with bootstrapped evidence documents, and exposes three structured browser tools: \texttt{search} for retrieving candidate pages, \texttt{open} for inspecting full document contents, and \texttt{find} for locating exact textual evidence within an opened document. The resulting trajectories capture long-horizon reasoning-action-observation loops in which the model iteratively decomposes research questions, gathers and inspects sources, localizes supporting evidence, and synthesizes grounded final answers. For \ourmodel, we did not regenerate the OpenResearcher data; instead, we curated the commercial-OK portion by removing examples whose source licensing was not cleared for commercial use. This filtering yields approximately $21.7$K SFT trajectories while preserving the original browser-tool interaction format for training open-ended research and evidence-grounded search behavior.

Finally, we work with data vendors to curate particularly challenging samples that require 50--100 searches, and collect SFT trajectories in our BrowseComp harness, described in Appendix~\ref{appendix:eval}. For these trajectories, we use MiniMax 2.5 and GLM 5.1 as teacher models.

\paragraph{Terminal-Use Capabilities.}
To develop strong terminal-use capabilities, we constructed a large-scale dataset of synthetic agentic trajectories covering a broad range of terminal tasks, including software engineering, data processing, file operations, and scientific computing. Seed instructions were sourced from a combination of publicly available datasets: OpenCodeReasoning \citep{nvidia2025opencodeReasoning}, OpenMathReasoning \citep{nvidia2025openmathReasoning}, SWE-bench \citep{jimenez2024swebenchlanguagemodelsresolve}, SWE-Fixer-Train-110K \citep{swefixer2025}, SWE-rebench \citep{badertdinov2025swerebenchautomatedpipelinetask}, and SWE-smith \citep{pan2025swesmith}. Tasks were assembled from two complementary sources: one portion was adapted from existing math and coding SFT data in Nemotron-Cascade, reformatted to align with the terminal-use environment, while the other was synthetically generated using DeepSeek-V3.2 \citep{deepseekai2025deepseekv3technicalreport} to cover a wider variety of terminal scenarios not well-represented in existing benchmarks. For trajectory generation, DeepSeek-V3.2 was used as the acting agent within the Terminus-2 agent provided by the Harbor framework \citep{harbor2025}. Each trajectory unfolds over multiple episodes in which the model receives an initial task prompt and subsequently interacts with a live terminal environment, issuing commands and incorporating execution feedback to make iterative progress toward task completion. This multi-step, environment-grounded collection process encourages the model to learn practical tool use, adaptive planning, and error recovery behaviors that are characteristic of effective terminal operation. The final dataset comprises approximately 370K multi-turn conversations, consisting of a mixture of reasoning and non-reasoning trajectories.

\paragraph{Conversational Tool Use Capabilities.}
We scale conversational tool-use data via a fully
synthetic, six-stage generation pipeline. This includes User and Environment simulation. This pipeline is similar to what is described in Nemotron-3 Super. 

\paragraph{Software Issue Resolution.}
To distill robust problem solving capabilities, we curated a diverse dataset of synthetic agent trajectories focused on resolving real-world GitHub issues. To this end, we generated synthetic trajectories by leveraging two distinct modeling paradigms: the reasoning-based (thinking) Minimax-M2.5 \citep{minimax2026m25} and the instruction-based (non-thinking) Qwen3-Coder-480B-A35B-Instruct \citep{yang2025qwen3technicalreport} for issue resolution. The underlying issue statements were curated from a diverse set of publicly available datasets, including SWE-Gym \citep{pan2025trainingsoftwareengineeringagents}, R2E-Gym \citep{jain2025r2egymproceduralenvironmentshybrid}, SWE-rebench \citep{badertdinov2025swerebenchautomatedpipelinetask}, and SWE-rebench-V2 \citep{badertdinov2026swe}. These trajectories were captured using the OpenHands \citep{wang2025openhandsopenplatformai}, SWE-agent \citep{yang2024sweagent}, and Mini-SWE-agent and Opencode harnesses. Raw agent rollouts contain many patterns that, while completing the task, would teach undesirable behaviors if used directly as SFT data, so we filter the rollout pool with a per-trajectory heuristic analyzer evaluates a fixed set of signals to make an include/exclude decision: submission
  integrity (trajectory must end with a valid submission action), disallowed git operations (any use of push, pull, fetch, clone, cherry-pick, reflog, fsck, remote, or ls-remote), edit-test loop anti-patterns (the model thrashes between edits and test runs without converging — either reverting its own edits or repeating the edit → test → edit cycle indefinitely), lost-in-exploration (the model spends most of its turns reading/searching the repo but rarely edits, indicating failure to localize and commit to a fix), tool-call hygiene (high rate of malformed tool calls), debug-artifact detection in the final patch (print(, pdb, breakpoint()), and a verification check that flags trajectories that edit but never run tests. This multi-model, multi-framework approach ensures the resulting dataset promotes generalization across both diverse problem-solving modes and varied agentic environments. 

\paragraph{Math / Proof Data.}
We use the Nemotron-Cascade-2~\citep{yang2026nemotron} math data, which sources problems primarily from Nemotron-Cascade~\citep{wang2025nemotroncascadescalingcascadedreinforcement,liu2025acereason} and Nemotron-Math-V2~\citep{du2025nemotronmath}.
For the non-proof math data, we collect 1.8M tool-calling samples and 1.9M non-tool samples, with responses generated by DeepSeek-V3.2 and DeepSeek-V3.2-Speciale, respectively.
For the mathematical natural language proof data, we source proof problems from the AOPS split of Nemotron-Math-Proofs-v1~\citep{du2025nemotronmath} and generate responses from DeepSeek-V3.2-Speciale for capabilities including proof generation, proof verification, and proof refinement.

\paragraph{Science.}
We prepare our science SFT data following the Nemotron Nano recipe~\citep{nvidia2025nemotron3nanoopen}, which combines synthetic, real-world, and document-derived seed data spanning physics, chemistry, and biology. Prompts and formats are diversified using NeMo Data Designer~\citep{nemo-data-designer}, and the resulting samples are filtered with an LLM judge for format compliance and reasoning quality. In addition to the Nano recipe, we generate web-search and web-search-with-Python reasoning traces using DeepSeek-V3.2~\citep{deepseekai2024deepseekv32}. For web-search traces, the model is provided access to the Tavily search engine (\url{https://www.tavily.com/}); for web-search-with-Python traces, the model additionally has access to a Python execution environment.

\paragraph{Chat.}
We create multi-turn chat SFT data by starting from seed prompts drawn from open conversational datasets such as LMArena \citep{chiang2024chatbot} and WildChat \citep{li2024wildchat}. For each prompt, we sample multiple candidate responses from GLM-5 \citep{glm5team2026glm5vibecodingagentic} and use Nemotron-GenRM \citep{Wang2025HelpSteer3Preference} to select the highest-quality response for the current turn. To extend the data into multi-turn conversations, we simulate the user with the same LLM under controlled prompting. The simulated user is guided by hand-crafted conversation strategies, such as building on prior content, asking for clarification, challenging assumptions, reframing the task, or applying the answer to a new context, in order to produce diverse and realistic dialogue trajectories. To build multi-turn robustness, we construct examples in which earlier assistant turns may contain suboptimal responses rather than the best-ranked candidates. The model is trained only on the final assistant response in SFT to produce high-quality current-turn outputs while remaining robust to imperfect previous turns.

\paragraph{Code.}
We collect our coding problems from modern competitive programming platforms such as Codeforces, AtCoder, AIZU, and CodeChef. Following \citet{yang2026nemotron}, we applied strict deduplication and aggressive filtering to improve the data quality and balance the problem difficulties. We choose GPT-OSS-120B~\citep{agarwal2025gpt} as our teacher model with the consideration of their strong reasoning ability and good verbosity, and apply the rejection sampling on those reasoning traces. This pipeline yields a final dataset comprising 1.2M Python reasoning traces, 1.0M C++14 reasoning traces, and 1.3M Python tool-calling reasoning traces for competitive coding.

\paragraph{CUDA.}
We constructed a large-scale synthetic CUDA dataset comprising approximately 100K samples for kernel generation, repair, and optimization. The dataset was built using an LLM-based synthetic data generation pipeline with DeepSeek-R1 and GPT-OSS-120B. Seed questions were sourced from popular open-source libraries, NVIDIA library API surfaces, and BackendBench~\citep{saroufimbackendbench}. The construction was motivated by the need for CUDA-specialized training data that reflects real-world GPU-kernel programming and benchmarking challenges, as highlighted by recent CUDA benchmark efforts such as SOL-ExecBench~\citep{lin2026solexecbenchspeedoflightbenchmarkingrealworld}. These seeds were used to generate two types of samples: \emph{PyTorch-reference-to-CUDA-kernel} samples and \emph{natural-language-specification-to-CUDA-kernel} samples, with each sample accompanied by reasoning. For each seed item, multiple candidate kernels were produced through LLM-based synthetic generation and rejection sampling. Candidate kernels were validated in an internal CUDA evaluation environment using compilation checks, numerical correctness tests, and runtime benchmarking. Invalid, non-compiling, or incorrect candidates were rejected. Among the remaining valid candidates, the best-performing kernel according to benchmarked runtime was retained. In addition, we collected traces from an internal CUDA agent to produce repair and optimization data. The repair samples contain a PyTorch reference, a faulty CUDA C++ kernel, the corresponding error message, and a corrected CUDA C++ kernel. The optimization samples contain a PyTorch reference, a slow CUDA C++ kernel, an Nsight Compute log, and an optimized CUDA C++ kernel. Beyond direct CUDA C++ kernel generation, we also generated CUDA-X library  data using publicly available documentation and official code samples. Following the same formulation as the CUDA-C data, we generated PyTorch references, aligned natural-language specifications, corresponding CUDA-X library implementations, and reasoning. The covered libraries include Thrust, CUB, cuBLAS, cuDNN, cuSPARSE, cuRAND, and cuSOLVER.

\paragraph{RTL.}
We use the ACE-RTL~\citep{deng2026ace} training data, which covers three major RTL task categories: specification-to-RTL generation, code editing, and code debugging. ACE-RTL builds on the seed RTL corpus from ScaleRTL~\citep{deng2025scalertl}, where seed designs are collected from license-checked open-source RTL repositories and processed through deduplication, filtering, and syntax validation before being used for data generation. For specification-to-RTL tasks, DeepSeek-R1 and GPT-OSS-120B are used to synthesize natural-language specifications paired with the corresponding golden RTL implementations derived from the seed designs. For editing and debugging tasks, the original seed RTL is treated as the golden implementation, while simplified variants with missing functionality or buggy variants with realistic injected design errors are generated as inputs; the corresponding specifications describe either the required feature extensions or diagnostic information needed to recover the golden RTL. After further filtering through syntax checks, benchmark decontamination, and semantic-alignment evaluation using human-defined rubrics, the final dataset contains around 1.2M training samples.

\paragraph{Multilingual.}
Our multilingual post-training data is a mixture of sentence-level parallel corpora and synthetic translations of English SFT examples in math, code, and science domains.
For the synthetic data, driven by data quality issues identified in the previous line-by-line translation pipeline, we introduced a new end-to-end translation pipeline that takes the full JSON object as input and generates the full JSON object as output.
Because this pipeline requires strong long-context capabilities from the translation model, we transitioned to \texttt{DeepSeek-V3-0324}.
After translation, we first perform heuristic format checking to ensure all outputs conform to the specified JSON format, followed by the same data filtration steps and light-weight post-editing step introduced in the Nemotron Super-V3 recipe~\citep{nvidia2026nemotron3superopen}.
Ablation studies on Japanese data showed significant improvements measured by MMLU-ProX~\citep{xuan2025mmluprox}, confirming quality improvements over the line-by-line pipeline.
For Nemotron Ultra-V3, we synthesized data for Hindi, Japanese, Korean, and Brazilian Portuguese using this pipeline, while reusing existing multilingual synthetic data from Super-V3 for the remaining languages.

\subsubsection{Data Packing}
To train efficiently on a large and diverse collection of datasets, we adopt a length-aware best-fit packing strategy~\citep{ding2024fewer}, which packs multiple conversations into sequences up to a maximum context length. 
Our packer is highly memory-efficient: it reads and interleaves all source files in a round-robin fashion, maintains only a fixed-size pool of open sequences in memory, and retires a sequence once its residual capacity falls below a small tolerance. 
Each incoming conversation is assigned to the partially filled sequence whose remaining capacity it most tightly fits, following a best-fit rule that minimizes padding overhead. 
We neither truncate nor split conversations, preserving complete context and reducing hallucinations. 
Additionally, we enforce an in-pack deduplication constraint to prevent identical prompts from co-occurring within the same sequence. 
After packing, we perform a final shuffle over all completed packs. 
Overall, our implementation ensures that each packed sequence draws from a broad, well-mixed cross-section of the data distribution rather than clustering examples from any single source. 
This thorough mixing is essential at scale, where naive concatenation of per-source shards would otherwise induce strong distributional locality and degrade optimization stability.
\subsection{Reinforcement Learning}
\label{subsec:RL}

To improve upon SFT model, we conduct a unified RLVR (Reinforcement Learning with Verifiable Reward) training stage spanning all available environments, targeting terminal usage, office and productivity workflows, software engineering, search, general tool-calling, math, code, STEM, safety, chat, instruction following, long-context QA, inductive and transductive reasoning, structured outputs, and general model usability. For harness-based environments, we construct training data using a diverse collection of harness implementations and interaction formats, improving robustness to variations in execution settings and reducing overfitting to any particular harness design.

We refresh the RL training data using recent collections and perform reward profiling prior to training. For data mixture and curriculum construction, we adopt the Gaussian-based approach introduced in~\cite{nvidia2025nemotron3nanoopen}. Our training procedure largely follows the asynchronous GRPO algorithm with the stability optimizations proposed in~\cite{nvidia2026nemotron3superopen}, while incorporating several improvements to the training infrastructure, detailed in Section~\ref{subsec:infrastructure}. To support training across a large and diverse set of environments, we use a global batch size of 8192, with each sample generating 16 rollouts. Training begins with a maximum generation length of 48K tokens, which is later increased to 64K tokens. 
\subsection{MOPD}
\label{subsec:MOPD}

Mixed-environment RLVR provides broad capability improvements across a wide range of domains. However, as the number of environments continues to grow, each domain contributes only a relatively small number of samples to any given training batch, diluting the per-domain learning signal and making it increasingly difficult to balance training across domains. To fully unlock performance and push the frontier in each capability area, we train more than ten specialized teacher models, each optimized through its own domain-specific training pipeline.

\begin{figure}[h]
    \centering
    \includegraphics[width=\linewidth]{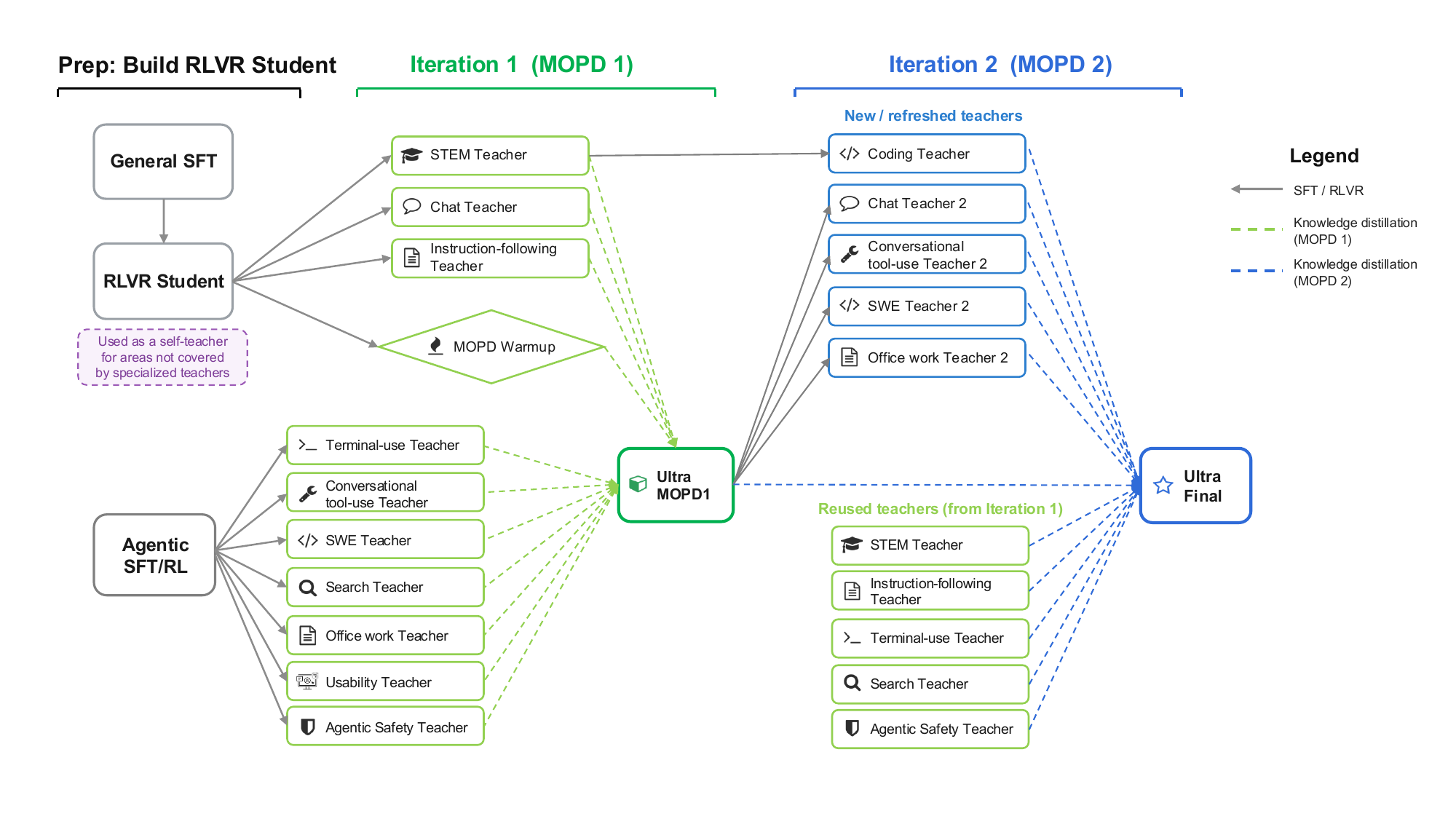}
    \vspace{-1em}
    \caption{Two-iteration MOPD training pipeline for \ourmodel.  Iteration 1 distills signals from general and agentic teachers into Ultra MOPD1. Iteration 2 initializes additional teachers from Ultra MOPD1, reuses first-round teachers, and distills all resulting signals into Ultra Final.  RLVR Student also serves as a self-teacher for areas not covered by specialized teachers.}
    \label{fig:mopd_detail}
    \vspace{-0.5em}
\end{figure}

During MOPD, the student model (obtained from RLVR) generates rollouts across all domains and receives dense reward signals from the corresponding teacher models. To maximize efficiency, MOPD is conducted asynchronously, with student rollout generation, teacher scoring, and student optimization fully pipelined. Moreover, MOPD is performed over multiple iterative cycles: after obtaining an MOPD-trained checkpoint, we branch out new rounds of teacher training initialized from the updated student model and subsequently merge the resulting improvements back into the next MOPD stage. This iterative co-evolution between student and teachers enables continuous capability improvement and progressively stronger specialization across domains. On training Nemotron 3 Ultra, we conduct two iterations of MOPD, with exact procedure shown in Figure~\ref{fig:mopd_detail}.

\subsubsection{Algorithm}

Building on prior work on on-policy distillation~\citep{yang2026nemotron,lu2025onpolicydistillation,xiao2026mimo}, we formulate asynchronous MOPD with a student policy $\pi_\theta$ and a set of domain-specialized teacher policies $\{\pi^{T_i}\}_{i=1}^{N}$, where each teacher $\pi^{T_i}$ is associated with a domain dataset $\mathcal{D}_i$. For a prompt $q \sim \mathcal{D}_i$ and a
student-generated completion $y=(y_1,\ldots,y_H)$, let
$s_t=(q,y_{<t})$ be the prefix state at token position $t$. MOPD trains the
student to match the corresponding teacher on states induced by the student
itself. In the fully on-policy case, this corresponds to maximizing the
negative reverse-KL objective
\begin{equation}
\mathcal{J}_{\mathrm{MOPD}}(\theta)
=
\sum_{i=1}^{N} \lambda_i
\mathbb{E}_{q \sim \mathcal{D}_i,\; y \sim \pi_\theta(\cdot|q)}
\left[
\sum_{t=1}^{H}
\log \pi^{T_i}(y_t|s_t)
-
\log \pi_\theta(y_t|s_t)
\right],
\label{eq:mopd_reverse_kl}
\end{equation}
where $\lambda_i$ controls the sampling or loss weight of domain $i$. Equivalently,
at each prefix $s_t$, the student minimizes
$D_{\mathrm{KL}}(\pi_\theta(\cdot|s_t)\|\pi^{T_i}(\cdot|s_t))$. Thus, unlike
mixed-environment RLVR, where the reward is typically sparse and environment
dependent, MOPD provides a dense token-level learning signal from the relevant
teacher distribution.

In our implementation, MOPD is executed asynchronously. Rollout workers,
teacher-scoring workers, and learner workers run in a pipeline. A trajectory
may therefore be generated by a stale behavior policy
$\pi_{\mathrm{behav}}$, while the learner optimizes a newer student snapshot.
To stabilize this setting, we decouple the behavior policy from the proximal
policy $\pi_{\mathrm{prox}}$ used as the trust-region center~\citep{fu2026areal}. For each sampled
token, we compute
\begin{align*}
\ell^{\mathrm{behav}}_t &= \log \pi_{\mathrm{behav}}(y_t|s_t), \\
\ell^{\mathrm{prox}}_t  &= \log \pi_{\mathrm{prox}}(y_t|s_t), \\
\ell^{T_i}_t            &= \log \pi^{T_i}(y_t|s_t).
\end{align*}
The dense distillation advantage is the sampled negative reverse-KL estimate
with respect to the proximal policy:
\begin{equation}
\widehat{A}_t
=
\mathrm{sg}\!\left[
\ell^{T_i}_t - \ell^{\mathrm{prox}}_t
\right],
\label{eq:mopd_advantage}
\end{equation}
where $\mathrm{sg}[\cdot]$ denotes stop-gradient. 
We denote the behavior-to-proximal importance ratio and the proximal-to-current
policy ratio by
\begin{align*}
c_t
=
\mathrm{sg}\!\left[
\frac{\pi_{\mathrm{prox}}(y_t|s_t)}
     {\pi_{\mathrm{behav}}(y_t|s_t)}
\right],
\qquad
r_t(\theta)
=
\frac{\pi_{\theta}(y_t|s_t)}
     {\pi_{\mathrm{prox}}(y_t|s_t)}.
\end{align*}
Here, $c_t$ accounts for the mismatch between the stale rollout policy
$\pi_{\mathrm{behav}}$ and the proximal learner policy $\pi_{\mathrm{prox}}$,
while $r_t(\theta)$ is the policy ratio optimized by the learner. PPO-style
clipping is applied to $r_t(\theta)$ around the proximal policy
$\pi_{\mathrm{prox}}$.
The learner maximizes the clipped asynchronous MOPD surrogate
\begin{align}
\mathcal{J}_{\mathrm{async\text{-}MOPD}}(\theta)
=
\mathbb{E}_{q \sim \mathcal{D}_i,\; y \sim \pi_{\mathrm{behav}}}
\left[
\sum_{t=1}^{H}
m_t c_t
\min\!\left(
r_t(\theta)\widehat{A}_t,\,
\mathrm{clip}\!\left(r_t(\theta),1-\epsilon,1+\epsilon\right)\widehat{A}_t
\right)
\right],
\label{eq:async_mopd}
\end{align}
where $m_t$ represents token-level masking using IcePop strategy~\citep{ring1t}.

MOPD training uses a maximum generation length of 192K tokens, matching the longest generation length used across all teacher training runs. Each training batch contains 1,024 prompts, with one rollout per prompt. In our ablation studies, using multiple rollouts did not yield additional benefits.

\subsubsection{Specialized Teachers}

\paragraph{Software Engineering Teacher.}
The SWE teacher was trained through a three-stage pipeline. We first applied SFT to the Ultra base model on a blend of agentic data. Next, we ran PivotRL~\citep{yi2026pivotrl} on single-step agentic environments. In the final end-to-end SWE-RL stage, the model interacts with a code repository over multiple turns, issuing tool and bash commands to produce a patch after which the verifier runs the hidden tests and assigns a binary reward used in GRPO. The final reward does not always help with trajectory-level behaviors and can sometimes wrongly reward or penalize trajectories which motivates the following adjustments. We mask the loss on unfinished trajectories (those that hit the maximum number of agent turns or trigger agent/eval timeouts), and also penalize malformed reasoning and tool calls by assigning negative advantage to the offending tokens. Additionally, to prevent the agent from cheating by reading the gold patch out of the task container, we close two leak channels. First, before the agent starts, we rewrite the in-container repository to look like a fresh clone taken at the moment of the task's base commit such that the future commits are not just hidden but physically deleted and cannot be recovered by any low-level git recovery command. Second, we install a runtime command filter that blocks every way the agent could pull that history back over the network using remote git operations or any attempt to download from GitHub's web, raw-content, or Pages domains with HTTP tools. Our end-to-end RL was conducted using 192K generation length and a maximum of 200 agent turns.

\paragraph{Office \& Workplace Task Teacher.}
To extend agentic capabilities beyond software engineering and technical domains, we trained a teacher specialized for the types of tasks measured by the GDPval benchmark \citep{patwardhan2025gdpval}. Each example is structured as a professional work assignment intended to capture meaningfully productive economic tasks typically performed by human professionals. The model is given a prompt, often with supporting reference files, and must produce a final set of deliverables, such as a spreadsheet, document or report, music/audio file, or other artifact. This makes GDPval qualitatively different as it depends not only on arriving at the correct conclusion, but also on reading the available materials, organizing intermediate evidence, following implicit professional conventions, and producing an output that must be qualitatively acceptable to a human evaluator or proxy judge.

We initialized the office and workplace task teacher from a Nemotron 3 Ultra checkpoint that completed the general SFT post-training phase. We then constructed a training distribution from AfterQuery (AQ) tasks that share important latent structure with GDPval, including file-grounded reasoning, professional deliverables, multi-step analysis, and judged final outputs. For each AQ task, we used a strong model to generate multiple full trajectory rollouts.

These rollouts were used in two stages. First, before pivot RL, we performed light SFT directly on the student Ultra model. The goal of this step was to transfer the strong model’s workflow priors for GDPval-like tasks to the student. Second, after this MOPD warmup, we proceeded with pivot RL in the MOPD stage, distilling the SFT-trained teacher into the student Ultra model using pivots derived from the strong model’s AQ rollouts.


\paragraph{Search Teacher.} 
Search-based agents often accumulate long and noisy interaction histories, since retrieved documents may be verbose, redundant, or only partially relevant. Without explicit context management, models can exhaust their context window before completing long-horizon questions that require iterative query refinement and evidence aggregation. Nemotron 3 Super~\citep{nvidia2026nemotron3superopen} was trained on search data without explicit context-management supervision. For Nemotron 3 Ultra, we therefore train a search-specialized teacher, initialized from an Ultra checkpoint, using SFT on trajectories augmented with context-management behavior.

The training data exposes the model to multiple strategies for operating under a finite context budget, including discard-all resets and summary-based compression. We focus primarily on discard-all context management, where earlier search observations are removed once the interaction history exceeds the context budget, which allows models to search for longer effective contexts that its official context length.

\paragraph{Terminal-use Teacher.}
We start with expert trajectories for tasks that was curated to challenge the model in long timeout settings, where the model needs to run the task for up-to one hour. We utilize PivotRL \citep{yi2026pivotrl} to iteratively improve the model on this data, introducing re-profiling steps whenever we observe accuracy saturation. 

\paragraph{Conversational Tool-use Teacher.}
We start with the same data and recipe from Nemotron 3 Super~\citep{nvidia2026nemotron3superopen}, training the model on conversational tool-use data through PivotRL~\citep{yi2026pivotrl}. For \ourmodel, we expand the data to tasks requiring sequential and dependent multi-step actions to discourage premature termination in conversational agent settings.

\paragraph{Model Usability Teacher.}
We expand our model usability training beyond the structured schema formatting in Nemotron 3 Super~\citep{nvidia2026nemotron3superopen} to cover three additional targets: document extraction, citation formatting, and freeform text formatting. For structured schema formatting, we create an improved dataset covering five schema types: JSON, YAML, XML, TOML, and CSV. In addition, we increase structured outputs tasks to six categories: direct extraction, translation, multistep-related (dependent follow-ups), multistep-unrelated (independent follow-ups), schema-only, and error correction. For document extraction, we provide varied structured extraction tools to the model with distractors to teach the model to invoke complex extraction tools with deeply nested fields. For citation formatting, we teach the model to use multiple inline citation formats to reference parts of a given document in its output. For freeform text formatting, we teach the model to follow diverse markdown styling instructions when answering a query grounded in a given document. All seed data was created using Nemo Data Designer~\citep{nemo-data-designer} with \texttt{openai/gpt-oss-120b}~\citep{openai2025gptoss120bgptoss20bmodel}, and all environments are implemented through Nemo-Gym~\citep{nemo-gym}.

\paragraph{Agentic Safety Teacher.}
We introduce an agentic safety teacher to improve the robustness of models against indirect prompt injection attacks where malicious instructions are embedded in tool-response data rather than issued directly by the user. 

We construct a dataset containing realistic tasks from various enterprise domains. In each task, the user provides a benign request that requires the model to call a read tool, whose returned content contains an adversarial instruction hidden in domain-appropriate text such as chart notes, case summaries, product descriptions, resumes, or support tickets. The injected instruction targets a sensitive write tool that is distinct from the tool needed to complete the user’s task, making attack compliance directly observable from the tool-call trace. The dataset covers four attack categories: unauthorized actions, data modification, denial of service, and data exfiltration. 

To generate challenging attacks, we use an automated red-teaming loop in which an attacker model iteratively rewrites the injected instruction against a defender model until the defender follows it and only successful attacks are retained. We use \supermodel as the attacker model and \nanomodel as the defender. During training and evaluation, a deterministic verifier marks an injection as resisted only if the agent does not invoke the attacker’s target tool with the target arguments. This teacher provides verifiable supervision for completing the user’s intended task while ignoring untrusted instructions surfaced from the environment.

\paragraph{Chat Teacher.}
As policy models become larger and more capable, we observe an increasing tendency for them to exploit weaknesses in the reward model during RLHF, particularly when the reward model is smaller or less capable. Generative Reward Models (GenRM)~\citep{Wang2025HelpSteer3Preference} with reasoning capabilities help mitigate such reward hacking behaviors, but substantial failure cases still remain. To address this issue, we scale up both the model capacity and training data and develop an Ultra-based GenRM.

The GenRM is trained to evaluate a pair of candidate responses given a conversational context. When user-defined principles are provided, the model performs judgment conditioned on those principles; otherwise, it evaluates responses according to general helpfulness and quality criteria.
We trained the GenRM on top of Ultra SFT model produced in section~\ref{subsec:SFT}. GenRM training follows the same RLVR method we used in \cite{nvidia2026nemotron3superopen}, where we assign rewards to teach it to predict both individual scores for two responses and a ranking score. When multiple principles are presented, GenRM predicts the triplet for each principle then come up with an overall judgment. During RLHF, only the overall scores are used as reward signal.

Chat teacher training involves multiple RLHF iterations. After each iteration, we evaluate the policy model on internal chat benchmarks, identify weaknesses, and curate targeted data to address them. A principle-following GenRM makes this process much more flexible: instead of relying on general helpfulness, it can adapt to different principles during training and evaluation, enabling targeted improvements across cycles without retraining the reward model itself.

\paragraph{Instruction-following and Factuality Teacher.}
To further advance instruction following and factuality, we performed domain-focused RLVR on top of the RL checkpoint described in Section~\ref{subsec:RL}. The training leveraged a combination of instruction-following, abstention-focused, and RLHF environments.

The instruction-following environments spanned a diverse set of challenging scenarios, including but not limited to strict format compliance, mid-conversation instruction changes, and long-horizon conversational coherence. These capabilities were evaluated either programmatically or through LLM-as-a-judge verification. Beyond instruction following, the teacher also underwent abstention training, where the model was encouraged to abstain when uncertain rather than hallucinate incorrect answers. We dynamically calibrated the abstention reward throughout training to achieve a favorable balance between accuracy and hallucination reduction.

To avoid behavioral collapse and overfitting to training environments, we additionally incorporated RLHF data during optimization. This helped preserve response quality, helpfulness, and alignment with human preferences while improving robustness on instruction-following and factuality-oriented tasks.

\paragraph{STEM Teacher.}
This teacher focuses on the general reasoning capabilities on a wide range of subjects including math, code, natural sciences, humanities, sociology, and tool use for these domains. Starting from the student model, we perform additional stages of SFT and RL on selected datasets. The resultant teacher model matches or outperforms DeepSeek V4 Pro (High) on challenging reasoning benchmarks such as GPQA, MMLU-Pro, LiveCodeBench v6, IMOAnswerBench, and Apex Shortlist (see Table \ref{reasoning_teacher_scores}). Below we discuss the data generation and blending strategy as well as our training processes.
\begin{table}[!htp]
\centering
\small
\begin{tabular}{l|ccc}
\toprule
Benchmark & DeepSeek V4 Pro (High) & Student & General Reasoning Teacher \\
\midrule
HLE (no tools) & \bf 34.5 & 25.6 & 32.1 \\
GPQA         & \bf 89.1 & 85.0 & 88.5 \\
MMLU-Pro     & 87.1 & 85.7 & \bf 87.7 \\
LiveCodeBench v6     & 89.8 & 87.4 & \bf 90.0 \\
IMOAnswerBench     & 88.0 & 84.5 & \bf 92.5 \\
Apex Shortlist     & \bf 85.5 & 68.9 & 85.4 \\
\bottomrule
\end{tabular}
\caption{\label{reasoning_teacher_scores} Reasoning performance of the general reasoning teacher. The highest scores on each benchmark are bold. Differences within a one point range are considered noise.}
\end{table}

\textbf{Science reasoning data.}
We constructed our science dataset, spanning both STEM and non-STEM domains, by generating new reasoning traces for existing problems. The seed problems were drawn from the Nemotron Nano SFT dataset\footnote{\url{https://huggingface.co/datasets/nvidia/Nemotron-Science-v1}}, a newly curated chemistry dataset~\citep{li2025chemicalqaevaluatingllms, zhang2024chemllm}, the Multi-subject-RLVR dataset~\citep{su2025expanding}, and an internal proprietary dataset. We filtered each source to remove the easiest problems. For each problem, we generated four solution traces using \texttt{DeepSeek-V4-Pro}~\citep{deepseekai2026deepseekv4}; for more difficult problems, we generated 16 traces per problem. All traces were graded for correctness by a separate LLM judge, \texttt{gpt-oss-120b}~\citep{openai2025gptoss120bgptoss20bmodel}. To improve coverage of long-form reasoning, we additionally resampled a subset of problems whose median correct-solution length exceeded 16k tokens, generating eight traces per problem. Finally, we reserved a held-out set of 3,000 problems for RL evaluation, selected to have pass rates between 0.25 and 0.80 and median correct-solution lengths below 64k tokens.

\textbf{Coding reasoning data.}
We built our competitive coding data from approximately 14K problems collected from international programming competitions around the world over the past 10 years. The dataset includes problems from diverse contest styles and difficulty levels, spanning Olympiad-style tasks, ICPC-style problems, and regional competitive programming benchmarks. To strengthen coverage of hard algorithmic reasoning, we further add 4K difficult problems from OpenCodeReasoning~\citep{ahmad2025opencodereasoning}, which emphasize long-horizon reasoning, algorithm design, and implementation-heavy problem solving. For each problem, we generate 10 candidate solutions with DeepSeek-V4 and filter the resulting traces based on compilation success, removing solutions that fail to compile.

\textbf{Mathematical Chain-of-Thought (COT) and Tool-Integrated Reasoning (TIR) data.}
We construct the main mathematical reasoning data following the Nemotron math data pipeline~\citep{du2025nemotron}. Starting from the same source collection, we filter out simple or trivial problems and retain 95{,}164 unique math problems. For each retained problem, we generate both COT and TIR solution trajectories with \texttt{DeepSeek-V4-Pro}~\citep{deepseekai2026deepseekv4} in high-inference mode, using the model provider's recommended generation parameters. We then check each trajectory against the reference answer using an LLM-based judging pipeline with \texttt{gpt-oss-120b}~\citep{openai2025gptoss120bgptoss20bmodel}, and retain only trajectories judged correct for SFT. The final validated pool contains 285{,}516 COT examples and 259{,}915 TIR examples, for 545{,}431 examples in total.

\textbf{Mathematical proof data.}
We additionally build a dedicated mathematical proof dataset to improve rigorous theorem-proving and verification-style reasoning. The proof seed problems are drawn from the AoPS \cite{AoPS} section of the Nemotron math data collection~\citep{du2025nemotron}, covering 5{,}751 unique proof problems. For each proof problem, we generate proof-oriented traces with \texttt{DeepSeek-V4-Pro}~\citep{deepseekai2026deepseekv4} in max-inference mode, using the model provider's recommended generation parameters. Following the methodology of \texttt{DeepSeekMath-V2}~\citep{shao2025deepseekmath}, we generate proof, verification, and meta-verification responses. We retain samples that follow the prompted output structure and remove responses that hit the maximum context length. The final validated pool contains 82{,}737 samples, including proof-, verification-, and meta-verification-style responses.

\textbf{SFT data blending.}
We construct the final SFT mixture using token-level target proportions rather than example counts, since average response lengths vary substantially across data sources.
The final blend consists of 40B generated tokens organized into four main components.
Science reasoning data contributes 23.5B tokens (58.75\%), covering both STEM and non-STEM reasoning tasks.
Mathematical chain-of-thought/tool-integrated reasoning data and mathematical proof data together contribute 9.45B tokens (23.63\%), while competitive coding data contributes 4.05B tokens (10.13\%). We additionally include a smaller general-domain SFT component, contributing 3.0B tokens in total (7.50\%), to preserve broad instruction-following and open-domain reasoning capabilities.
For sources that exceed their token budget, we randomly downsample examples; for sources below their target budget, we upsample examples while preserving the original within-source mixture ratios where applicable. Training examples are packed to a maximum sequence length of $294{,}912$ tokens, and we follow the same experimental setup as Section~\ref{subsec:SFT} to train for one full epoch.

\textbf{Reinforcement learning.} After SFT, the model shows outstanding performance on math, code, and natural sciences, so in the RL stage that follows, we focus on non-STEM domains such as humanities and sociology. We mostly adopt the same setup as in Section \ref{subsec:RL} except that we use a smaller number of prompts per batch (128) and a resultant smaller global batch size (2048). We are excited to find that the training generalizes well, leading to significant improvements on all domains, instead of just non-STEM.

\paragraph{Competitive Coding Teacher.} We conduct additional Competitive Coding RL on top of General Reasoning Teacher with coding data originally from the Nemotron-Cascade~\citep{wang2025nemotroncascadescalingcascadedreinforcement, yang2026nemotron}, which contains coding prompts sourced from multiple competitive programming platforms with strong test cases for reward verification. We filter out prompts that General Reasoning Teacher solves correctly in all 8 of 8 rollouts, yielding a compact final set of only 3.5K samples. This Competitive Coding RL gives us +2.4 improvements beyond General Reasoning Teacher on LiveCodeBench v6.

\subsubsection{MOPD Warmup}

One key finding from our MOPD trials is that teacher models trained with substantially different training pipelines cannot be effectively combined through a straightforward MOPD merge, resulting in suboptimal performance. We hypothesize that when the teacher and student are trained on different SFT data, they acquire different reasoning behaviors and induce different output distributions. This distribution mismatch can cause student-generated trajectories to be out-of-distribution for the teacher, reducing the quality and reliability of the supervision signals provided by the teacher. We encountered this issue in practice because the teacher and student models were developed in parallel, with many of the agentic and reasoning teachers trained using their own specialized SFT pipelines.

To mitigate the distribution mismatch between teacher and student models, we introduce a brief warmup stage before MOPD. Specifically, the student undergoes a very light SFT on data drawn from the teacher's training distribution. The objective is to align the student's reasoning trajectories and output distribution with those expected by the teacher. This improves the reliability of teacher supervision by increasing the likelihood that student-generated trajectories remain within the teacher's support. Since the warmup stage is intentionally limited in scale, it induces minimal regression on unrelated domains, and any residual degradation is subsequently recovered through MOPD training.

Table~\ref{tab:mopd_warmup} presents ablation results across three representative domains. The results show that, in agentic domains, the warmup stage substantially improves performance after MOPD training. In contrast, for general reasoning tasks such as HLE, the warmup provides only negligible gains. We discuss possible explanations for this discrepancy in Section~\ref{sec:mopd_discussion}.

\begin{table}[!htp]
\centering
\small
\begin{tabular}{l|cccc}
\toprule
Benchmark & Student & Warmup & No Warmup & Teacher\\
\midrule
GDPVal         & 28.9 & 46.7 & 35.3 & 49.5 \\
BrowseComp     & 31.0 & 44.4 & 33.0 & 51.0\\
HLE (no tools) & 25.6 & 26.7 & 26.3 & 32.1\\
\bottomrule
\end{tabular}
\caption{Warmup ablation for MOPD across three representative domains.}
\label{tab:mopd_warmup}
\end{table}

\subsubsection{Results and Discussions}
\label{sec:mopd_discussion}

\begin{table*}[!htp]
\centering
\footnotesize
\setlength{\tabcolsep}{2.5pt}
\renewcommand{\arraystretch}{1.05}

\resizebox{0.92\textwidth}{!}{%
\begin{tabular}{l|c c c c c c}
\toprule
\textbf{Benchmark}
& \makecell{\textbf{SFT}}
& \makecell{\textbf{RLVR}}
& \makecell{\textbf{MOPD1}}
& \makecell{\textbf{MOPD2}}
& \makecell{\textbf{Teacher}}
& \makecell{\textbf{Recovery(\%)}} \\
\midrule

Terminal Bench 2.0 & 34.5 & 44.5 & 50.8 & 54.0 & 50.0  & 172.7\% \\
GDPVal & 23.2 & 28.9 & 46.7 & 46.7 & 49.5  & 86.4\% \\
SWE-Bench Verified & 63.5 & 65.8 & 70.1 & 71.7 & 72.5 &  88.1\% \\

TauBench Telecom & 55.7 & 82.7 & 91.2 & 92.9 & 94.0 & 90.3\% \\

BrowseComp & 14.3 & 31.0 & 41.0 & 44.4 & 51.0 & 67.0\% \\

LiveCodeBench (v6) & 85.5 & 87.4 & 90.0 & 89.0 & 92.4 & 32.0\% \\
IMOAnswerBench (no tools) & 85.1 & 84.5 & 88.1 & 88.6 & 92.5 & 51.3\% \\

HLE (no tools) & 19.7 & 25.6 & 25.9 & 26.7 & 32.1 & 16.9\% \\

OmniScience Non-Hallucination & 4.8 & 46.3 & 77.9 & 78.7 & 87.0  & 79.6\% \\

IFBench (prompt loose) & 62.3 & 78.4 & 80.0 & 81.7 & 83.0 & 71.7\% \\
Multi-Challenge & 53.3 & 60.3 & 62.8 & 63.8 & 63.3 & 116.7\% \\

\bottomrule
\end{tabular}%
}

\caption{MOPD results across domains, showing gains over the RLVR student and recovery toward specialized teachers.}

\label{tab:mopd_student_teacher}
\end{table*}

We report the main MOPD results in Table~\ref{tab:mopd_student_teacher}, where \texttt{RLVR} denotes the initial student checkpoint, and \texttt{MOPD1} and \texttt{MOPD2} denote the checkpoints after the first and second MOPD iterations, respectively. Recovery rate is defined as $(\texttt{MOPD2}-\texttt{RLVR})/(\texttt{Teacher}-\texttt{RLVR})$, representing the fraction of the teacher-student performance gap closed by MOPD. MOPD improves over the \texttt{RLVR} student across the evaluation suite, with strong recovery on both agentic benchmarks, such as Terminal Bench, GDPVal, SWE-Bench Verified, TauBench Telecom, and BrowseComp, and instruction-following/factuality benchmarks, such as OmniScience, IFBench, and Multi-Challenge. On several benchmarks, MOPD even surpasses the corresponding specialized teacher, suggesting positive cross-domain generalization from merging supervision across multiple teachers. For example, we find that \texttt{MOPD2} checkpoint significantly outperforms teacher on data science related tasks in Terminal Bench, indicating a potential knowledge transfer from office/productivity workflows. Overall, these results indicate that MOPD is particularly effective when the teacher's advantage can be expressed as token-level preferences over trajectories that the student is already able to sample, such as tool-use decisions, environment interactions, abstention behavior, and multi-step execution patterns.

The gains are smaller on self-contained reasoning benchmarks, especially HLE. We believe this reflects a limitation of the on-policy distillation setting rather than a failure of the teacher. The general reasoning teacher is initialized from the student, but its gains come from additional large-scale SFT and RL on a separate reasoning mixture generated by DeepSeek-V4-Pro. The student has not directly seen this data. As a result, the teacher's advantage is not simply a different preference over trajectories already produced by the student; it also comes from capabilities acquired through additional off-policy data exposure. Since MOPD scores student-generated trajectories, its learning signal is strongest when those trajectories lie within the teacher's support. When the missing capability requires reasoning paths that the student rarely samples, student rollouts become effectively out-of-distribution for the teacher, making the token-level supervision less informative.

This interpretation is consistent with the warmup ablation in Table~\ref{tab:mopd_warmup}. Warmup substantially improves MOPD in agentic domains, where increasing overlap between student rollouts and teacher-supported trajectories makes teacher scoring more informative. In contrast, warmup has little effect on HLE, suggesting that the remaining gap is driven less by a shallow trajectory mismatch and more by capabilities introduced through the general reasoning teacher's additional SFT/RL training.

\subsubsection{Limitations and Open Problems}
In this section we will discuss several aspects of MOPD that remain unresolved and would benefit from further study.
During the development of MOPD, we evaluated several technically plausible variants that did not improve performance under our current experimental setup. We share these observations to make the empirical picture more complete, but they should not be interpreted as evidence that these approaches are fundamentally ineffective. With further research, we believe these directions could yield MOPD approaches that outperform our current setup, and we encourage the community to continue exploring them.

\begin{itemize}
\item \textbf{Logit matching.}
A natural alternative to the sampled-token objective is distribution-level distillation, where the student is trained to match the teacher's predictive distribution over the top-$k$ tokens or the full vocabulary at each prefix. In our preliminary experiments, these objectives did not improve MOPD performance and consistently underperformed the sampled-token objective on some agentic benchmarks such as Terminal Bench. We hypothesize that full-distribution matching may impose an overly strong local constraint on prefixes sampled from the student policy, particularly when those prefixes have limited support under the teacher distribution. In this regime, teacher logits can become poorly calibrated or less informative, and matching the full distribution may amplify noise from off-support states. By contrast, the sampled-token objective applies supervision only to realized actions and may therefore provide a more stable on-policy learning signal. Characterizing when broader distributional supervision is beneficial remains an open problem.

\item \textbf{Foundations for MOPD.}
As discussed above, MOPD is most effective when student-generated trajectories remain within the teacher's support, allowing the teacher to provide reliable supervision. One possible approach is to ensure that the teacher and student share a unified SFT stage before specialization. Another is to first develop specialized teachers, use them to generate SFT data, and then train the student with this data before applying RL or MOPD. Due to time and resource constraints, we did not systematically evaluate these alternatives in this project and will leave them for future work.
\item \textbf{MOPD on long-horizon tasks.}
Agentic workflows require many turns of tool calls and environment interactions, whereas reasoning tasks are typically single-turn. When mixing end-to-end agentic environments with reasoning environments in MOPD, we observed substantial training inefficiency because rollout times can differ dramatically. Balancing efficiency and accuracy requires sophisticated training infrastructure together with careful asynchronous algorithm design. In practice, we use single-turn rollouts, similar to PivotRL~\citep{yi2026pivotrl}, for most of the agentic tasks. This approach performs relatively well, but whether end-to-end rollouts can yield further gains, and how to make it robust to potential distribution mismatch~\citep{wang2026tcod}, remains an open area for exploration.

\end{itemize}

\subsection{MTP Boosting}
\label{subsec:mtp_boosting}
\ourmodel ships with native speculative-decoding support via a Multi-Token Prediction (MTP) head trained throughout all stages of training, following \supermodel~\citep{nvidia2026nemotron3superopen}. The MTP head is an internal drafter that predicts multiple future tokens from the backbone's hidden states; at inference, these draft tokens are verified against the backbone (the target model), and several can be accepted in a single verification step, either through rejection sampling~\citep{specdec} or naive 1-1 token matching. As in \supermodel, we use a \emph{shared} MTP-head formulation applied recursively for several MTP steps, so the draft horizon grows without requiring additional parameters.

\paragraph{Train-Inference Mismatch.}
Even with a shared head, naive teacher-forced MTP training does not match autoregressive MTP inference. Let the input to the first MTP step be $(h_1, \ldots, h_n)$, and let the corresponding outputs be $(h^{mtp_1}_2, \ldots, h^{mtp_1}_{n+1})$. During training, the input to the second MTP step is the full shifted sequence $(h^{mtp_1}_2, \ldots, h^{mtp_1}_{n+1})$, all of which originate from the previous MTP step. At inference time, however, the conditioning structure differs: the newly produced state $h^{mtp_1}_{n+1}$ is generated while attending to the \emph{previous} backbone hidden states $(h_1, \ldots, h_n)$, so the effective input to the second MTP step becomes $(h_1, \ldots, h_n, h^{mtp_1}_{n+1})$. One step further, it becomes $(h_1, \ldots, h_n, h^{mtp_1}_{n+1}, h^{mtp_2}_{n+2})$, and as the draft length grows, later MTP steps condition on an increasingly noisy mixture of target-model and MTP-generated hidden states. This distribution differs from the teacher-forced training distribution and degrades acceptance at deeper draft positions.

\paragraph{Training Procedure.}
The goal of \emph{MTP Boosting} is to make the MTP head match the backbone's next-token distribution under the input conditions or noise it encounters at inference. To address this, we continue to train the MTP starting from the MOPD checkpoint (Section~\ref{subsec:MOPD}). The backbone is fixed for the entire training phase, and only the MTP head receives gradient updates. This ensures there is no risk of regressing backbone quality, and substantially reduces the optimizer-state and activation memory footprint of each step. At this stage, we modify the MTP forward pass such that the hidden states passed as input to the MTP at step $k$ are sampled from the set of hidden states produced at MTP steps $1,..,k-1$, rather than simply taking the previous MTP step's generated hidden states. This procedure exposes the head at training time to a similar noise it encounters at inference, and produces a drafting head that handles longer draft lengths more gracefully.

\paragraph{Data.}
We generate on-policy rollouts using the MOPD checkpoint described in Section~\ref{subsec:MOPD} by starting from seed prompts drawn from \textit{Nemotron-Post-Training-Dataset-v2}\footnote{\url{https://huggingface.co/datasets/nvidia/Nemotron-Post-Training-Dataset-v2}} and \textit{Nemotron-RL-Super-Training-Blends}\footnote{\url{https://huggingface.co/datasets/nvidia/Nemotron-RL-Super-Training-Blends} } to cover general-purpose and agentic inputs, respectively. Rollouts are sampled with $temp=1$. The MTP head was trained on these rollouts for 12K steps at a global batch size of 64, with sequences capped at 8K tokens. The loss was accumulated over the assistant response in each sample.

\paragraph{Loss.}
We use a temperature-scaled forward-KL loss against the backbone's logits. The standard cross-entropy term against the gold token is disabled, so the head matches the backbone's full distribution rather than the one-hot label. Let $\mathcal{A}$ denote the set of assistant-token positions. For each assistant token position $t\in \mathcal{A}$ and MTP generation step $k \in \{1,\ldots,N_{\mathrm{mtp}}\}$, let $z_t$ denote the backbone's logits at position $t$ and $z^{mtp_k}_{t+k}$ the MTP logits at step $k$ starting from position $t$. The boosting objective is
\begin{equation}
\mathcal{L}_{\mathrm{MTP}}(\theta)
=
\frac{T^{2}}{N_{\mathrm{mtp}}|\mathcal{A}|}
\sum_{k=1}^{N_{\mathrm{mtp}}}
\sum_{t \in \mathcal{A}}
D_{\mathrm{KL}}\!\left(
\sigma\!\left(z_{t+k}/T\right)
\,\Big\|\,
\sigma\!\left(z^{\mathrm{mtp}_k}_{t+k}/T\right)
\right).
\label{eq:mtp_kd}
\end{equation}
where $\sigma$ denotes the softmax operator, $T = 2$ is the distillation temperature, $N_{\mathrm{mtp}}=7$ is the number of MTP steps. The $T^{2}$ factor follows~\citep{hinton2015distilling}.

\paragraph{Results.}
We evaluate MTP accuracy using SPEED-Bench~\citep{2026speedbench}, measuring the per-sample acceptance lengths (ALs) across the categories in the \textit{qualitative} data split. Table~\ref{tab:mtp_speedbench} reports the average ALs by category. The Boosted-MTP head delivers consistent gains over the baseline at all draft positions, with the improvement most pronounced at deep draft positions where the train-inference mismatch is most visible. 
\begin{table}[ht!]
\centering
\setlength{\tabcolsep}{4pt}
\resizebox{0.9\linewidth}{!}{
\begin{tabular}{l|cc|cc}
\toprule
Category
& \ourmodel
& \ourmodel + MTP-Boosting
& Qwen3.5-397B-A17B
& DeepSeek-V4-Flash \\
\midrule
Coding        & 5.152 (4.739) & 5.452 (4.872) & \textbf{5.550} (\textbf{5.307}) & 2.835 (2.584) \\
Humanities    & 3.950 (3.738) & \textbf{4.102} (3.857) & 4.095 (\textbf{3.875}) & 2.611 (2.402) \\
Math          & 5.127 (4.661) & \textbf{5.343} (\textbf{4.733}) & 5.002 (4.646) & 2.934 (2.747) \\
Multilingual  & 5.141 (4.937) & \textbf{5.382} (\textbf{5.179}) & 4.951 (4.836) & 2.811 (2.666) \\
QA            & 4.251 (4.089) & \textbf{4.469} (\textbf{4.331}) & 4.371 (4.219) & 2.586 (2.419) \\
RAG           & 5.207 (5.098) & \textbf{5.380} (\textbf{5.367}) & 5.093 (5.051) & 2.868 (2.768) \\
Reasoning     & 4.672 (4.315) & \textbf{4.896} (\textbf{4.502}) & 4.572 (4.466) & 2.782 (2.631) \\
Roleplay      & 2.801 (2.767) & 2.940 (2.856) & \textbf{3.972} (\textbf{3.807}) & 2.149 (2.056) \\
STEM          & 4.258 (3.935) & \textbf{4.435} (\textbf{4.094}) & 4.323 (4.062) & 2.703 (2.462) \\
Summarization & 4.413 (4.321) & 4.552 (4.468) & \textbf{4.787} (\textbf{4.785}) & 2.642 (2.560) \\
Writing       & 3.285 (3.209) & 3.476 (3.385) & \textbf{3.667} (\textbf{3.605}) & 2.419 (2.201) \\
\midrule
Average       & 4.387 (4.165) & \textbf{4.584} (4.331) & 4.580 (\textbf{4.423}) & 2.667 (2.500) \\
\bottomrule
\end{tabular}}
\captionof{table}{MTP average acceptance lengths on the SPEED-Bench \textit{qualitative} split using draft length 7. Main values are obtained with greedy decoding; values in parentheses use temperature sampling with $temp=1$. MTP-Boosting consistently increases acceptance length over the base MTP, yielding relative speculative-decoding speedup improvements from 3.15\% on summarization tasks to 5.82\% on coding tasks.}
\label{tab:mtp_speedbench}
\end{table}

\subsection{Reasoning Efficiency and Control}
\label{subsec:reasoning_effort}

Nemotron 3 Ultra is trained for three reasoning modes: reasoning-off, regular and medium-effort. The regular and medium-effort reasoning modes have the option to be used in conjunction with inference-time budget control. These combinations of controls provide flexibilities that cover the entire spectrum of accuracy-efficiency trade-off to meet customers’ needs in various application scenarios and complement task-level controls such as turn-limits in agentic applications.

The medium-effort reasoning mode is introduced during the SFT stage and later optimized during the RLVR stage. Approximately 2.5\% of the RLVR training prompts are in the medium-effort mode and they cover math, STEM and coding, and length-based adjustments are applied on the RL rewards for them. This recipe has trained both Nemotron 3 Ultra and Super and the effects of optimization generalize to a variety of tasks beyond math, STEM and coding, and we were able to calibrate the resulting effort mode by adjusting hyperparameters.

\begin{figure}[ht!]
    \centering
    \includegraphics[width=0.9\linewidth]{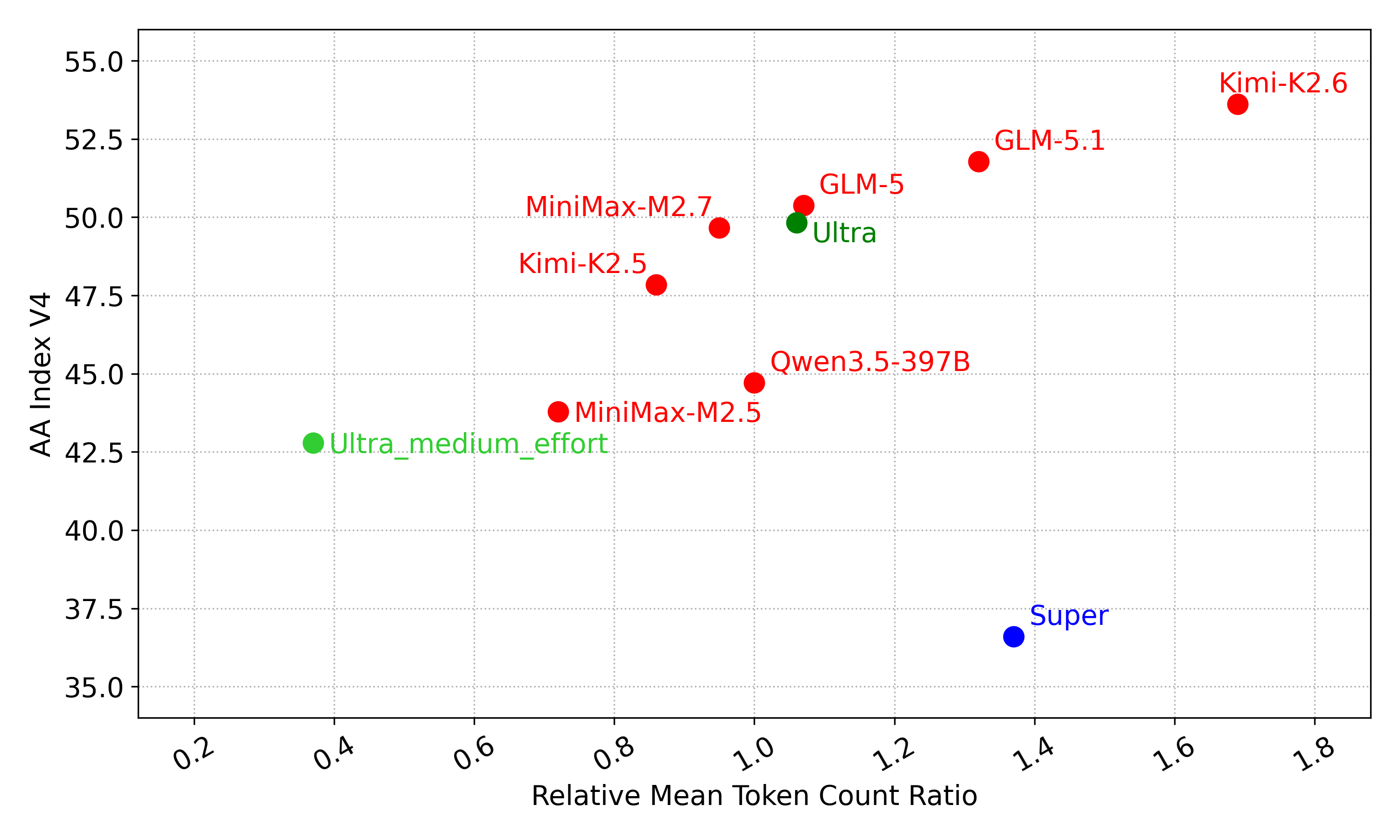}
    \vspace{-0.5em}
    \caption{Accuracy-efficiency comparisons on Artificial Analysis Intelligence Index V4 tasks.}
    \label{fig:reason_effort}
    \vspace{-0.5em}
\end{figure}

Figure~\ref{fig:reason_effort} compares Nemotron 3 Ultra, Nemotron 3 Ultra medium-effort and Nemotron 3 Super against a set of open-source models. The y-axis is Artificial Analysis Intelligence Index V4 while the x-axis is a relative verbosity measure by using Qwen 3.5 397B's average token usage on each task as reference and by averaging over the 10 tasks in AA Index V4. It shows that Ultra's medium-effort mode uses on average around 2.5X less tokens than the regular mode at the cost of approximately 7\% drop in accuracy.

\subsection{Infrastructure}
\label{subsec:infrastructure}

\subsubsection{Accelerating Rollout Generation with Multi-Token Prediction}

During RL and MOPD, we train using a one-step off-policy asynchronous RL setup, so rollout generation is overlapped with the policy update, and the step time is bounded by whichever stage is slower.
In our setting, the slower stage is typically rollout generation, whose time is in turn dominated by a small fraction of straggler generations that run substantially longer than the rest of the batch.
In order to accelerate rollout generation, we use speculative decoding with Multi-Token Prediction (MTP).
At each decoding iteration, the MTP head is applied recurrently to propose $k$ candidate tokens, which the base model verifies in a single forward pass.
Accepted tokens are committed without additional sequential decoding steps.

\begin{figure}[!ht]
    \centering
    \includegraphics[width=0.9\linewidth]{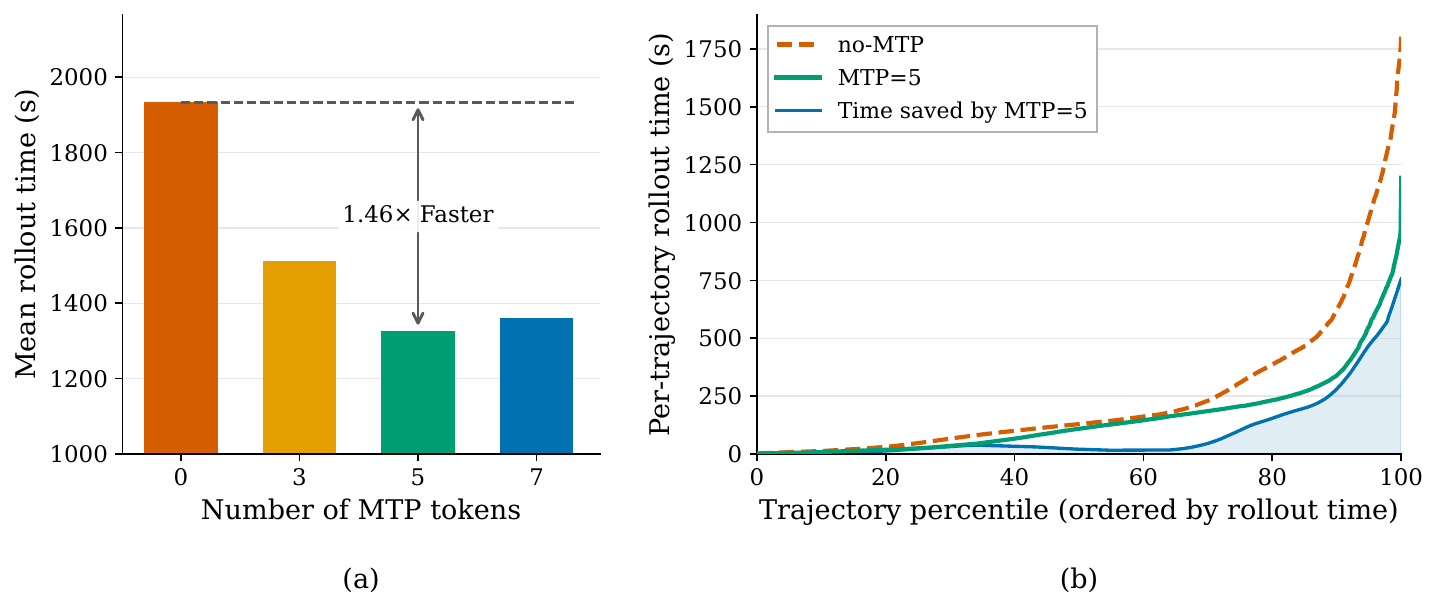}
    \vspace{-0.5em}
    \caption{(a) Mean per-step rollout-generation time during RLVR training vs.\ number of MTP tokens $k$. $k{=}5$ gives a $1.46\times$ speedup over the $k{=}0$ (no MTP) dashed baseline. (b) Per-trajectory rollout time for a representative RLVR rollout step, $k{=}5$ vs.\ no-MTP (trajectories ordered by rollout time), and the time saved by MTP. The benefit is concentrated in the long-tail (slowest) generations.}
    \label{fig:mtp_rollouts}
    \vspace{-0.5em}
\end{figure}

To find the best $k$, we sweep $k \in \{0, 3, 5, 7\}$, where $k{=}0$ is the standard no-MTP baseline.
As shown in Figure~\ref{fig:mtp_rollouts}(a), enabling MTP speeds up rollout generation, with $k{=}5$ giving the largest gain at $1.46\times$ over the baseline.
Examining the per-trajectory rollout times for a representative rollout step (Figure~\ref{fig:mtp_rollouts}(b)), we find MTP is particularly beneficial for the long-tail (slowest) generations.
We attribute this to a combination of factors: these long generations emit many more tokens, and they decode toward the end of the batch at lower concurrency, where speculative decoding tends to be most effective.

\subsubsection{Scaling RL Infrastructure}
\noindent Our production cluster is deployed on NVIDIA GB200 nodes with Slurm orchestration and co-located CPUs for sandbox execution. The observed RL software failure breakdown is shown in Table~\ref{tab:failure-attribution}.
\begin{table}[!ht]
\centering
\begin{tabular}{lr}
\hline
\textbf{Failure Category} & \textbf{\%} \\
\hline
Generation engine failures / timeouts & 56 \\
Sandbox / tool calling & 36 \\
Other Software issues & 8 \\
\hline
\end{tabular}
\caption{Failure attribution. Generation and sandbox/tool calling failures account for $\sim$92\% of the failures.}
\label{tab:failure-attribution}
\end{table}

\begin{table}[!ht]
\centering
\begin{tabular}{lcc}
\toprule
\textbf{Issue} & \textbf{Before} & \textbf{After} \\
\midrule
Ray GCS startup           & 30+ min          & 10 min \\
Checkpoint blocking       & 60 sec              & {$<$1 sec} \\
Cold init (JIT)           & 38.8 min         & 0.4 min \\
Multi-node vLLM startup   & 25 min           & 9.5 min \\
Container extraction      & 2--3 min (with cascading failures) & $\sim$0s (warm) \\
\bottomrule
\end{tabular}
\caption{Summary of key infrastructure optimizations and their impact.}
\label{tab:improvements}
\end{table}

Achieving high resiliency required a sustained engineering effort through systematic instrumentation and optimizations, which we subsequently review in detail. 

\subsubsection*{Ray GCS Scalability and Slurm Launch Overheads}
The NeMo-RL job is heterogeneous, it must launch different roles (training workers, vLLM generation workers, gym workers, and judge workers), each with different resource requirements,  across different node subsets. At the start of post-training, our RL launch script issued multiple, separate \texttt{srun} invocations per node to assign each node its role, start its processes, and monitor progress. Each \texttt{srun} is an RPC to the Slurm controller daemon (\texttt{slurmctld}), which must process them serially. At scale, this meant hundreds of serial RPCs queued against \texttt{slurmctld}, a pattern that scaled $O(n)$ with the node count and quickly became a scaling bottleneck. Restructuring to a single multi-node \texttt{srun} reduced Slurm controller interactions to $O(1)$. Together, startup cost dropped from over 30 minutes to 10 minutes..

Additionally, the RL job creates a massive number of Ray actors simultaneously across policy, generation, and environment workers. At 3K+ GPU scale, Ray's single-threaded Global Control Service (GCS) was overwhelmed by actor registrations, causing long startups (25-49 minutes) and thundering-herd startup failures. We eliminated 40\% of actor registrations by converting short-lived actors to tasks, pooling initialization actors per node, and applied aggressive GCS tuning. Anyscale resolved the GCS scalability regressions with the fixes shipped in the Ray 2.55 public release.

\subsubsection*{Topology-Aware NVLink Domain Placement}

On GB200 NVL72, the NVLink domain spans an entire rack (72 GPUs across 18 nodes). Without
topology awareness during worker placement, Megatron's Expert Parallelism (EP) groups can span multiple
racks, forcing MoE all-to-all communication over InfiniBand instead of NVLink.

NeMo RL had no concept of NVLink domains: Ray node IDs are random UUIDs with no
topological relation, and Megatron trusts the rank order it receives. If the
framework assigns ranks without NVLink domain awareness, EP groups will silently
span racks. The fix was to make rank assignment domain-aware and ensure all GPUs
within an EP group are co-located on the same rack. Note that inter-rack ordering
is not a concern here, as our DP size does not require exploiting cross-rack rail
optimizations.

The fix was threefold. First, at container startup (before \texttt{ray start}),
\texttt{ray.sub} generates and sources a probe script that parses the NVLink fabric
\texttt{ClusterUUID} from \texttt{nvidia-smi -q} -- a hex string identical for all
GPUs in the same NVLink domain -- and registers it as a Ray custom resource
(\texttt{nvlink\_domain\_<ClusterUUID>}), giving the scheduler rack-membership
information it previously lacked. The same probe derives a topology rank
(\texttt{topo\_rank}) from \texttt{SLURM\_TOPOLOGY\_ADDR}, falling back to
\texttt{SLURM\_PROCID} or hostname digits. Second, inside NeMo RL's
\texttt{RayVirtualCluster}, bundle indices are sorted by a
$(\texttt{domain\_min\_topo\_rank},\ \texttt{topo\_rank},\ \texttt{gpu\_id})$
composite key, ordering NVLink domains by their minimum topology rank and nodes
within each domain by physical position, so that EP groups remain within a single
NVLink domain. When \texttt{segment\_size} is set, bundles from domains that cannot
contribute a complete segment are discarded. The sorted order drives per-worker
\texttt{RANK} and \texttt{CUDA\_VISIBLE\_DEVICES} assignment, and Megatron runs with
\texttt{external\_gpu\_device\_mapping=True} so it trusts Ray's GPU pinning rather
than re-deriving device placement. Third, the \texttt{topo\_rank} (from
\texttt{SLURM\_TOPOLOGY\_ADDR}) corrects the scrambled Slurm block ordering, with
\texttt{ray.sub} additionally applying a deterministic hostname sort before
Ray head-node assignment.

With these changes, training and generation actors are co-located in the same
NVLink domain, and EP all-to-all traffic stays on NVLink instead of falling back to InfiniBand.

This optimization delivered a \textbf{20\% end-to-end throughput improvement} on GB200.

\subsubsection*{NUMA Binding for Policy and vLLM Workers}

On GB200 NVL72, each compute tray contains a Grace CPU with two sockets and multiple NUMA 
nodes. GPUs are physically affiliated with specific CPU sockets: GPUs~0 and~1 map to NUMA 
node~0, while GPUs~2 and~3 map to NUMA node~1. Without explicit NUMA binding, Ray worker 
processes, both Megatron policy workers and vLLM generation workers, could be scheduled on 
CPU cores belonging to the remote socket relative to their assigned GPUs. On 
Grace-Blackwell, the NVLink-C2C interconnect between Grace CPU and Blackwell GPU is 
socket-local, so cross-socket placement forces memory traffic to traverse the inter-socket 
coherence link before reaching NVLink-C2C, degrading GPU memory bandwidth.

The fix was to explicitly bind policy and vLLM worker processes to the CPU socket local to 
their assigned GPUs. This ensures that optimizer state offloading 
(e.g., CPU$\leftrightarrow$GPU transfers during checkpointing and weight refit), tokenization, 
data preprocessing, and pinned-memory allocations all hit local DRAM and use the 
socket-local C2C path. This optimization delivered a \textbf{10\% end-to-end throughput improvement} on GB200.

\subsubsection*{Checkpoint Save Blocking}
Synchronous checkpoint saves blocked training for $\sim$60s per save. Enabling asynchronous 
checkpointing from the Nvidia Resiliency Extension (NVRx), where model parameters are copied to CPU and persisted in the background 
while training continues, reduced the exposed blocking time to \textbf{$\sim$6--8 seconds}. 
Two additional optimizations reduced this further: overlapped NCCL transfers with D2H copies 
to shrink the synchronous staging window, and persistent checkpoint worker processes that avoid fork/spawn overhead on each save.  We further
moved checkpoint finalization, the cross-worker synchronization and the final
commit, onto a background thread so it never blocks training, and we cache the
distributed save plan so it is computed once rather than on every save. With the Megatron Core Distributed Optimizer, which shards optimizer state across 
data-parallel ranks so each rank only saves its local slice, the exposed save time reduces 
to \textbf{$<$1 second}.

\subsubsection*{JIT Cache and Initialization}

Cold-start initialization at 1K+ GPU scale took $\sim$49~minutes, of which $\sim$38.8~minutes 
was dominated by JIT compilation. Table~\ref{tab:jit-cache} shows the breakdown.

\begin{table}[ht!]
\centering
\caption{Cold-start vs.\ warm-start JIT compilation time benchmarked at 1K GPU scale.}
\label{tab:jit-cache}
\begin{tabular}{lcc}
\toprule
\textbf{Component} & \textbf{Cold Start} & \textbf{Warm Start} \\
\midrule
FlashInfer cubin compilation   & 28.0 min & 0 (cached) \\
Inductor / \texttt{torch.compile} & 5.5 min  & 0 (cached) \\
Triton kernel autotuning       & 2.0 min  & 0 (cached) \\
vLLM CUDA graph capture        & 2.5 min  & 0 (cached) \\
Model load                     & 0.4 min  & 0.4 min    \\
\midrule
\textbf{Total init/total}      & \textbf{38.8 min} & \textbf{0.4 min} \\
\bottomrule
\end{tabular}
\end{table}

Each of these frameworks maintains its own on-disk cache of compiled artifacts. On a cold 
start, when nodes are freshly allocated with empty local storage, every vLLM worker, 
policy worker, and judge model server independently JIT-compiles the same kernels from 
scratch. At scale, this means hundreds of redundant compilations 
executing in parallel, each taking seconds to minutes per kernel instance.

The fix was a three-pronged cache management strategy:

\begin{itemize}
  \item \textbf{Persistent warm cache on shared storage.} At job completion, all JIT 
    artifacts (Inductor graphs, Triton autotuned kernels, FlashInfer cubins, vLLM compiled 
    graphs) are written back to a persistent shared directory as compressed tarballs.

  \item \textbf{Node-local seeding at startup.} Before Ray initialization, a setup script 
    on each node extracts the warm cache tarballs into node-local \texttt{/tmp}, giving every 
    worker immediate access to pre-compiled artifacts without shared-filesystem contention. 
    All subsequent JIT writes target \texttt{/tmp} to avoid shared storage metadata storms from 
    parallel compilation.

  \item \textbf{Container-baked artifacts.} For FlashInfer specifically, precompiled cubins 
    were baked into the container image at build time via \texttt{flashinfer download-cubin}, 
    ensuring zero runtime compilation regardless of node allocation.
\end{itemize}

\noindent With warm caches, the init phase dropped from 38.8~minutes to 0.4~minutes, a 
99\% reduction, making initialization time negligible relative to step time.

\subsubsection*{Multi-Node vLLM Operational Stability}

Multi-node vLLM uses Ray as its distributed executor. Each vLLM data-parallel leader spawns 
\texttt{EngineCore} subprocesses that each call \texttt{ray.init()} to connect to the Global 
Control Store (GCS). At scale, this creates additional GCS connections on top of the actor 
creation burst. Beyond the JIT cache contention addressed in the previous section, three 
additional issues caused startup failures at scale:

\begin{itemize}
  \item \textbf{Package and kernel ABI mismatches.} Independent components of the RL stack 
    (policy training, generation, environment services) installed different versions of 
    shared GPU kernel libraries. ABI differences between versions meant precompiled kernels 
    from one component could not be reused by another, causing silent crashes at 
    initialization.

  \item \textbf{Subprocess environment divergence.} Environment variables set in the main 
    Ray actor process were not propagated to subprocesses spawned via 
    \texttt{multiprocessing.spawn}, causing them to resolve different library versions than 
    the parent and fail at kernel load time.

  \item \textbf{Collective-communication incompatibilities.} Certain JIT-compiled kernels 
    were incompatible with NCCL's multi-node NVLink memory registration on GB200, causing 
    hangs during distributed initialization.
\end{itemize}

\noindent These were addressed as follows:

\begin{itemize}
  \item \textbf{Dependency unification.} All components of the RL stack were aligned to a 
    single version of shared GPU kernel libraries, ensuring ABI consistency across the job.

  \item \textbf{Environment propagation.} Library paths and environment variables were 
    explicitly forwarded to all subprocesses at spawn time.

  \item \textbf{Collective-communication workaround.} Multi-node NVLink memory registration 
    was disabled for the affected kernel paths until an upstream fix in FlashInfer was available.

  \item \textbf{vLLM} Health checks to validate device allocations and communications over the TP NCCL group, RPC timeouts to prevent infinite waits on stalled workers, and graceful shutdown along with orphan process cleanup were all implemented to improve operational stability across jobs.

\end{itemize}

\subsubsection*{Container and Storage I/O}

At scale, every node simultaneously reads the container image ($\sim$44~GB squashfs) from 
shared storage at job startup, producing tens of terabytes of concurrent reads across 
thousands of parallel I/O streams. This overwhelmed the storage subsystem: a subset of nodes 
either hit \texttt{Input/output error} or stalled for 12+ minutes during container 
extraction (vs.\ $\sim$2 to 3~minutes under normal conditions). Because all workers must be 
ready before Ray initialization begins, a single slow node delays the entire job. 
Additionally, at job completion all nodes concurrently write back JIT caches to shared 
storage, creating a second I/O storm that degraded storage availability for other jobs on 
the cluster.

These were addressed with two complementary strategies:

\begin{itemize}
  \item \textbf{Container caching.} Enroot's local squashfs cache ensures that once a node 
    has extracted the container image, subsequent jobs reuse the cached local copy, 
    effectively eliminating the shared storage read load for warm nodes.

  \item \textbf{Asymmetric read/write cache paths.} During training, all JIT writes go to 
    node-local storage, producing zero shared storage writes. At job completion, a single 
    designated sidecar process archives the caches back to shared storage as compressed 
    tarballs, rather than all nodes writing simultaneously. Since all nodes compile identical 
    kernels, only one node's cache needs to be persisted. At startup, each node extracts 
    these tarballs into local storage via a single sequential read per cache type, avoiding 
    metadata-intensive small-file I/O entirely.
\end{itemize}

\subsubsection{Future Work}
Active work targets the two dominant failure categories: fail-fast fault isolation to prevent retry-and-cascade failure modes, enabling component-level recovery where individual generation workers or sandbox instances restart independently without full job restart. The sandbox and tool-calling infrastructure is being disaggregated to eliminate cascading failures and allow independent scaling. Fine-grained checkpointing of in-flight rollouts, KV cache, and conversation state further reduces recovery cost by enabling replay from the last consistent snapshot rather than from scratch.

\subsection{Post-trained Model Evaluations}
\label{subsec:final_model_evals}
\subsubsection{Evaluation Setup}

We evaluate \ourmodel on a comprehensive suite of benchmarks spanning agentic capabilities, reasoning and knowledge, conversational ability and instruction following, long-context understanding, and multilingual performance.
All evaluation results for \ourmodel and baselines were collected via Nemo Evaluator SDK\footnote{\url{https://github.com/NVIDIA-NeMo/Evaluator}}. We used three main evaluation harnesses: Nemo Gym\footnote{\url{https://github.com/NVIDIA-NeMo/Gym}}, Nemo Skills\footnote{\url{https://github.com/NVIDIA-NeMo/Skills}}, and Harbor\footnote{\url{https://github.com/harbor-framework/harbor}} with extended sandboxing support via AWS ECS on Nemo Evaluator. In addition, the evaluations also used dedicated open-source packaged containers for Multi-Challenge Multi-Turn Instruction Following. For reproducibility purposes, more details on the evaluation settings and pinned containers can be found in the Nemo Evaluator SDK examples folder\footnote{\url{https://github.com/NVIDIA-NeMo/Evaluator/blob/main/examples/nemotron/nemotron-3-ultra}} and the corresponding reproducibility tutorial.

The following benchmarks are not onboarded yet in our open source tools and for these we used either their official open source implementation or otherwise an internal scaffolding that we plan to open source in the future: BrowseComp, Tau Bench 3, ProfBench, PinchBench, Vals.ai Financial Agent, LongBench v2.

All models were evaluated under the same evaluation settings including agentic resources (CPU compute and timeouts), input data, prompt templates, number of repeats, and extraction/metric implementation. Inference parameters such as temperature and top\_p, and maximum number of tokens were directly extracted from the recommendations of the corresponding model cards. 

\paragraph{Agentic Capabilities.}
We evaluate agentic capabilities using Terminal-Bench 2.1~\citep{merrill2026terminal} for terminal-based task execution; GDPVal~\citep{patwardhan2025gdpval} for economically valuable real-world workplace tasks; SWE-Bench Verified~\citep{jimenez2023swe} and SWE-Bench Multilingual~\citep{jimenez2023swe} for software engineering agents; ProfBench~\citep{wang2026profbench} for professional-domain reasoning and deep-research tasks; PinchBench~\citep{pinchbench_skill2026} for coding agents in OpenClaw environments; TauBench V3~\citep{barres2025tau} for conversational tool use; BrowseComp~\citep{wei2025browsecomp} for web browsing and information-seeking tasks; and the Vals.ai Financial Agent Benchmark~\citep{bigeard2025finance} for financial analysis and decision-support tasks.

\paragraph{Reasoning and Knowledge.}
We evaluate reasoning and knowledge capabilities across coding, mathematics, scientific reasoning, and expert-level knowledge domains. For coding and algorithmic problem solving, we use International Olympiad in Informatics (IOI) 2025~\citep{ioi2025} and LiveCodeBench v6~\citep{jain2024livecodebench}. Mathematical reasoning is assessed using IMO-AnswerBench~\citep{luong2025towards} and MathArena Apex-Shortlist~\citep{dekoninck2026matharena}. For IMO-AnswerBench and Apex-Shortlist, we report both \emph{without-tools} and \emph{with-tools} settings, where the latter allows the model to use Python for computation and verification during problem solving. We further evaluate expert-level knowledge and reasoning using GPQA~\citep{rein2023gpqa}, MMLU-Pro~\citep{wang2024mmlupro}, and Humanity's Last Exam (HLE)~\citep{phan2025humanitysexam}; scientific coding with SciCode~\citep{tian2024scicoderesearchcodingbenchmark}; research-level physics reasoning with CritPt~\citep{zhu2025critpt}; and factual reliability and hallucination resistance with AA-Omniscience~\citep{jackson2025aaomniscience}. Together, these benchmarks provide a comprehensive assessment of the model's reasoning, problem-solving, and knowledge capabilities across a wide range of domains.

\paragraph{Conversation and Instruction Following.}
We assess conversational quality and instruction-following ability using IFBench~\citep{pyatkin2025generalizing}, and Multi-Challenge~\citep{deshpande2025multichallenge}. IFBench measures instruction-following accuracy under verifiable constraints. Multi-Challenge evaluates multi-turn conversational competence.

\paragraph{Long-Context Understanding.}
We evaluate long-context capabilities using AA-LCR~\citep{artificialanalysis2025lcr}, RULER~\citep{hsieh2024ruler}, and LongBench v2~\citep{bai2025longbenchv2}, which measure information retrieval, synthesis, and reasoning over long contexts of up to one million tokens.

\paragraph{Multilingual Performance.}
We evaluate multilingual capabilities using MMLU-ProX~\citep{xuan2025mmluprox} and WMT24++~\citep{deutsch2025wmt24pp} for multilingual reasoning, knowledge, and machine translation.

For more details, please refer to Appendix \ref{appendix:eval}.

\subsubsection{Evaluation Results}

\begin{table*}[!t]
\centering
\footnotesize
\setlength{\tabcolsep}{2.5pt}
\renewcommand{\arraystretch}{1.05}

\resizebox{\textwidth}{!}{%
\begin{tabular}{l|c c c c c c c}
\toprule
\textbf{Benchmark}
& \makecell{\textbf{N-3-Ultra}\\\textbf{550B-A55B}}
& \makecell{\textbf{MiniMax-2.7}\\\textbf{230B-A10B}}
& \makecell{\textbf{GLM-5.1}\\\textbf{744B-A40B}}
& \makecell{\textbf{Kimi-K2.6}\\\textbf{1T-A32B}}
& \makecell{\textbf{Qwen-3.5}\\\textbf{397B-17B}}
& \makecell{\textbf{DS-v4-Pro}\\\textbf{1.6T-A49B}}
& \makecell{\textbf{DS-v4-Flash}\\\textbf{284B-A13B}} \\
\midrule
\rowcolor{black!5}
\multicolumn{8}{l}{\textbf{Agentic}} \\
\midrule
Terminal Bench 2.1 & 56.4 & 55.5 & 59.3 & 67.2 & 49.9 & 49.2 & 54.2 \\
GDPVal & 46.7 & 47.6 & 54.7 & 50.4 & 34.6 & 54.6 & 50.2\\
SWE-Bench Verified & 70.7 & 75.3 & 76.2 & 75.7 & 73.6 & 74.5 & 73.5 \\
SWE-Bench Multilingual & 67.7 & 71.8 & 74.8 & 77.1 & 70.9 & 76.5 & 75.0 \\
ProfBench (Search) & 56.0 & 52.0 & 46.0 & 56.0 & 53.0 & 59.9 & 57.0 \\
PinchBench & 90.0 & 77.6 & 81.2 & 90.2 & 86.6 & 88.6 & 91.3 \\
TauBench V3 & & & & & & \\
\quad Airline & 81.5 & 75.3 & 85.0 & 85.8 & 76.5 & 80.8 & 80.8\\
\quad Retail & 86.4 & 84.9 & 84.1 & 82.9 & 88.5 & 88.9 & 89.1 \\
\quad Telecom & 92.9 & 89.6 & 96.9 & 97.8 & 98.0 & 96.3 & 98.3\\
\quad Banking & 22.6 & 14.6 & 12.8 & 23.1 & 20.9 & 25.9 & 26.7 \\
\quad Average & 70.9 & 66.1 & 69.7 & 72.4 & 71.0 & 73.2 & 73.7 \\
BrowseComp & 44.4 & 54.1 & 59.4 & 61.3 & 40.5 & 59.4 & 46.9  \\
Vals.ai Financial Agent 1.1 & & & & & & \\
\quad without web search & 60.1 & 51.3 & 60.2 & 54.0 & 61.3 & 58.9 & 58.4\\
\quad with web search & 53.7 & 50.5 & 60.7 & 58.8 & 59.0 & 62.3 & 60.1\\

\midrule
\rowcolor{black!5}
\multicolumn{8}{l}{\textbf{Reasoning and Knowledge}} \\
\midrule
IOI 2025 & 570.0 & -- & 456.5 & 585.0 & 441.3 & 580.1 & --\\
LiveCodeBench (v6) & 89.0 & 77.2 & 85.7 & 90.2 & 79.3 & 92.5 & 90.9 \\
IMOAnswerBench (no tools) & 88.6 & 68.3 & 86.8 & 91.1 & 83.1 & 93.0 & 91.1 \\
IMOAnswerBench (with tools) & 92.3 & 75.1 & 91.1 & 93.71 & 84.51 & 85.4 & 89.6 \\
Apex-Shortlist (no tools) & 74.9 & 28.9 & 71.1 & 77.4 & 61.4 & 85.8 & 82.4 \\
Apex-Shortlist (with tools) & 84.8 & 51.9 & 79.0 & 73.2 & 60.4 & 86.5 & 82.0 \\
GPQA (no tools) & 87.0 & 86.6 & 86.1 & 91.0 & 87.1 & 87.8 & 88.5 \\
SciCode (subtask) & 44.6 & 38.3 & 47.7 & 52.0 & 48.0 & 50.5 & 48.2 \\
HLE (no tools) & 26.7 & 23.1 & 27.2 & 34.8 & 28.5 & 37.7 & 32.2 \\
HLE (with tools) & 37.4 & -- & 50.4 & 54.0 & 48.3 & 48.2 & 45.1 \\
CritPt (no tools) & 3.1 & 0.6 & 3.7 & 9.1 & 2.4 & 14.0 & 10.6\\
MMLU-Pro & 86.8 & 81.9 & 85.9 & 88.1 & 88.3 & 87.5 & 86.4\\
OmniScience Accuracy & 24.1 & 20.5 & 31.3 & 35.5 & 35.9 & 46.8 & 39.9\\
OmniScience Non-Hallucination & 78.7 & 74.4 & 66.8 & 67.1 & 7.4 & 5.7 & 2.8\\

\midrule
\rowcolor{black!5}
\multicolumn{8}{l}{\textbf{Chat \& Instruction Following}} \\
\midrule
IFBench (prompt loose) & 81.7 & 74.6 & 76.6 & 73.7 & 78.2 & 79.1 & 82.0 \\
Multi-Challenge & 63.8 & 42.5 & 63.0 & 63.1 & 63.9 & 64.1 & 63.5\\
\midrule
\rowcolor{black!5}
\multicolumn{8}{l}{\textbf{Long Context}} \\
\midrule
AA-LCR & 65.4 & 69.8 & 66.9 & 70.2 & 68.3 & 67.3 & 62.7 \\
RULER (1M) & 94.7 & -- & -- & -- & 90.1 & 94.2 & 87.7 \\
Longbench v2 ($\leq$ 1M) & 61.9 & -- & -- & -- & 68.9 & 62.1 & 57.0 \\
\midrule
\rowcolor{black!5}
\multicolumn{8}{l}{\textbf{Multilingual}} \\
\midrule
MMLU-ProX (avg en/de/fr/es/it/ja/zh/hi/pt/ko) & 83.0 & 78.4 & 85.8 & 85.0 & 86.4 & 85.6 & 84.3\\
WMT24++ (en$\rightarrow$xx) & 83.7 & 82.8 & 84.4 & 84.5 & 86.8 & 85.9 & 85.9 \\

\bottomrule
\end{tabular}%
}

\caption{
Evaluation suite for \ourmodel. We compare against six open models.
}
\label{tab:ultra_comparison}
\end{table*}

Table~\ref{tab:ultra_comparison} shows that \ourmodel is a strong, agentic-first, and well-rounded post-trained model. It performs strongly across terminal-based task execution, productivity workflows, software engineering, real-world agent benchmarks, professional-domain deep research, and conversational tool use. Across the evaluation suite, \ourmodel remains competitive with leading open models, including several with substantially larger total parameter counts.

A key part of our evaluation protocol is the use of PinchBench and ProfBench as \textbf{held-out generalization gates}. These benchmarks were reserved for final validation: they were not used for training-time monitoring, checkpoint selection, or any other development decisions, and were evaluated only once after the final model was produced. This makes them a stringent test of whether improvements observed on development benchmarks transfer to unseen agentic settings. On PinchBench, \ourmodel achieves 90.0, within 1.3 points of the best model in the comparison and in the top tier of evaluated open models. On ProfBench, \ourmodel reaches 56.0, tying Kimi-K2.6, a 1T-parameter model, and remaining close to the strongest reported results. The strong performance on these held-out gates provides evidence that the gains achieved during development are not limited to benchmark suite, but generalize to unseen agentic tasks and environments.

Beyond agentic tasks, \ourmodel demonstrates strong chain-of-thought reasoning and tool-integrated reasoning capabilities. On IOI 2025, \ourmodel obtains a score of 570.0, corresponding to \textbf{top-3-human-level} competitive programming performance; this score would fall between the second- and third-ranked official human contestants on the IOI 2025 scoreboard.\footnote{\url{https://stats.ioinformatics.org/results/2025}} For mathematical reasoning, \ourmodel achieves 92.3 on IMOAnswerBench with tools, indicating that it can effectively combine internal reasoning with external computation and verification.

\ourmodel also maintains strong performance on conversational quality, long-context understanding, and reliability. For long-context understanding, \ourmodel supports contexts of up to 1M tokens and achieves competitive performance on million-token evaluation suites, demonstrating effective retrieval and context tracking at long sequence lengths. On AA-Omniscience, \ourmodel achieves the highest non-hallucination score, 78.7, suggesting a favorable reliability profile and a lower tendency to produce unsupported answers when knowledge is uncertain.

Overall, the results suggest that \ourmodel achieves its primary design goal of strong agentic capability while maintaining competitive reasoning, conversational quality, and long-context performance. In particular, its strong performance on held-out agentic evaluation gates provides evidence that the observed gains extend beyond benchmarks monitored during development. Despite being smaller than several leading open models, \ourmodel remains competitive across a broad range of challenging evaluations.

\begin{figure*}[!t]
\centering
\begin{minipage}[t]{0.45\textwidth}
\vspace{0pt}          
\centering
\footnotesize
\renewcommand{\arraystretch}{1.1}
\begin{tabular}{l|c}
\toprule
\textbf{Competition} & \textbf{Accuracy} \\
\midrule
IMO-ProofBench Advanced & 82.3\% (173/210) \\
IMO 2025                & 83.3\% (35/42) \\
Putnam 2025             & 96.7\% (116/120) \\
USAMO 2026              & 97.6\% (41/42) \\
\bottomrule
\end{tabular}
\captionof{table}{Test-time scaling results for \ourmodel on Olympiad-level competition mathematics. Accuracy is reported with the corresponding graded score in parentheses. Scores are from human expert graders, except USAMO 2026, which follows \citet{dekoninck2026matharena}.}
\label{tab:math_ttc}
\end{minipage}
\hfill
\begin{minipage}[t]{0.52\textwidth}
\vspace{0pt}          
\centering
\includegraphics[width=\linewidth]{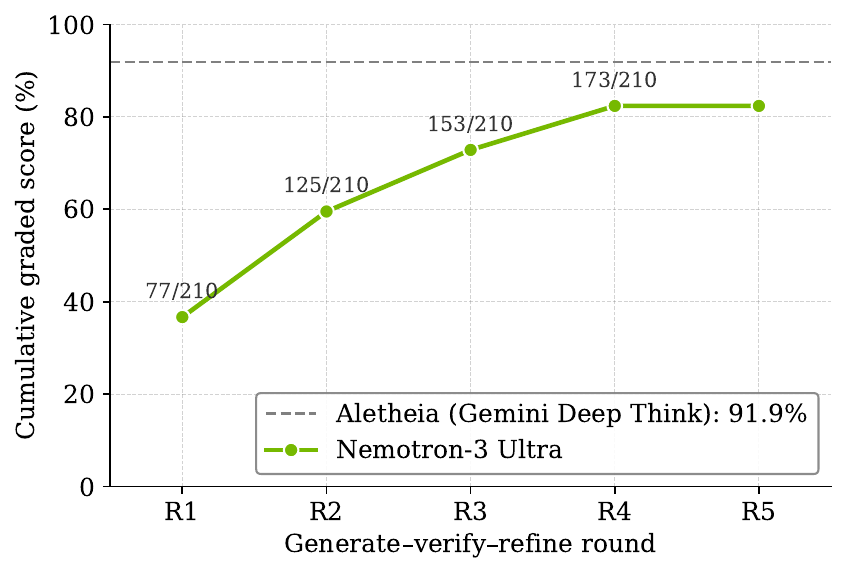}
\captionof{figure}{Test-time scaling of \ourmodel on IMO-ProofBench Advanced by round}
\label{fig:math_ttc}
\end{minipage}
\end{figure*}

\subsubsection{Test-time Scaling in Math Olympiad Problems}

\ourmodel achieves strong performance as the underlying model of a high-compute, search-based test-time scaling strategy for Olympiad-level competition mathematics. Following the generate--verify--refine methodology of \citet{shao2025deepseekmath}\footnote{In contrast with the original paper, we started with 128 proof attempts per problem and allowed 512k context length. All other pipeline hyperparameters were identical.} and Nemotron-Cascade-2~\citep{yang2026nemotron}, we evaluate on the IMO-ProofBench Advanced subset \citep{luong2025towards}, IMO 2025, Putnam 2025, and USAMO 2026, with final accuracies reported in Table~\ref{tab:math_ttc}.

Figure~\ref{fig:math_ttc} shows the accuracy of the test-time scaling strategy vs rounds of refinement in the pipeline, as a proxy for compute. The SOTA performance of the Aletheia math research agent \citep{feng2026autonomousmathematicsresearch} is also plotted.


\clearpage
\section{Quantization}
\label{sec:quantization}

We apply post-training quantization (PTQ) using Model-Optimizer to quantize the Nemotron 3 Ultra checkpoint to NVFP4 \citep{nvpfp4_ptq_default} for efficient inference on NVIDIA Blackwell GPUs.
The quantization format per operator (GEMM, KVCache, Mamba Cache) is summarized in Table~\ref{tab:quant-recipe}. We begin from a heuristic mixed per-layer precision recipe informed by
Model-Optimizer AutoQuantize\footnote{\url{https://github.com/NVIDIA/Model-Optimizer/tree/main/examples/llm_ptq\#autoquantize}}
sensitivity analysis on Nemotron 3 Super~\citep{nvidia2026nemotron3superopen}.
We then perform ablations over the model's effective
bits per element (BPE) and FP4 weight quantization algorithms to refine this
recipe and select the final operating point.

\begin{table}[!htp]
\centering
\small
\setlength{\tabcolsep}{8pt}
\renewcommand{\arraystretch}{1.3}
\resizebox{\textwidth}{!}{
\begin{tabular}{@{}lcc@{}}
\toprule
\textbf{Layer / Operator} & \textbf{BF16 Baseline} &
\textbf{Quantized Checkpoint Precision} \\
\midrule
Embedding, Output classification layer, MTP layers & BF16 & BF16 \\
MoE routed experts & BF16 & NVFP4 \\
MoE shared experts & BF16 & FP8 per-tensor \\
Mamba mixer linears & BF16 & FP8 per-tensor \\
Attention linears & BF16 & BF16 \\
Latent MoE & BF16 & BF16 \\
Mamba conv1d & BF16 & BF16 \\
KV cache & BF16 & FP8 \\
Mamba SSM cache & FP32 & FP16 with stochastic rounding \\
\bottomrule
\end{tabular}
}
\caption{
Quantization recipe for \ourmodel, mapping each layer or operator
from its BF16 baseline.
}
\label{tab:quant-recipe}
\end{table}

\subsection{Bits per Element Selection}
\label{sec:bpe-sweep}

We selected the model's bits-per-element (BPE) budget empirically, by quantizing a
fixed intermediate checkpoint at a range of BPE settings and scoring each against a
curated suite of evaluations, served through the Nemo Evaluator SDK on vLLM~v0.20.0.
We raised the number of repeats per benchmark to suppress run-to-run variance and
report the averaged pass@1 scores. The suite spans seven benchmarks: coding (SciCode),
scientific reasoning (GPQA~Diamond, HLE, CritPt), instruction following and knowledge
(IFBench, AA-Omniscience), and long-context reasoning (AA-LCR).

The results are summarized in Table~\ref{tab:bpe-sweep}.
Here BPE is best read as a
summary axis over a family of quantization recipes rather than a single tunable knob:
the higher-BPE points correspond to qualitatively different strategies (router-only
quantization, keeping the last MoE layer in BF16, and skipping the top-8 most
sensitive experts), each of which lands at a different effective BPE.

Across the sweep, most capabilities are already saturated at the lowest BPE we tried.
GPQA Diamond, SciCode, HLE,
IFBench, and AA-Omniscience accuracy are flat to within
run-to-run noise across the entire 4.85--7.19 BPE range. The single discriminating
axis is long-context reasoning: AA-LCR improves by $+2.4$ points
at the 4.85~$\rightarrow$~5.03 step and then plateaus (64.2--65.0) all the way
to 7.19~BPE. This step is precisely the introduction of the mixed-FP8 layers on top of
the NVFP4-amax recipe, so the long-context recovery is attributable to those targeted
higher-precision layers rather than to additional bits in general.

Above 5.03~BPE we observe no further gains: increasing the budget to 7.19~BPE
(a 43\% increase in bits) leaves every benchmark unchanged within noise. We therefore
selected \textbf{5.03~BPE} (NVFP4 with mixed-FP8) as the operating point, as it is the
smallest budget that recovers long-context performance while leaving no measurable
quality on the table at higher precision.

We note two caveats in reading Table~\ref{tab:bpe-sweep}. First, CritPt scores sit near
the floor of the benchmark (${\sim}3$--$5\%$) and are non-monotonic across the sweep; we
treat it as noise-dominated rather than a deciding signal here, which is also why the
repeat count was raised. Second, AA-Omniscience's non-hallucination rate mildly favors
the lowest BPE (54.13 at 4.85 versus 51.59 at 5.03); given the magnitude relative to the
other axes we attribute this to variance rather than a genuine precision trade-off.
\label{app:bpe-sweep}

\begin{table}[ht!]
\centering
\small
\setlength{\tabcolsep}{6pt}
\renewcommand{\arraystretch}{1.3}

\begin{tabular}{lr|c >{\columncolor{blue!8}}c c c c}
\toprule
& & \multicolumn{5}{c}{\textbf{Quantization (bits-per-element)}} \\
\cmidrule(lr){3-7}
\textbf{Task} & \textbf{Metric} &
4.85 & \textbf{5.03}$^{\dagger}$ & 5.25 & 5.43 & 7.19 \\
\midrule

\rowcolor{black!5}
\multicolumn{7}{l}{\textbf{Coding}} \\
SciCode & \textit{pass@1 (avg-16), subtask acc} & 43.82 & \textbf{43.88} & 43.45 & 43.27 & 43.44 \\
\midrule

\rowcolor{black!5}
\multicolumn{7}{l}{\textbf{Scientific Reasoning}} \\
GPQA Diamond & \textit{pass@1 (avg-32), sym.\ correct} & 84.66 & 84.33 & \textbf{84.75} & 84.12 & 84.52 \\
HLE & \textit{pass@1, judge correct} & 24.24 & 24.84 & 25.00 & 24.98 & \textbf{25.44} \\
CritPt & \textit{pass@1 (avg-8), accuracy} & 3.04 & 3.93 & \textbf{5.18} & 4.82 & 4.46 \\
\midrule

\rowcolor{black!5}
\multicolumn{7}{l}{\textbf{General}} \\
AA-Omniscience & \textit{pass@1 (avg-20), judge correct} & 29.21 & \textbf{29.75} & 29.18 & 29.29 & 29.00 \\
          & \textit{pass@1 (avg-20), non-hallucination} & \textbf{54.13} & 51.59 & 51.84 & 51.70 & 52.81 \\
IFBench & \textit{pass@1 (avg-8), avg.\ score} & 79.34 & 79.26 & \textbf{79.83} & 79.53 & \textbf{79.83} \\
\midrule

\rowcolor{black!5}
\multicolumn{7}{l}{\textbf{Long Context}} \\
AA-LCR & \textit{pass@1 (avg-16), judge correct} & 62.25 & \textbf{64.69} & 64.19 & 64.94 & \textbf{65.00} \\
\bottomrule
\end{tabular}
\caption[Bits-per-element sweep]{
  Bits-per-element (BPE) sweep on a fixed intermediate checkpoint, evaluated with the
  Nemo Evaluator SDK served on vLLM~v0.20.0. Columns are quantization configurations
  at increasing effective BPE; the specific recipes are described in
  Section~\ref{sec:bpe-sweep}. Rows
  are averaged pass@1 scores, and the best result per row is in bold. The shaded
  \textbf{5.03} column ($^{\dagger}$NVFP4 with mixed-FP8) is the selected operating
  point: it is the smallest budget that recovers long-context (AA-LCR) performance,
  while no benchmark improves beyond run-to-run noise at higher BPE.
}
\label{tab:bpe-sweep}
\end{table}

\subsection{FP4 Algorithm Experiments}

We explored FP4 PTQ algorithms focused on scale selection on BPE settings around 5.03 to study the impact of different algorithms on different quantization configurations. For FP4 input activations, we follow the default NVFP4 PTQ
recipe~\citep{nvpfp4_ptq_default}, using max-based scaling selected from
calibration statistics. We then vary only the FP4 weight scale-selection rule.
Because this only changes offline weight scales computed during PTQ, no
additional inference kernel support is required; the checkpoint remains on the
standard NVFP4 inference path. In mixed-precision settings we ablate over different algorithms for NVFP4 weights but keep FP8 weights using max-based scaling. 

For weights, we experimented with max-based, MSE-based, and Four-Over-Six ~\citep{four_over_six} scaling. 

Each algorithm changes the per-block NVFP4 scale -- max-based scaling uses
the block absolute maximum; MSE-based scaling minimizes reconstruction
error~\citep{nvidia2026nemotron3superopen}; and
Four-Over-Six selects between the $M=4$ and $M=6$
weight grids per block, using the option that minimizes reconstruction error.
Table~\ref{tab:fp4-weight-calib} compares these weight scale-selection
strategies across mixed-precision settings and BPE targets. We find that in more conservative quantization settings of 5.03 or 5.43 BPE, MSE slightly outperforms max, whereas in the aggressive quantization setting of 4.85 BPE, they are comparable. However, Four-Over-Six shows an improvement in the balanced 5.03 BPE setting with large degradation in the 4.85 BPE setting. The degradation in 4.85 BPE could be due to mamba linear layers being sensitive to outliers and preferring max scales.

Four-Over-Six increases the global per-tensor weight scale by $1.75\times$ and
allows each weight microblock to choose between the $M=4$ and $M=6$ FP4 grids,
where $M=4$ uses a $1.5\times$ larger block scale than $M=6$.
This trades a small amount of additional zero-rounding for better handling of
high-magnitude tails, similar in spirit to MSE calibration. In full routed-expert
tensor analysis, max-calibrated Four-Over-Six reduced the median relative MSE of
quantized weight reconstruction by 16.4\% compared with standard max
calibration, with improvement across all 49,152 projection weights from 48 MoE
expert layers. Although MSE calibration gave an additional weight-MSE reduction of 27.1\%,
downstream evaluations did not show a consistent accuracy increase across
benchmarks. We therefore selected Four-Over-Six for setting the FP4 routed-expert
weight scales in the mixed precision 5.03 BPE setting.

\begin{table*}[!ht]
\centering
\footnotesize
\setlength{\tabcolsep}{4pt}
\renewcommand{\arraystretch}{1.4}
\begin{tabular}{@{}p{0.3\textwidth} | *{3}{>{\raggedright\arraybackslash}p{0.18\textwidth}}@{}}
\toprule
\textbf{BPE} &
\makecell[l]{\textbf{Max per-block}} &
\makecell[l]{\textbf{MSE per-block}} &
\makecell[l]{\textbf{Four-Over-Six}\\\textbf{per-block}} \\
\midrule
5.43 (NVFP4 Routed Experts only) & 97.44 & 98.27 & n/a \\
5.03 (NVFP4 Routed Experts + mixed FP8) & 96.78 & 98.40 & \textbf{98.50} \\
4.85 (NVFP4 experts + Mamba) & 98.32 & 97.57 & 84.71 \\
\bottomrule
\end{tabular}
\caption{
Summary of FP4 weight scale-selection ablations showing median accuracy recovery
on 6 AA benchmarks (GPQA, SciCode, HLE AA, IFBench, CritPT, Omniscience)
relative to BF16 on an \textbf{intermediate checkpoint}. Activation scales use
max calibration in all columns and SSM Cache used FP32.
}
\label{tab:fp4-weight-calib}
\end{table*}

\subsection{Final Weight and GEMM Quantization Recipe}

Overall, the final GEMM and MoE PTQ recipe combines:
\begin{itemize}
    \item NVFP4 routed-expert GEMMs; Dynamic max-based activation scaling and
    max-calibrated 4/6 weight scaling.
    \item FP8 per-tensor GEMMs for shared experts and Mamba linear layers;
    static max-calibrated per-tensor scales.
    \item BF16 precision for attention linear layers, MoE latent projection layers (i.e, layers for which accuracy degradation outweighed the expected
    inference-cost benefit after quantization).
\end{itemize}
This combination reduces the accuracy loss of naive NVFP4 PTQ while preserving the
runtime efficiency needed for deployment. The resulting checkpoint operates at
5.03 bits-per-element (BPE); the BPE selection is detailed in
Section~\ref{sec:bpe-sweep}. The final per-operator precision assignments are
summarized in Table~\ref{tab:quant-recipe}.


\subsection{Software Support in Model-Optimizer }
 While Model-Optimizer supports both HuggingFace and Megatron-LM for quantization, due to Ultra’s large size we chose Megatron-LM for its efficient multi-node distributed parallelism and MoE support  ~\citep{yan2026scalabletrainingmixtureofexpertsmodels}. 

Megatron-LM supports multi-node inference and n-D parallelism which allow us to shard the model across multiple GPUs and across nodes, making quantization much faster. Megatron-LM’s expert parallelism and data parallelism speed up calibration by sharding experts across GPUs and enabling large global batch sizes by parallelizing model forward. In addition, context parallelism allowed us to scale to large sequence lengths beyond 32k tokens, which can help improve calibration accuracy. A comparison against HuggingFace transformers showed that Megatron-LM allowed us to experiment faster and perform more complex long-sequence experiments on multi-node compute (Table~\ref{tab:framework-comparison}). 

In contrast, Model-Optimizer's HuggingFace PTQ script uses transformers native inference, which does not support multi-node due to being limited to single process execution. Ultra’s 550B parameter size means the BF16 model is approximately 1.1TB and cannot comfortably fit inside one node, requiring CPU offloading inactive layers for inference in transformers. We used transformers on a single node with CPU offloading to quantize Ultra layer-by-layer.

Depending on your individual resources and preference, both HuggingFace \footnote{https://github.com/NVIDIA/Model-Optimizer/tree/main/examples/llm\_ptq} and Megatron-LM \footnote{https://github.com/NVIDIA/Megatron-LM/tree/main/examples/post\_training/modelopt} are suitable frameworks for quantizing large models with HuggingFace layerwise PTQ taking around 2 hours vs Megatron-LM taking 42 minutes on Ultra. 

\begin{table*}[!ht]
\centering
\resizebox{\textwidth}{!}{%
\small
\renewcommand{\arraystretch}{1.3}
\begin{tabular}{@{} l p{0.32\textwidth} p{0.38\textwidth} @{}}
\toprule
\textbf{Metric} & \textbf{HuggingFace transformers} & \textbf{Megatron-LM} \\
\midrule
Compute & 4 x B300 & 16 x B300s; \newline Expert Parallelism = \newline Data Parallelism = 16 \\
Model loading time & 40 minutes & < 2 minutes \\
Model loading \& Calibration time & 85 minutes & 9 minutes \\
Export & 42 min & 33 min \\
\midrule
\textbf{Total Time} & \textbf{2 hours} & \textbf{45 minutes} \\
\bottomrule
\end{tabular}%
}
\caption{PTQ performance comparison on HuggingFace transformers vs. Megatron-LM with details on compute setups, loading, calibration, and export times.}
\label{tab:framework-comparison}
\end{table*}

\subsection{SSM Cache Optimization}
\label{sec:ssm-cache-optimization}

\begin{figure}[t]
\centering
\includegraphics[width=0.72\linewidth]{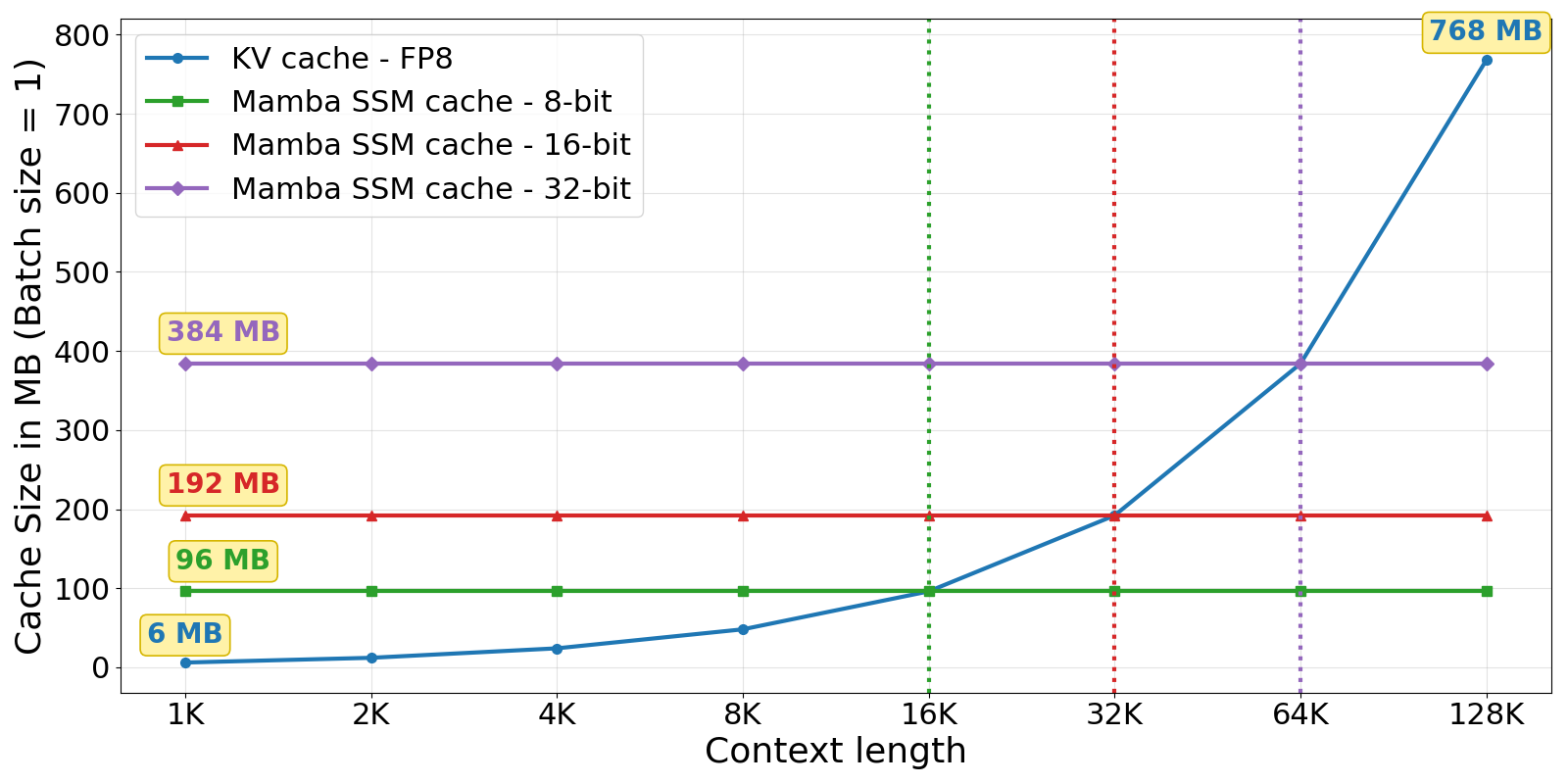}
\caption{Cache size comparison for FP8 KV cache and Mamba SSM cache at
different cache precisions for batch size 1.}
\label{fig:mamba-vs-kv}
\end{figure}

In Mamba autoregressive decoding, the SSM state is stored in a
constant-sized cache for each batch element. The KV
cache size, on the other hand, grows with sequence length.
However, there is a crossover point below which the Mamba cache size is larger than the KV cache size.
For example, in Nemotron 3 Ultra, the 32-bit Mamba cache is larger than the FP8 KV cache at sequence lengths up to 64K. This is illustrated in Figure~\ref{fig:mamba-vs-kv}. 

Therefore, depending on the sequence length and batch size, the Mamba cache can become both a DRAM-footprint bottleneck and a source of DRAM-read pressure that limits decoding speed.

To address this, we quantize the Mamba SSM cache from the original FP32 precision to lower precisions.
Following Nemotron 3 Super~\citep{nvidia2026nemotron3superopen}, we first adopt 16-bit cache precision, where FP16 with stochastic rounding preserves FP32-cache accuracy and verbosity.

To further improve cache compression, we also explore 8-bit Mamba cache quantization formats.
We evaluated this on Nemotron 3 Super.

Consistent with Nemotron 3 Super~\citep{nvidia2026nemotron3superopen}, we find that the mantissa precision and stochastic rounding are key ingredients for preserving accuracy after Mamba cache quantization.
Block-scaled INT8 quantization with stochastic rounding largely preserves FP32-cache accuracy and verbosity. FP8 E4M3 quantization degrades accuracy, likely because it provides lower effective precision for these cache values than INT8.
The Mamba cache quantization results are summarized in Table~\ref{tab:mamba-cache-checkpointing}.

To further reduce the quantization error caused by Mamba SSM state caching, we explore the idea of periodic cache checkpointing: Instead of storing (thus, quantizing) the state in each decoding step, we store it every $CC$ steps for an integer $CC>1$. To compensate for the state being behind, we cache the input activations and apply activation replay to forward the state on-the-fly. This reduces the number of sequential quantization steps by a factor of $CC$ and can also save time due to saving on cache writes (though there is a tradeoff with the additional compute and IO used for the activation replay).

We ran initial experiments on Nemotron 3 Super NVFP4, using emulated quantization. The results are summarized in Table~\ref{tab:mamba-cache-checkpointing}.

\begin{table}[!htp]
\centering
\small
\setlength{\tabcolsep}{5pt}
\renewcommand{\arraystretch}{1.2}
\resizebox{\textwidth}{!}{%
\begin{tabular}{@{}>{\raggedright\arraybackslash}p{0.30\textwidth}
                  *{4}{>{\centering\arraybackslash}p{0.175\textwidth}}@{}}
\toprule
\textbf{Mamba cache precision} &
\multicolumn{2}{c}{\textbf{Quantization (No Checkpointing)}} &
\multicolumn{2}{c}{\textbf{Quantization + Checkpointing (CC = 8)}} \\
\cmidrule(lr){2-3}
\cmidrule(lr){4-5}
&
\textbf{Mean accuracy drop from FP32 cache} &
\textbf{Average verbosity increase from FP32 cache} &
\textbf{Mean accuracy drop from FP32 cache} &
\textbf{Average verbosity increase from FP32 cache} \\
\midrule
FP16 RTN & 1.07\% & 9.98\% & 0.03\% & 3.07\% \\
FP16 SR & -0.26\% & 0.36\% & 0.19\% & 0.40\% \\
INT8 RTN, block size = 128 & 1.42\% & 9.91\% & 0.30\% & 4.16\% \\
INT8 SR, block size = 128 & -0.15\% & 1.28\% & -0.42\% & 0.60\% \\
FP8 RTN, block size = 128 & 4.68\% & 25.17\% & 0.46\% & 3.69\% \\
FP8 SR, block size = 128 & 0.61\% & 3.12\% & 0.29\% & 1.41\% \\
\bottomrule
\end{tabular}
}
\caption{
Mamba cache checkpointing results on the Nemotron 3 Super NVFP4 model with emulated quantization, measured relative to the same model with FP32 Mamba cache. CC denotes the checkpointing period. Accuracy and verbosity are measured based on the following evaluation benchmarks: MMLU Pro, GPQA, HLE, LiveCodeBench, IFBench, OmniScience, AA-LCR, Ruler 128K, and Ruler 256K. For both metrics, lower is better.
}
\label{tab:mamba-cache-checkpointing}
\end{table}

Optimized 8-bit Mamba cache kernels with checkpointing support are under development.
For the current release, we use FP16 SSM cache storage with stochastic rounding, as described in Nemotron Super~\citep{nvidia2026nemotron3superopen}, which avoids block-scaling overhead while mitigating the recurrent rounding bias observed with naive FP16 round-to-nearest cache storage.

\subsection{One NVFP4 checkpoint}

We release a single NVFP4 checkpoint for \ourmodel{} that targets both
Blackwell, where it runs with native FP4 math, and Hopper, where it runs
as W4A16 (weights NVFP4, activations BF16). This is because Hopper lacks native FP4
tensor cores. KV Cache is quantized to FP8 and Mamba cache data type is FP16 with stochastic rounding.

The natural candidate for Hopper would have been a separate FP8 checkpoint to un-lock FP8 tensor-core math.
At first glance W8A8 should be the better Hopper choice: FP8 tensor
cores have higher peak throughput than the BF16 math the W4A16 path
relies on. In practice the opposite holds at \ourmodel's scale. With
TP=8 on an 8-GPU H100 node ($640$~GiB of aggregate HBM), the FP8
checkpoint ($\approx 540$~GiB) leaves roughly $10$~GiB per GPU for
activations, KV cache, and Mamba state, against roughly $40$~GiB for
the NVFP4 checkpoint ($\approx 330$~GiB). The tight FP8 cache budget
caps the maximum batch size at the operating points we care about,
which keeps the workload memory-bandwidth-bound and prevents us from
ever reaching the compute-bound regime where FP8 tensor cores would
matter. The measured throughput-versus-user-latency Pareto places
W4A16 on or above W8A8 across the relevant range.

The case for the single NVFP4 checkpoint becomes much stronger once
MTP is in the picture. The reduced W4 footprint leaves enough headroom
to fit MTP weights on the same 8-GPU node, while the FP8 checkpoint
cannot fit MTP without scaling to two H100 nodes and surrendering the
single-NVLink-domain property. With MTP enabled on the W4A16 path, the
throughput-versus-user-latency Pareto both moves up across the relevant
range and extends much further into the low-latency regime than W8A8
reaches at any operating point.

A third option, W4A8, NVFP4 weights with FP8 activations, would
in principle combine the smaller weight footprint with FP8 tensor-core
math. We chose not to ship it. Naively downcasting NVFP4 weights to FP8
causes catastrophic accuracy degradation: NVFP4 uses E2M1 elements with
E4M3 block scales, giving an effective E6M4 representation whose range
exceeds FP8's E4M3, so the resulting weights saturate. Preserving
accuracy therefore requires a W4 $\rightarrow$ BF16 $\rightarrow$ FP8
round-trip rather than a direct W4 $\rightarrow$ FP8 cast, which adds
an extra cast op before each GEMM. Combined with the result above, 
even pure W8A8 is slower than W4A16 in our regime, the extra op
makes W4A8 strictly worse than W4A16, with no accuracy upside.

\begin{table}[ht!]
\centering
\small
\setlength{\tabcolsep}{5pt}
\renewcommand{\arraystretch}{1.3}

\begin{tabular}{l|cc|cc|cc}
\toprule
& \multicolumn{2}{c|}{\textbf{BF16}}
& \multicolumn{2}{c|}{\textbf{W4A16}}
& \multicolumn{2}{c}{\textbf{NVFP4 (W4A4)}} \\
\cmidrule(lr){2-3} \cmidrule(lr){4-5} \cmidrule(lr){6-7}
\textbf{Task} & \textbf{Score} & \textbf{Tok.}
            & \textbf{Score} & \textbf{Tok.}
            & \textbf{Score} & \textbf{Tok.} \\
\midrule
GPQA           & \textbf{86.67} & 14408 & \textbf{86.67} & 14841 & 86.36          & 15134 \\
HLE            & \textbf{26.92} & 37479 & 25.12          & 39103 & 25.67          & 39103 \\
IFBench        & 82.12          & 5272  & \textbf{82.75} & 5466  & 82.42          & 5764  \\
AA-Omniscience & 24.38          & 1140  & \textbf{25.50} & 1396  & 24.55          & 1226  \\
AA-LCR         & 63.67          & 4110  & \textbf{65.33} & 4286  & 64.00          & 4346  \\
\bottomrule
\end{tabular}
\caption[BF16 vs.\ W4A16 vs.\ NVFP4]{
  Accuracy and verbosity of the single NVFP4 checkpoint deployed as W4A16 (NVFP4
  weights, BF16 activations) and as NVFP4 (W4A4, native FP4; the
  Blackwell path), against the BF16 reference. \textbf{Score} is pass@1; \textbf{Tok.}\
  is the average completion length in tokens. Higher score per task is in bold.
  W4A16 outperforms NVFP4 on four of five tasks (all but HLE), stays within $1$~point
  of or above BF16 on four of five tasks (all but HLE), and produces no more completion
  tokens than NVFP4 on four of five tasks (all but AA-Omniscience).
}
\label{tab:w4a16-vs-nvfp4}
\end{table}

\begin{table*}[!htp]
\centering
\footnotesize
\setlength{\tabcolsep}{2.5pt}
\renewcommand{\arraystretch}{1.05}
\begin{tabular*}{\textwidth}{@{\extracolsep{\fill}}l|c c }
\toprule
\textbf{Benchmark}
& \makecell{\textbf{N-3-Ultra}\\\textbf{BF16}}
& \makecell{\textbf{N-3-Ultra}\\\textbf{NVFP4}}\\

\midrule
\rowcolor{black!5}
\multicolumn{3}{l}{\textbf{Agentic}} \\
\midrule
Terminal Bench 2.1 & 56.4 &  53.9 \\
GDPVal & 46.7 & 47.9   \\
SWE-Bench Verified & 70.7 & 69.5  \\
SWE-Bench Multilingual & 67.7 & 69.1  \\
ProfBench (Search) & 56 & 56.4 \\
PinchBench & 90 & 89.8  \\
TauBench V3 & &   \\
\quad Airline & 81.5 & 80.0 \\
\quad Retail  & 86.4 & 88.4 \\
\quad Telecom & 92.9 & 93.6 \\
\quad Banking & 22.6 & 19.2 \\
\quad Average & 70.9 & 70.3 \\
BrowseComp & 44.4 & 41.4 \\

\midrule
\rowcolor{black!5}
\multicolumn{3}{l}{\textbf{Reasoning and Knowledge}} \\
\midrule
IOI 2025 & 570.0 & 564.7 \\
GPQA (no tools) & 87.0 &  87.9 \\
SciCode (subtask) & 44.6 & 43.5   \\
HLE (no tools) & 26.7 & 26.1   \\
CritPt (no tools) & 3.1 & 3.4  \\
OmniScience Accuracy & 24.1 & 24.6  \\
OmniScience Non-Hallucination & 78.7 & 75.5 \\

\midrule
\rowcolor{black!5}
\multicolumn{3}{l}{\textbf{Chat \& Instruction Following}} \\
\midrule
IFBench (prompt) & 81.7 & 82.3  \\

\midrule
\rowcolor{black!5}
\multicolumn{3}{l}{\textbf{Long Context}} \\
\midrule
AA-LCR & 65.4 & 65.5  \\
RULER 1M & 94.7 & 94.0  \\


\bottomrule
\end{tabular*}

\caption{
Evaluation suite comparing BF16 and NVFP4 of \ourmodel. BF16 uses vLLM 0.17.1, NVFP4 uses vLLM 0.22.0.
}
\label{tab:super_comparison}
\end{table*}

\clearpage
\section{Inference}
\label{sec:inference}

\ourmodel inherits the inference-aware architecture of
\supermodel~\citep{nvidia2026nemotron3superopen}: LatentMoE~\citep{latentmoe_tr},
which buys more routed experts at fixed inference cost by trading away
hidden-dimension width; a hybrid Mamba-2 stack with sparse global Attention
anchors, which gives sub-quadratic sequence-length scaling during prefill
and a bounded KV-cache footprint during decode; and Multi-Token Prediction
for native speculative decoding (see \S\ref{sec:pretraining}). Below we first
characterize how these choices play out in practice across serving regimes ---
prefill- versus decode-heavy workloads and small- versus large-batch operation
--- and then turn to the considerations that come into focus when serving a
model of this size, considerations that did not bind at \supermodel{} or
\nanomodel{} scale.
Quantization recipes and the SSM-state cache treatment are covered in
\S\ref{sec:quantization}; headline throughput comparisons appear in
\S\ref{sec:intro}.

\subsection{Performance across serving regimes}
\label{sec:inference:regimes}
\ourmodel{}'s effective inference performance depends strongly on the
operating point: the prefill/decode balance of the workload and the batch
size at which it is served. We characterize both below.

\begin{figure}[t]
    \centering
    \includegraphics[width=0.7\linewidth]{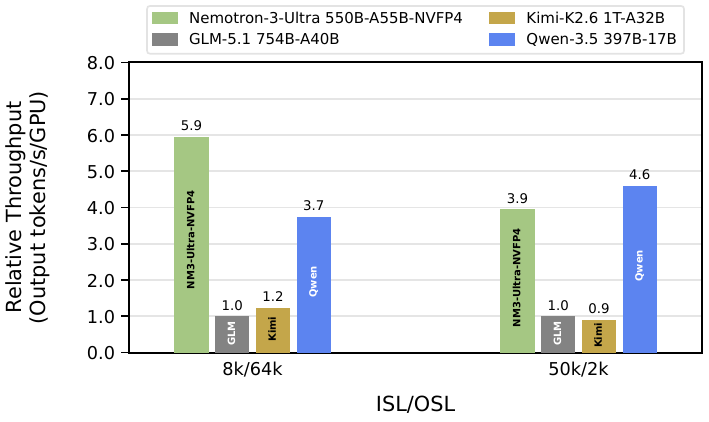}
    \caption{Relative throughput on a decode-heavy (8K/64K) and a
    prefill-heavy (50K/2K) ISL/OSL setting, normalized to GLM-5.1 in both
    settings. The decode-heavy setting and measurement methodology match
    Figure~\ref{fig:intro}. \ourmodel{} leads on the decode-heavy setting
    ($1.6\times$ over Qwen-3.5) but trails Qwen-3.5 on the prefill-heavy
    setting, consistent with the active-parameter (prefill) versus
    total-weight-I/O (decode) analysis in the text.}
    \label{fig:throughput_workloads}
\end{figure}

\paragraph{Throughput across prefill- and decode-heavy workloads.}
\label{sec:inference:workloads}
Figure~\ref{fig:throughput_workloads} compares max-throughput serving on
GB200 NVL72 for two representative settings:
a decode-heavy 8K~input / 64K~output workload and a prefill-heavy
50K~input / 2K~output workload, both at NVFP4 precision with speculative
decoding disabled.
Prefill is compute-bound, so per-token cost tracks FLOPs, which are set by
the number of active parameters; \ourmodel{} pays roughly a $3.2\times$ FLOPs
penalty against Qwen-3.5-397B-17B (55B versus 17B active parameters), so the
MoE GEMMs are the dominant bottleneck and \ourmodel{} trails on the
prefill-heavy workload.
In large-batch decode, routing activates essentially all experts and per-step
cost is instead set by total weight I/O, where the gap shrinks to roughly
$1.39\times$ (550B versus 397B total parameters). With the MoE penalty that
much smaller, the token-mixing mechanism becomes the deciding factor, and
\ourmodel{}'s Mamba-2 state-space layers, whose per-step decode cost is
constant in sequence length, let it lead on the decode-heavy workload despite
trailing on prefill.

\begin{figure}[h]
    \centering
    \includegraphics[width=0.55\linewidth]{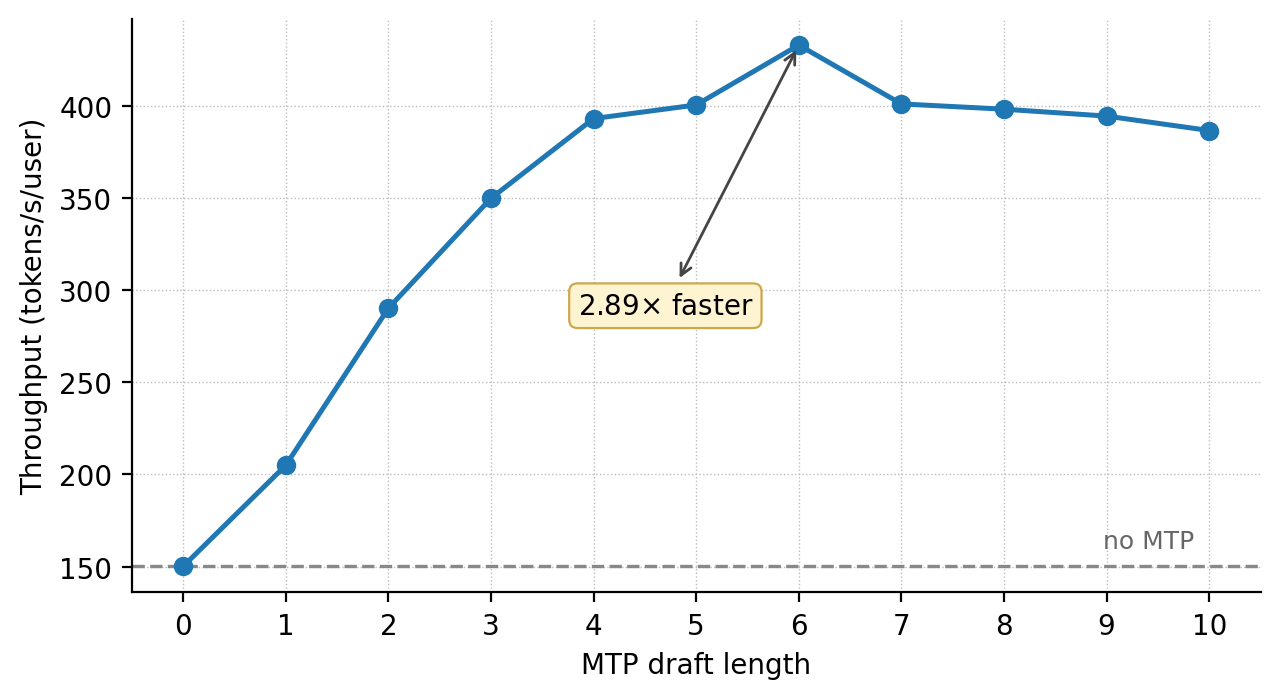}
    \vspace{-0.5em}
    \caption{Decode throughput of the NVFP4 checkpoint as a function of
    MTP draft length on a single-user workload (ISL/OSL/BS = 10K/16K/1,
    single GB200 node, TP=4), using acceptance lengths measured on the
    SPEED-Bench benchmark~\citep{2026speedbench}. The dashed line is the no-MTP
    baseline; throughput peaks at DL=6, giving a $2.89\times$ speedup
    before gradually rolling off as verification overhead outweighs the
    marginal acceptance gain.}
    \label{fig:mtp_dl_throughput}
    \vspace{-0.5em}
\end{figure}

\paragraph{Speculative decoding for hybrid models.}
Multi-Token Prediction (see \S\ref{sec:pretraining}) reduces main-model
forward passes at the cost of running the draft head and verifying
candidates. The operating point depends on batch size: at small batches
per-pass cost is dominated by weight reads, verification overhead is
cheap, and high draft length wins for latency; at large batches per-pass
cost is dominated by compute, verification overhead eats into
throughput, and lower draft lengths, or disabling MTP, often win.
We expose draft length as a deployment-time knob.
Figure~\ref{fig:mtp_dl_throughput} sweeps draft length on a
representative low-latency operating point: the NVFP4 checkpoint on a
single-user workload with ISL/OSL/BS = 10K/16K/1, on a single GB200
node at TP=4.

Hybrid Mamba models raise a second question: how to roll back state when
a draft token is rejected. For pure Attention this is a per-token KV
truncation; the Mamba SSM state, however, is a single fixed-size entry
per sequence that is overwritten every token, so no state corresponding
to an earlier token is directly available. We address this by
snapshotting the SSM state at every draft step. The same snapshotting
mechanism, run at a coarser cadence (every fixed number of tokens), also
gives us prefix caching across requests, otherwise unavailable for
Mamba, since prefix reuse in pure Attention is a free consequence of the
per-token KV cache.

\subsection{Inference at Ultra Scale}
\label{sec:inference:ultra-scale}

\paragraph{Parallelism choices and wide expert parallelism.}
\ourmodel{} is too large to fit on a single GPU and must be parallelized across
multiple ranks. The relevant axes are tensor parallelism (TP, which shards
the weight matrices of each linear layer across ranks), expert parallelism
(EP, which distributes full routed experts across ranks while Attention
and Mamba remain data-parallel), and combinations of the two, TP within
a group of ranks for Attention and Mamba, with routed experts EP-sharded
across and within those groups.
At small batch sizes inference is bound by weight-read memory bandwidth,
and wider TP wins: spreading weights across more ranks reduces per-step
HBM traffic by the TP factor, while the NVLink AllReduce of a single
hidden-state vector is negligible by comparison. At large batch sizes
communication of activation tensors becomes the bottleneck, and EP wins:
inter-rank traffic is confined to the all-to-all dispatch and combine that
bracket each routed-expert block, with no AllReduce on the critical path
of any GEMM, Attention, Mamba, and the routed experts all run
rank-local under EP.
For \ourmodel{}, this makes wide EP the practical choice for high-throughput
serving and wide TP the practical choice for low-latency serving; in some
low-latency settings we have also observed that combining TP and EP, 
along with DP for Attention and Mamba, outperforms either alone. Wide-EP serving in particular requires careful
load balancing across data-parallel ranks: because the routed-expert
all-to-all is synchronous, an under-loaded rank stalls the entire group,
so we route incoming requests to the least-loaded worker to keep per-rank
token counts close. Expert-level imbalance within each MoE layer can be
similarly addressed by replicating hot experts across EP ranks (EPLB). GB200 NVL72 is well-matched to all of these
configurations, most significantly for high-throughput: all 72 GPUs share
a single NVLink domain, so wide EP can span the full system without paying
for cross-domain interconnects in the all-to-all path. Wide TP benefits
from the same property for latency-bound serving, though only up to the
point at which per-rank weight reads shrink enough that AllReduce
becomes the new bottleneck.

\paragraph{Prefill-decode disaggregation for hybrid models.}
Prefill and decode stress the hardware very differently, as the workload
comparison above shows; serving both phases on a single replica therefore
forces a single parallelism configuration and scheduling policy across two
workloads with very different requirements.
Prefill-decode disaggregation, which runs the two phases on separate
workers with their own parallelism and scheduling, is the established
response, and we adopt it for \ourmodel{}. Disaggregation requires transferring
the KV cache from prefill to decode workers on every request; for hybrid
Mamba-Attention models extra care is needed to ensure both the KV cache
and the Mamba SSM state are transferred correctly, and that consumers of
cache-state events can tell the two apart. We landed the necessary
upstream changes, including the semantic KV-event metadata path that
identifies hybrid cache groups and the NIXL side-channel host-resolution
fix needed for multi-node Ray, and disaggregation for hybrid
Mamba-Attention models now works out-of-the-box in
vLLM~\citep{kwon2023efficientmemorymanagementlarge}. End-to-end we
currently measure up to roughly 10\% throughput improvement from
disaggregation on \ourmodel{} on prefill-heavy workloads, and expect
further gains as the software stack matures.

\paragraph{All-to-all backend.}
Under expert parallelism, the routed-expert all-to-all is the dominant
inter-rank communication during MoE serving. It accounts for roughly
15\% to 20\% of total runtime on prefill-heavy workloads
(50K input / 2K output tokens). The default vLLM scheme implements this all-to-all as an AllGather of
activations followed by a ReduceScatter of expert outputs, simple,
dependency-free, and topology-agnostic, but intrinsically wasteful:
AllGather replicates every token on every rank, when each rank only
needs the subset routed to its experts. We
evaluated several true all-to-all backends and adopted FlashInfer's
NVLinkOneSided implementation~\citep{nvidia2026nvlinkonesided} as the
best end-to-end choice for \ourmodel{} serving on GB200 NVL72. End-to-end we measure roughly 5\% throughput
improvement from this backend over the default vLLM scheme; performance
analysis indicates the all-to-all kernel itself still has room for
further optimization. A more fundamental alternative is to avoid the
all-to-all altogether: DWDP~\citep{nvidia2026dwdp} keeps execution
data-parallel and instead pulls each layer's expert weights to every rank,
prefetched behind the preceding compute so the transfer is hidden.

\paragraph{Prefill and MoE chunking.}
On prefill-heavy workloads (e.g.\ 50K input / 2K output), the natural
scheduling choice is to chunk prefill at the request level so chunks
can be interleaved with decode forward passes. Small chunks at
moderate-to-high batch sizes leave essentially no pure-decode passes,
hurting user interactivity (TPOT), and at large batch sizes the number
of prefill passes per request can exceed the number of decode passes.
Growing the chunk size resolves this, but exposes a second issue under
wide expert parallelism: while Attention and Mamba are DP-sharded and
each rank processes only $\text{concurrency} / \text{DP}$ requests,
the routed-expert kernels process the union of tokens from all
data-parallel ranks. Large prefill chunks therefore drive the routed-expert kernels into a
regime where kernel-level resource limits become binding. The fix is
MoE-side chunking, splitting the token batch within the MoE kernel
itself, so chunk size can grow without hitting these limits. We
landed this fix upstream in vLLM.

\paragraph{GEMM dimensions and weight padding.}
Several combinations of tensor parallelism factor, quantization format,
and target hardware land the per-rank GEMM dimensions at sizes that
violate kernel alignment requirements. In every such case the route we
have taken is to pad the affected weight matrices at load time so the
kernel sees an acceptable shape, with runtime infrastructure adjusted
to ignore the padded positions. Examples include MoE NVFP4 kernels
whose hidden- or intermediate-dimension alignment requirements exceed
what naturally falls out of the architecture, and Marlin NVFP4 linear
and MoE kernels on Hopper that impose tile- and thread-level alignment
constraints on their inputs. A natural improvement for future model
generations is to choose the model's inner dimensions such that all
expected (TP, quantization, hardware) tuples produce kernel-friendly
shapes without requiring load-time padding.

\section{Conclusion}
\label{sec:conclusion}
We present our most capable model yet – Nemotron 3 Ultra with 550 billion total and 55 billion active parameters. Nemotron 3 Ultra uses a MoE Hybrid Mamba-Attention architecture along with LatentMoE and MTP for optimal inference and accuracy. Nemotron 3 Ultra was pre-trained on 20 trillion text tokens and then post-trained using SFT, RL, and MOPD. We show that our model attains ~5x higher inference throughput than other state-of-the-art open LLMs while achieving on-par accuracy. We open-source the pre-trained, post-trained, and quantized checkpoints along with the training data for Nemotron 3 Ultra on HuggingFace.

\section*{Contributors}

We thank the following people for their invaluable contributions to \ourmodelfull.

Aaron Blakeman, Aaron Thomas, Aastha Jhunjhunwala, Abhibha Gupta, Abhinav Khattar, Adam Rajfer, Adi Renduchintala, Adil Asif, Aditya Vavre, Adriana Flores Miranda, Ahmad Bilal, Aileen Zaman, Ajay Hotchandani, Akanksha Shukla, Akhiad Bercovich, Aleksander Ficek, Alex Gronskiy, Alex Kondratenko, Alex Steiner, Alex Ye, Alexander Bukharin, Alexandre Milesi, Ali Taghibakhshi, Alice Gatti, Alisa Liu, Alok Kumar, Amar Phanishayee, Ameya Sunil Mahabaleshwarkar, Amir Klein, Amit Zuker, Amnon Geifman, Anahita Bhiwandiwalla, Ananth Subramaniam, Andrea Santilli, Andrew Fulks, Andrew McHarg, Andrew Tao, Andrii Skliar, Anjulie Agrusa, Ankur Srivastava, Ankur Verma, Anna Shors, Anna Warno, Antoni-Joan Solergibert I Llaquet, Arham Mehta, Arkadiusz Nowaczynski, Arti Jain, Ashwath Aithal, Ashwin Poojary, Asif Ahamed, Asit Mishra, Asma Kuriparambil Thekkumpate, Atefeh Sohrabizadeh, Avinash Kaur, Avinash Vem, Ayush Dattagupta, Barath Subramaniam Anandan, Bardiya Sadeghi, Ben Lanir, Benedikt Schifferer, Besmira Nushi, Bilal Kartal, Bill Thiede, Bita Darvish Rouhani, Bo Deng, Bob Schatz, Boris Ginsburg, Boxin Wang, Brad Nemire, Brandon Norick, Brian Dang, Brian Westphal, Brian Yu, Brucek Khailany, Bryan Catanzaro, Carlo del Mundo, Caryln Aarish, Chankyu Lee, Chantal Hwang, Charbel Sakr, Charles Wang, Charlie Truong, Chen Cui, Cheng Cheng, Cheng-Ping Hsieh, Chenghao Zhang, Chenhui Deng, Chintan Patel, Chris Alexiuk, Christian Cosgrove, Christian Munley, Christine Harvey, Christopher Parisien, Chunyang Shen, Coco Li, Collin Neale, Cynthia Gao, Cyril Meurillon, Dan Gil, Dan Su, Dan Zhao, Dane Corneil, Daniel Afrimi, Daniel Egert, Daniel Korzekwa, Daniel Lo, Daniel Machlab, Daniel Serebrenik, Daniil Sorokin, Daria Gitman, Daria Levy, Darko Stosic, David Mosallanezhad, David Yu, Davit Karamyan, Deena Donia, Deep Debroy, Deepak Narayanan, Devin O’Kelly, Dheeraj Peri, Dhruv Nathawani, Di (Allan) Wu, Dima Rekesh, Divyanshu Kakwani, Donald Plummer, Dong Anh, Dongfeng Yu, Dongfu Jiang, Donnie Kim, Dorrin Poorkay, Duncan Riach, Dusan Stosic, Dustin VanStee, Eavan Meng, Edgar Minasyan, Edward Lin, Eileen Margaret Peters Long, Elad Sarafin, Elad Segal, Elena Lantz, Ellie Evans, Elliott Ning, Eric Chung, Eric Harper, Eric Pham-Hung, Eric Tramel, Eric Yang, Erick Galinkin, Erik Pounds, Erika Goncalves Goncalves, Evan Briones, Evan Wu, Evelina Bakhturina, Evgeny Tsykunov, Ewa Dobrowolska, Faisal Ladhak, Farzan Memarian, Fay Wang, Fei Jia, Felipe Soares, Felipe Vieira Frujeri, Feng Chen, Fengguang Lin, Ferenc Galko, Frank Sun, Frankie Siino, Frida Hou, Gal Hubara Agam, Gal Kaplun, Gantavya Bhatt, Gargi Prasad, Garvit Kulshreshtha, George Armstrong, Gerald Shen, Giulio Borghesi, Gordana Neskovic, Gorkem Batmaz, Grace Lam, Greg Mason, Greg Pauloski, Grigor Nalbandyan, Grzegorz Chlebus, Grzegorz Karch, Guan-Ting Liu, Guoming Zhang, Guyue Huang, Haggai Maron, Haifeng Qian, Haim Elisha, Haoxing Ren, Haran Kumar Shiv Kumar, Haribhau Hud, Harris Nover, Harrison Saturley-Hall, Hayate Iso, Helen Ngo, Herbert Hum, Herman Sahota, Hexin Wang, Himanshu Soni, Hovhannes Tamoyan, Hua Li, Huanhuan Chen, Hui Li, Hui Wang, Huy Nguyen, Ian Chiles, Ido Galil, Ido Shahaf, Igor Gitman, Igor Shovkun, Ilya Loshchilov, Ingo Guehring, Itamar Schen, Itay Levy, Itay Neeman, Ivan Moshkov, Izik Golan, Izzy Putterman, Jaemin Choi, Jakub Slowikowski, Jan Kautz, Jane Polak Scowcroft, Jared Casper, Jatin Mitra, Jeffrey Glick, Jenny Chen, Jesse Oliver, Jiacheng Xu, Jiafan Zhu, Jialin Song, Jian Zhang, Jiantao Jiao, Jiaqi Zeng, Jie Lou, Jim King, Jimmy Zhang, Jingquan Wang, Jinhang Choi, Jinju Chu, Joey Conway, Joey Guman, Johan Jatko, Johannes Rausch, John Kamalu, John Roberts, Johnny Greco, Johnny Mensel, Jonah Alben, Jonas Yang, Jonathan Cohen, Jonathan Raiman, Joseph Jennings, Joshua Mabry, Joshua Pierce, Joyjit Daw, Julien Veron Vialard, Junkeun Yi, Jupinder Parmar, Kajal Jain, Kan Zhu, Kari Briski, Katherine Cheung, Katherine Luna, Keith Willowhawk, Keith Wyss, Keshav Santhanam, Kevin Shih, Kezhi Kong, Khanh Nguyen, Khushi Bhardwaj, Kirthi Shankar Sivamani, Konstantinos Krommydas, Krishna C. Puvvada, Krzysztof Pawelec, Kumar Anik, Kyle Keprios, Kylie Day, Lawrence McAfee, Leo Du, Leon Derczynski, Li Ding, Linda Liu, Lingjie Wu, Lior Kadoch, Lizzie Wei, Luis Vega, Luke Robison, Lun Su, Maarten Van Segbroeck, Maciej Jakub Mikulski, Maer Rodrigues de Melo, Magda Sypula, Mahan Fathi, Makesh Narsimhan Sreedhar, Makesh Tarun Chandran, Manoj Kilaru, Maor Ashkenazi, Marc Cuevas, Marc Romeijn, Marcin Chochowski, Mark Cai, Mark Mozolewski, Markus Kliegl, Marta Stepniewska-Dziubinska, Martyna Patelka, Mattei Machczynski, Matvei Novikov, Mauricio Ferrato, Maximilian Golub, Mehrzad Samadi, Melissa Corpuz, Mengru Wang, Mengxi Wu, Meredith Price, Meriem Boubdir, Micah Schaffer, Michael Andersch, Michael Boone, Michael Gschwind, Michael Lightstone, Michael Loh, Michal Bien, Michal Zawalski, Michelle Gill, Miguel Martinez, Mikail Khona, Mike Chrzanowski, Mike Houston, Mingyuan Ma, Minseok Lee, Mohamed Fawzy, Mohammad Dabbah, Mohammad Shoeybi, Mostofa Patwary, Nabin Mulepati, Najeeb Nabwani, Namit Dhameja, Narimane Hennouni, Natalie Hereth, Nathaniel Pinckney, Nave Algarici, Nave Assaf, Netanel Haber, Nicholas Knight, Nick Reamaroon, Nickson Quak, Nidhi Bhatia, Nikhil Desai, Nikolai Ludwig, Nima Tajbakhsh, Ning Xu, Nir Ailon, Nirmal Juluru, Nitin Nitin, Ofri Masad, Oleg Rybakov, Oleksii Hrinchuk, Oleksii Kuchaiev, Olivia Viessmann, Olivier Delalleau, Oluwatobi Olabiyi, Omer Ullman Argov, Omri Puny, Oren Tropp, Pablo Ribalta, Pallab Bhattacharya, Panos Lampropoulos, Parth Mannan, Pasha Shamis, Patrick Legresley, Paul Gibbons, Pavlo Molchanov, Pawel Morkisz, Peter Dykas, Peter Jin, Pierre-Yves Aquilanti, Pinky Xu, Piotr Januszewski, Piotr Laskiewicz, Pooya Jannaty, Prakash Gurumurthy, Pranav Prashant Thombre, Prasoon Varshney, Pritam Gundecha, Przemek Tredak, Puhui Meng, Qiyu Wan, Rabeeh Karimi Mahabadi, Rachel Oberman, Rachit Garg, Radha Sri-Tharan, Rahul Kandu, Rakshit Sanadhya, Ran El-Yaniv, Ran Zilberstein, Rasoul Shafipour, Ray Macalisang, Rayen Tian, Reka Kovacs, Renjie Pi, Rick Izzo, Rima Shahbazyan, Rishabh Garg, Rishi Puri, Rita Fernandes Neves, Ritchie Zhao, Ritika Borkar, Ritu Gala, Riyad Islam, Robert Clark, Robert Hesse, Robert Kirby, Roger Waleffe, Rohit Watve, Roi Koren, Ron Banner, Ruoxi Zhang, Russell J. Hewett, Ryan Prenger, Ryan Stewart, Ryota Egashira, Sadegh Mahdavi, Saee Paliwal, Sagar Singh, Sahil Modi, Salika Dave, Samantha Shinagawa, Samuel Kriman, Sandip Bhaskar, Sangkug Lym, Sanjay Kariyappa, Sanjeev Satheesh, Saran Vikas Murari, Satish Pasumarthi, Saurabh Mishra, Saurav Muralidharan, Scott Hara, Sean Narentharen, Selvaraj Anandaraj, Seonjin Na, Seonmeyong Bak, Seonmyeong Bak, Sepehr Sameni, Seph Mard, Serge Panev, Seth Henneman, Seth Poulos, Shahar Mor, Shantanu Acharya, Shaona Ghosh, Sharath Turuvekere Sreenivas, Sharon Mendelson, Shaun Kotek, Shawn Wang, Shay Aharon, Shaya Gharghabi , Sheng-Chieh Lin, Shi Chen, Shiqing Fan, Shirish Baskaran, Shreya Gopa, Shrimai Prabhumoye, Shubham Pachori, Shubham Toshniwal, Shuoyang Ding, Shwetha Krishnamurthy, Siddharth Singh, Simeng Sun, Sirshak Das, Sivakumar Arayandi Thottakara, Smita Ithape, Somshubra Majumdar, Soumye Singhal, Sri Harsha Singudasu, Sridhar Bhuvanapalli, Srimukh Veccham, Stas Sergienko, Stefania Alborghetti, Stephen Ge, Su Rong, Sugam Dipak Devare, Sukrit Rao, Sumeet Kumar Barua, Sungsoo Ha, Sunny Gai, Suriya Gunasekar, Suseella Panguluri, Suyog Gupta, Sviataslau Hinzburh, Sweta Priyadarshi, Syeda Nahida Akter, Talor Abramovich, Tan Bui, Tanay Varshney, Tatevik Ter-Hovhannisyan, Teodor-Dumitru Ene, Terry Kong, Thanh Do, Tianhe Zhang, Tiffany Moore, Tijmen Blankevoort, Tim Moon, Tiyasa Mitra, Tom Balough, Tomasz Grzegorzek, Tomasz Hliwiak, Tomer Asida, Tomer Bar Natan, Tomer Keren, Tomer Ronen, Tony Salim, Tony Wang, Traian Rebedea, Tugrul Konuk, Twinkle Vashishth, Udi Karpas, Ushnish De, Vahid Noorozi, Venkat Srinivasan, Venmugil Elango, Vibhor Agrawal, Victor Cui, Vijay Korthikanti, Vikas Mehta, Vinay Rao, Virginia Wu, Vitaly Kurin, Vitaly Lavrukhin, Vladimir Anisimov, Vu Pham, Wanli Jiang, Wasi Uddin Ahmad, Wataru Ishihara, Wei Du, Wei Ping, Weiheng Chai, Wenliang Dai, Wesley Helmholz, Will Jennings, Will Zhu, Wojciech Prazuch, Xiaowei Ren, Xiwen Yu, Yan Breek, Yang Chen, Yang Yu, Yangyi Chen, Yaniv Galron, Yashaswi Karnati, Yejin Choi, Yev Meyer, Yi-Fu Wu, Yian Zhang, Ying Lin, Yonatan Geifman, Yonggan Fu, Youngeun Kwon, Yu Yao, Yugi Guvvla, Yuki Huang, Yunsheng Liu, Zach Moshe, Zachary Newell, Zhilin Wang, Zhiyu Li, Zhongbo Zhu, Zhuolin Yang, Zihan Liu, Zijie Yan, Zsolt-Alon Wertheimer.
\newpage

\bibliography{references}
\bibliographystyle{references}

\appendix

\Needspace{0.8\textheight}

\clearpage
\section{Post-training Evaluations}
\label{appendix:eval}
\subsection{Benchmark Details}
We provide details on specific benchmark settings as below.
\begin{itemize}
    \item \textbf{TauBench V3.} To mitigate premature termination from the user model, we added an additional prompt\footnote{\url{https://github.com/AGI-Eval-Official/tau2-bench-revised}} to the user simulation across all domains. For the banking domain, we used the (terminal\_use) setting which allows the agent to search through the knowledge base through a terminal tool. DeepSeek-V4 models were evaluated under max reasoning. User simulator: GPT-5.2 (low reasoning effort). 8 trial average.
    
    \item \textbf{ProfBench (Search)} We use ProfBench \citep{wang2026profbench} to evaluate model's deep research capability in Professional Work - specifically across Finance MBA, Consulting MBA, Scientific Research in Chemistry PhD and Physics PhD. ProfBench tasks are based on real-world workflows and judged with rubric criteria annotated by professionals working in these domains. Specifically, we run evaluations with a search tool and browse tool enabled in order for the model to identify relevant context from the internet. We run all evaluations at 256K context length without any context management, and report scores averaged over 16 times. 
    
    \item \textbf{Browsecomp}. We evaluate BrowseComp with a custom agentic search harness using Tavily as the search and browsing provider, together with terminal access. To avoid loading all retrieved web content into the model context, search and browse results are persisted to a per-task disk workspace; the model receives only metadata/snippets and can selectively inspect saved pages with shell commands such as grep, head, and sed. This allows the model to retrieve relevant evidence surgically, slows context growth, and preserves all previously collected search information across context resets.
   
    \item \textbf{Vals.ai Finance Agent Benchmark (FAB v1.1).} We evaluate on the Vals.ai Finance Agent Benchmark \citep{bigeard2025finance}, which tests agents on entry-level financial analyst tasks spanning nine different categories, from simple retrieval to financial modeling and market analysis. While the full benchmark reports on a private test set of 337 questions, we evaluate on 200 questions: the 50 publicly available validation samples augmented with 150 additional samples from the privately licensed validation set. Agents are provided four tools in the no-web-search condition: EDGAR search (via SEC API), an HTML page parser, a retrieval tool for querying previously extracted content, and a \texttt{submit} tool for producing the final answer. In the web search condition, a fifth tool, Google web search via Tavily, is additionally available. Since the benchmark is grounded in SEC filings, most questions are expected to be answerable from EDGAR directly; web search serves as a supplementary signal. Accuracy is judged against expert-written ground-truth answers using an LLM-as-judge (GPT-5.2, mode of three evaluations). We closely follow the open-source evaluation harness.\footnote{\url{https://github.com/vals-ai/finance-agent}}
    
\end{itemize}

\subsection{Harness Robustness}

\begin{figure*}[ht]
    \centering

    \begin{subfigure}[t]{0.48\linewidth}
        \centering
        \includegraphics[width=\linewidth]{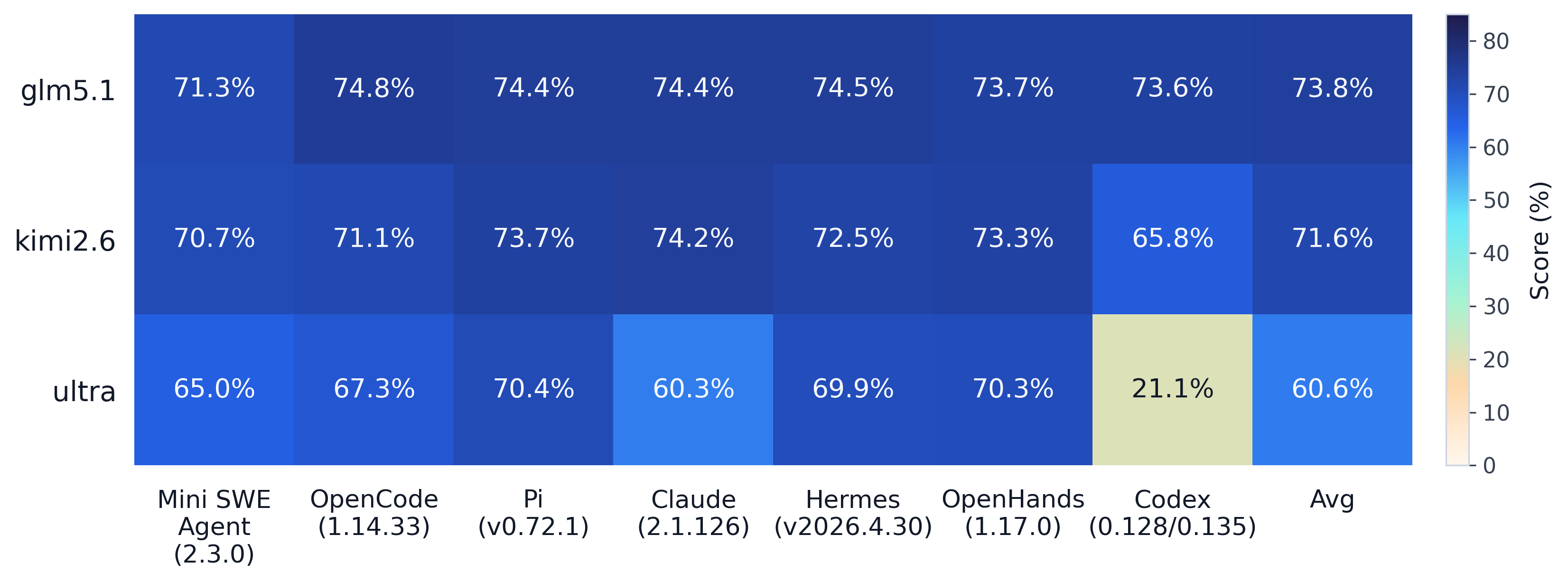}
        \caption{SWE-bench Verified}
        \label{fig:swebench_agent_matrix}
    \end{subfigure}
    \hfill
    \begin{subfigure}[t]{0.48\linewidth}
        \centering
        \includegraphics[width=\linewidth]{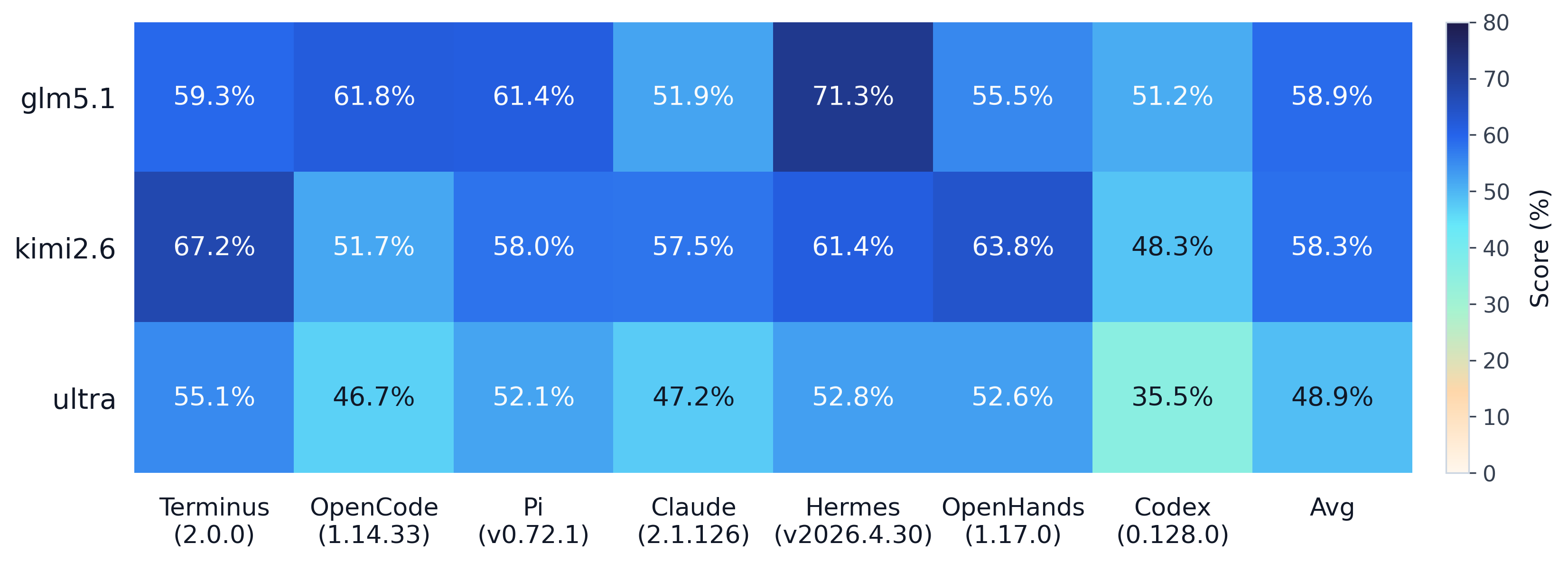}
        \caption{Terminal-Bench 2.1}
        \label{fig:terminal_agent_matrix}
    \end{subfigure}

    \caption{Agent and model matrices for SWE-bench Verified and Terminal-Bench 2.1.}
    \label{fig:agent_model_matrix}
\end{figure*}

Harness Robustness is a key part of the post-training stage for Nemotron-3 Ultra. We categorize all of the task distributions into five verticals, including:
\begin{itemize}
    \item Zero to One Terminal Use and Software Engineering Task
    \item Existing Repo Bug Fixing Tasks
    \item Office and General Productivity Tasks
    \item General/Multi-domain Knowledge Tasks
    \item Search Tasks
\end{itemize}

For each of these input task distributions, we ensure that the model is trained under at least two of the following harnesses:
\begin{itemize}
    \item Stirrup
    \item OpenHands
    \item OpenCode
    \item Terminus
    \item Droid
    \item Custom Internal Harnesses
\end{itemize}

The prevention of the model being trained under a single harness for any given task distribution allows for better generalization and robustness when the model is used under dynamic execution contexts in real world settings. The performance of \ourmodel on different harnesses are shown in Figure \ref{fig:agent_model_matrix}.

\end{document}